\documentclass[10pt]{article} 
\usepackage{times}
\usepackage[left=1.25in, right=1.25in, top=1in, bottom=1in]{geometry}
\usepackage{graphicx}
\usepackage{amsmath}
\usepackage{amssymb}
\usepackage{amsthm}

\newtheorem{definition}{Definition}
\newtheorem{proposition}{Proposition}
\newtheorem{remark}{Remark}
\newtheorem{lemma}{Lemma}
\newtheorem{theorem}{Theorem}

\usepackage{bm}        
\usepackage{lipsum}
\usepackage{amsfonts}
\usepackage{graphicx}
\usepackage{epstopdf}
\usepackage{tabularx}
\usepackage{booktabs}
\usepackage{comment}
\usepackage{subcaption} 
\usepackage{algpseudocode}
\usepackage{algorithm}
\usepackage{placeins}
\usepackage{mathtools}

\usepackage{ae,aecompl}       

\usepackage[utf8]{inputenc} 
\usepackage[T1]{fontenc}    
\usepackage{hyperref}       
\usepackage{url}            
\usepackage{booktabs}       
\usepackage{amsfonts}       
\usepackage{microtype}      

\usepackage{graphics}  
\usepackage{bm}        
\usepackage{epstopdf}
\usepackage{placeins}
\usepackage{float}
\usepackage{dcolumn}   
\usepackage{bm}        
\usepackage{comment}
\usepackage{tikz,ifthen}
\usepackage{color}
\usepackage{hyperref}
\usepackage{placeins}
\usepackage{tabularx}
\usepackage{float}
\usepackage{comment}

\usepackage{natbib}

\DeclareMathOperator{\Tr}{Tr}

\DeclareMathOperator{\Diag}{Diag}

\newcommand{\rmd}{\mathrm{d}}

\newcommand{\rev}[1]{\textcolor{black}{#1}}

\begin{document}

\title{Nystr\"om landmark sampling and regularized Christoffel functions}

\author{ Micha\"el Fanuel\thanks{Most of this work was done when Micha\"el Fanuel was at KU Leuven.}  \\
\normalsize  Univ. Lille, CNRS, Centrale Lille,\\ 
\normalsize UMR 9189 – CRIStAL\\
\normalsize  F-59000 Lille, France \\
\normalsize  \texttt{michael.fanuel@univ-lille.fr} \and Joachim Schreurs \\
 \normalsize ESAT-STADIUS, KU Leuven\\
\normalsize  Kasteelpark Arenberg 10, B-3001 Leuven, Belgium \\
\normalsize  \texttt{joachim.schreurs@kuleuven.be} \and Johan A.K. Suykens \\
\normalsize  ESAT-STADIUS, KU Leuven\\
\normalsize  Kasteelpark Arenberg 10, B-3001 Leuven, Belgium\\
 \normalsize \texttt{johan.suykens@kuleuven.be}}

\maketitle          

\begin{abstract}
Selecting diverse and important items, called landmarks, from a large set is a problem of interest in machine learning.
As a specific example, in order to deal with large training sets, kernel methods often rely on low rank matrix Nystr\"om approximations based on the selection or sampling of landmarks.
In this context, we propose a deterministic and a randomized adaptive algorithm for selecting landmark points within a training data set. These landmarks are related to the minima of a sequence of kernelized Christoffel functions. Beyond the known connection between Christoffel functions and \emph{leverage scores}, a connection of our method with finite \emph{determinantal point processes} (DPPs) is also explained. Namely, our construction promotes diversity among important landmark points in a way similar to DPPs. Also, we explain how our randomized adaptive algorithm can influence the accuracy of Kernel Ridge Regression.
\end{abstract}


\section{Introduction \label{Sec:Intro}}
Kernel methods have the advantage of being efficient, good performing machine learning methods which allow for a theoretical understanding. These methods make use of a kernel function that maps the data from input space to a kernel-induced feature space. Let $\{x_1, \dots, x_n\}$ be a data set of points in $\mathbb{R}^d$ \rev{and let $k(x,y)$ be a strictly positive definite kernel. The associated kernel matrix} is given by $K = [k(x_i,x_j)]_{i,j=1}^n$ and defines the similarity between data points. Just constructing the kernel matrix already requires \rev{$O(n^2)$} operations. Solving a linear system involving $K$, such as in kernel ridge regression, has a complexity $O(n^3)$. Therefore, improving the scalability of kernel methods has been an active research question. Two popular approaches for large-scale kernel methods are often considered: Random Fourier features~\citep{RandomFeatures} and 
the Nystr\"om approximation~\citep{WilliamSeeger}. The latter is considered more specifically in this paper. The accuracy of Nystr\"om approximation depends on the selection of good landmark points and thus is a question of major interest. A second motivation, besides improving the accuracy of Nystr\"om approximation, is selecting a good subset for regression tasks.  Namely in~\citet{cortes2010impact} and \citet{Fastdpp}, it was shown that a better kernel approximation results in improving the performance of learning tasks.  This phenomenon is especially important in stratified data sets~\citep{valverde2014100,oakden2019hidden,chen2019slice}. In these data sets, it is important to predict well in specific subpopulations or outlying points and not only in data dense regions. These outlying points can, e.g., correspond to serious diseases in a medical data set, being less common than mild diseases. Naturally, incorrectly classifying these outliers could lead to significant harm to patients. Unfortunately, the performance in these subpopulations is often overlooked. This is because aggregate performance measures such as MSE or sensitivity can be dominated by larger subsets, overshadowing the fact that there may be a small `important' subset where performance is bad. These stratifications often occur in data sets with `a long tail', i.e., the data distribution of each class is viewed as a mixture of distinct subpopulations~\citep{feldman2019does}. Sampling algorithms need to select landmarks out each subpopulation to achieve a good generalization error. By sampling a \emph{diverse} subset, there is a higher chance of every subpopulation being included in the sample. In this paper, the performance of learning tasks on stratified data is therefore measured on two parts: the bulk and tail of data. The bulk and the tail of data correspond to points with low and high outlyingness respectively, where outlyingness is measured by the ridge leverage scores \rev{ which are described below.}
\paragraph{Leverage score sampling.}
A commonly used approach for matrix approximation is the so-called \rev{Ridge Leverage Score (RLS) sampling. The RLSs} intuitively correspond to the correlation between the singular vectors of a matrix and the canonical basis elements. In the context of large-scale kernel methods, recent efficient approaches indeed consider leverage score sampling~\citep{Gittens2016,Drineas:2005,Bach2013} which results in various theoretical bounds on the quality of the Nystr\"om approximation. Several refined methods also involve recursive sampling procedures~\citep{ElAlaouiMahoney,MuscoMusco,Rudi2017,Rudi2018} yielding statistical guarantees. In view of certain applications where random sampling is less convenient, deterministic landmark selection based on leverage scores
can also be used as for instance in~\citet{Belabbas369}, \citet{Papailiopoulos} and \citet{McCurdy}, whereas a greedy approach was also proposed in~\citet{GreedyNystr}. 
While methods based on leverage scores often select `important' landmarks, it is interesting for applications that require high sparsity to promote samples with `diverse' landmarks since fewer landmarks correspond to fewer parameters for instance for Kernel Ridge Regression and LS-SVM~\citep{suykens2002}. This sparsity is crucial for certain (embedded) applications that often necessitate models with a \emph{small number of parameters} for having a \emph{fast prediction} or due to memory limitations. In the same spirit, the selection of  informative data points~\citep{Guyon:1996,gauthier2018optimal} is an interesting problem which also motivated entropy-based landmark selection methods~\citep{suykens2002,Girolami}.
\paragraph{Determinantal Point Processes.}
More recently, selecting diverse data points by using Determinantal Point Processes (DPPs) has been a topic of active research~\citep{hough2006,KuleszaT12}, finding applications in document or video summarization, image search taks or pose estimation~\citep{gong2014diverse,kulesza2010structured,kulesza2011k,kulesza2011learning}. DPPs are ingenious probabilistic models of repulsion inspired by physics, and which are kernel-based. While the number of sampled points generated by a DPP is in general random, it is possible to define $m$-Determinantal Point Processes ($m$-DPPs) which only generate samples of $m$ points~\citep{kulesza2011k}. $m$-DPP-based landmark selection has shown to give superior performance for the Nystr\"om approximation compared with existing approaches~\citep{Efficientkdpp}. 
However, the exact sampling of a DPP or a $m$-DPP from a set of $n$ items takes $O(n^3)$ operations and therefore several improvements or approximations of the sampling algorithm have been studied in~\citet{Derezinski}, \citet{Fastdpp} \citet{Efficientkdpp} and \citet{Tremblay}. In this paper, we show that \rev{our methods} are capable of sampling equally or more diverse subsets as a DPP, which \rev{gives} an interesting option to sample larger diverse subsets. 
Our paper explains how DPPs and leverage scores sampling methods can be connected in the context of Christoffel functions. These functions are known in approximation theory and defined in Appendix~\ref{app:Christoffel}. They have been `kernelized' in the last years as a tool for machine learning~\citep{ChristOutliers,Pauwels,Lasserre} and are also interestingly connected to the problem of support estimation~\citep{RudiSupport}.
\begin{figure}[t]
        \centering
        \begin{subfigure}[b]{0.32\textwidth}
                \includegraphics[width=\textwidth]{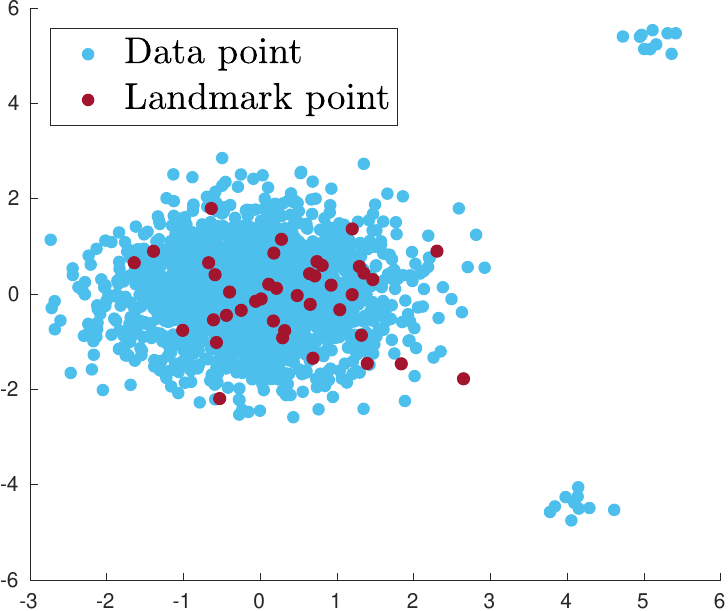}
                \caption{Uniform}              
        \end{subfigure}
        \begin{subfigure}[b]{0.32\textwidth}
                \includegraphics[width=\textwidth]{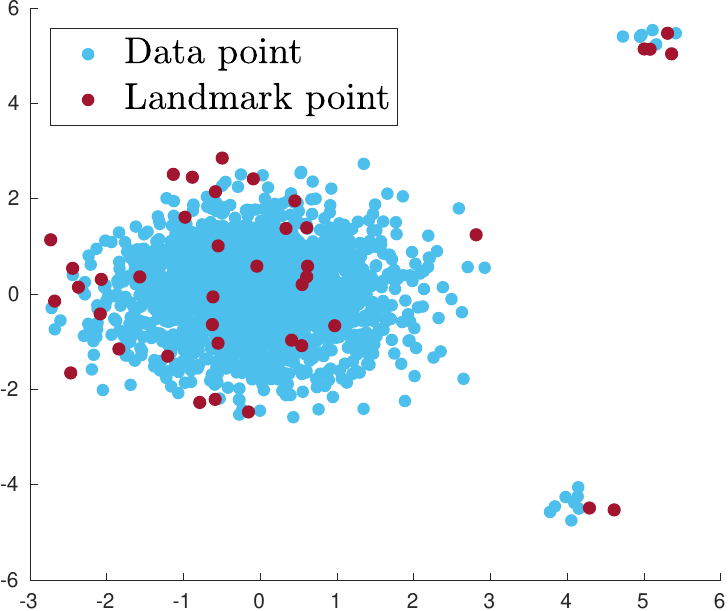}
                \caption{RLS}
        \end{subfigure}
                \begin{subfigure}[b]{0.32\textwidth}
                        \includegraphics[width=\textwidth]{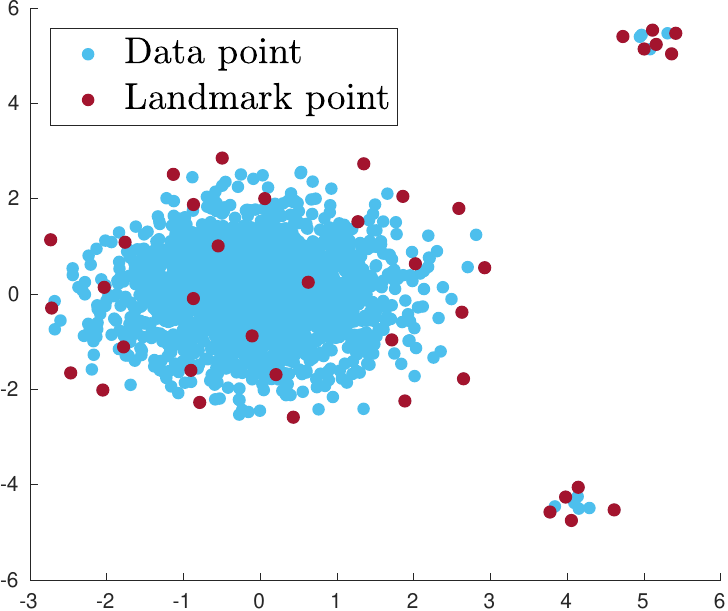}
                        \caption{DAS}
                \end{subfigure}
        \caption{Samplings of an artificial data set. Uniform sampling oversamples dense parts. RLS sampling overcomes this limitation while DAS (proposed in this paper) promotes more diversity.\label{fig:Example}}
\end{figure}

\newpage
\paragraph{Roadmap} 
\begin{enumerate}
    \item We start with an overview of RLS and DPP sampling for kernel methods. The connection between the two sampling methods is emphasized:  both make use of the same matrix that we call here the projector kernel matrix. We also explain Christoffel functions, which are known to be connected to RLSs.
    \item Motivated by this connection, we introduce conditioned Christoffel functions. These {\sl conditioned}  Christoffel functions are shown to be connected to {\sl regularized} conditional probabilities for DPPs in Section~\ref{sec:Christoffel}, yielding a relation between Ridge Leverage Score sampling and DPP sampling.
    \item Motivated by this analogy, two landmark sampling methods are analysed:  Deterministic Adaptive Selection ({DAS}) and  Randomized Adaptive Sampling ({RAS}) respectively in Section~\ref{sec:DAS} and Section~\ref{sec:RAS}. Theoretical analyses of both methods are provided for the accuracy of the Nystr\"om approximation. In particular, RAS is a simplified and weighted version of sequential sampling of a DPP~\citep{Poulson,Desolneux}. Although slower compared to non-adaptive methods, RAS  achieves a better kernel approximation and produces samples with a similar diversity compared to DPP sampling. The deterministic algorithm DAS is shown to be more suited to smaller scale data sets with a fast spectral decay.
    \item Numerical results illustrating the advantages of our methods are given in Section~\ref{sec:Numerics}. The proofs of our results are given in Appendix.
\end{enumerate}

\paragraph{Notations}
Calligraphic letters ($\mathcal{C}, \mathcal{Y}, \dots$) are used to denote sets with the exception of $[n] = \{1,\dots , n\}$. Matrices ($A,K,\dots$) are denoted by upper case letters, while random variables are written with boldface letters. Denote by $a\vee b$ the maximum of the real numbers $a$ and $b$. The $n\times n$ identity matrix is $\mathbb{I}_n$ or simply $\mathbb{I}$ when no ambiguity is possible. The max norm and the operator norm of a matrix $A$ are given respectively by
$\|A\|_{\infty} = \max_{i,j} |A_{ij}|$ and 
$\|A\|_{2} = \max_{\|x\|_2=1} \|A x\|_2$. The Moore-Penrose pseudo-inverse of a matrix $A$ is denoted by $A^+$. Also, we write $A\succeq B$ (resp. $A\succ B$) if $A-B$ is positive semidefinite (resp. definite). Let $v= (v_1,\dots,v_n)^\top\in \mathbb{R}^n$. Then, if $c$ is a real number and $v$ a vector, $v\geq c$ denotes an entry-wise inequality. The $i$-th component of a vector $v$ indicated by $[v]_i$, while $[A]_{ij}$ is the entry $(i,j)$ of the matrix $A$. \rev{Given an $n\times n$ matrix A and $\mathcal{C}\subseteq [n]$, $A_\mathcal{C}$ denotes the submatrix obtained by selecting the columns of $A$ indexed by $\mathcal{C}$, whereas the submatrix $A_{\mathcal{C}\mathcal{C}}$ is obtained by selecting both the rows and columns indexed by $\mathcal{C}$.}
The canonical basis of $\mathbb{R}^d$ is denoted by $\{e_\ell\}_{\ell =1}^d$. 

Let $\mathcal{H}$ be a Reproducing Kernel Hilbert  \rev{Space} (RKHS) with a strictly positive definite kernel $k:\mathbb{R}^d\times \mathbb{R}^d\to \mathbb{R}$, written as $k\succ 0$. Typical examples of strictly positive definite kernels are given by the Gaussian RBF kernel $k(x,y) = e^{-\|x-y\|_2^2/(2\sigma^2)}$ or the Laplace kernel\footnote{By Bochner's theorem, a kernel matrix $K = [k(x_i,x_j)]_{i,j}$ of the Gaussian or Laplace kernel is non-singular, if the $x_i$'s are sampled i.i.d.\ for instance from the Lebesgue measure.} $k(x,y) = e^{-\|x-y\|_2/\sigma}$.  The parameter $\sigma>0$ is often called the bandwidth of $k$. It is common to denote $k_{x}(\cdot) = k(x,\cdot)\in\mathcal{H}$ for all $x\in \mathbb{R}^d$. In this paper, only the Gaussian kernel will be used.
Finally, we write $f(x)\sim g(x)$ if $\lim_{x\to +\infty} f(x)/g(x) = 1$.
\section{Kernel matrix approximation and Determinantal Point Processes \label{sec:KernelApproximation}}
Let $K$  be \rev{an} $n\times n$ positive semidefinite kernel matrix.
In the context of this paper, landmarks are specific data points, within a data set, which are chosen to construct a low rank approximation of $K$. These points can be sampled independently by using some weights such as the ridge leverage scores or thanks to a point process which accounts for both importance and diversity of data points. These two approaches can be related in the framework of Christoffel functions. 
\subsection{Nystr\"om approximation \label{sec:DefNystr\"om}}
In the context of Nystr\"om approximations, the low rank kernel approximation relies on the selection of a subset of data points $\{x_i|i\in \mathcal{C}\}$ for $\mathcal{C}\subseteq [n]$. This is a special case of `optimal experiment design'~\citep{GaoKovalskyDaubechies}. For a small budget $|\mathcal{C}|$ of data points, it is essential to select points so that they are sufficiently diverse and representative of all data points in order to have a good kernel approximation. These points are called \emph{landmarks} in order to emphasize that we aim to obtain both important and diverse data points.
It is common to associate to a set of landmarks a sparse sampling matrix.
\begin{definition}[Sampling matrix]\label{def:Sampling}
Let $\mathcal{C}\subseteq [n]$. A sampling matrix $S(\mathcal{C})$ is \rev{an} $n\times |\mathcal{C}|$ matrix obtained by selecting and possibly scaling by a strictly positive factor the columns of the identity matrix which are indexed by $\mathcal{C}$. When the rescaling factors are all ones, the corresponding unweighted sampling matrix is simply denoted by $C$.
\end{definition}
A simple connection between weighted and unweighted sampling matrices goes as follows. Let $S(\mathcal{C})$ be a sampling matrix. Then, there exists a vector of strictly positive weights $w\in (\mathbb{R}_{>0})^n$ such that $S(\mathcal{C}) = C\Diag(w)$, where $C$ is an unweighted sampling matrix. 
\rev{
An unweighted sampling matrix can be used to select submatrices as follows: let $A$ be an $n\times n$ matrix and $\mathcal{C}\subseteq [n]$, then $A_{\mathcal{C}} = AC$ and $A_{\mathcal{C}\mathcal{C}} = C^\top A C$.} In practice, it is customary to use the unweighted sampling matrix to calculate the kernel approximation whereas the weighted sampling matrix is convenient to obtain theoretical results.

\begin{definition}[Nystr\"om approximation]\label{def:Nystr}
Let $K$ be \rev{an} $n\times n$ positive semidefinite symmetric matrix and let $\mathcal{C}\subseteq [n]$. Given $\mu> 0$  and \rev{an} $n\times |\mathcal{C}|$ sampling matrix $S(\mathcal{C})$, the  \emph{regularized} Nystr\"om approximation of $K$ is
$$
L_{\mu , S}(K) = K S( S^\top K S+ \mu \mathbb{I}_{|\mathcal{C}|})^{-1} S^\top K,
$$
where $S = S(\mathcal{C})$.
The \emph{common} Nystr\"om approximation is simply obtained in the limit $\mu\to 0$ as follows $L_{0, S}(K) = K S( S^\top K S)^{+} S^\top K$.
\end{definition}
\rev{The use of the letter $L$ for the Nystr\"om approximation refers to a Low rank approximation.} The following remarks are in order. 
\begin{itemize}
    \item First, in this paper, the kernel is assumed to be strictly positive definite, \rev{ implying that $K\succ 0$.} In this case, the common Nystr\"om approximation involves the inverse of the kernel submatrix in the above definition rather than the pseudo-inverse. \rev{Thus, the common Nystr\"om approximation is then independent of the weights of the sampling matrix $S(\mathcal{C})$, i.e.,
    \[
    L_{0 , S(\mathcal{C})}(K) = L_{0 , C}(K).
    \]
    }
    \item Second, the Nystr\"om approximation satisfies  the structural inequality $$0\preceq K - L_{\mu, S}(K),$$ for all sampling matrices $S$ and for all $\mu\geq 0$. This can be shown thanks to Lemma~\ref{lem:WNys} in Appendix.
    \item  \rev{Third, if $S = C\Diag(w)$ with $w> 0$ and $K$ is nonsingular}, then we have
    $$
     K- L_{0, C}(K) \preceq K- L_{\mu, S}(K),
    $$
    for all $\mu>0$. The above inequality indicates that we can bound the error on the common Nystr\"om approximation by upper bounding the error on the regularized Nystr\"om approximation.
\end{itemize}

\subsection{Projector Kernel Matrix }
Ridge Leverage Scores and DPPs can be used to sample landmarks. The former allows guarantees with high probability on the Nystr\"om error \rev{by using matrix concentration inequalities involving a sum of i.i.d. terms.} The latter allows particularly simple error bounds on expectation as we also explain hereafter. 
Both these approaches and the algorithms proposed in this paper involve the use of another matrix constructed from $K$ and that we call the \emph{projector kernel matrix}, which is defined as follows
\begin{equation}
P_{n\gamma}(K) = K(K+n\gamma \mathbb{I})^{-1},\label{eq:proj}
\end{equation}
where $\gamma>0$ is a regularization parameter. This parameter is used to filter out small eigenvalues of $K$.
When no ambiguity is possible, we will simply omit the dependence on $n\gamma$ and $K$ to ease the reading.
The terminology `projector' can be understood as follows: if $K$ is singular, then $P_{n\gamma}(K)$ is a mollification of the projector $KK^+$ onto the range of $K$.

We review the role of $P$ for the approximation of $K$ with RLS and DPP sampling in the next two subsections.
These considerations motivate the algorithms proposed in this paper where $P$ plays a central role.
\subsection{RLS sampling \label{sec:RLS}}
Ridge Leverage Scores are a measure of uniqueness of data points and are defined as follows
\begin{equation}
\label{eq:RLS}
\ell_{n\gamma,K}(x_z) = [P_{n\gamma}(K)]_{zz},   
\end{equation}
 with $z\in [n]$. When no ambiguity is possible, the dependence on $K$ is omitted in the above definition. \rev{RLS} sampling is used to design approximation schemes in kernel methods as, e.g., in~\citet{MuscoMusco} and \citet{Rudi2017}. The trace of the projector kernel matrix is the so-called effective dimension~\citep{Zhang}, i.e.,  $d_{\rm eff}(\gamma) = \sum_{i\in [n]}\ell_{n\gamma}(x_i)$. For a given regularization $\gamma >0$, this effective dimension is typically associated with the number of landmarks needed to have an `accurate' low rank Nystr\"om approximation of a kernel matrix as it is explained, e.g., in~\cite{ElAlaouiMahoney}.

To provide an intuition of the kind of guarantees obtained thanks to RLS sampling, we adapt an error bound from~\cite{CalandrielloAISTATS} for Nystr\"om approximation holding with high probability, which was obtained originally in~\citet{ElAlaouiMahoney} in a slightly different form. To do so, we firstly define the probability to sample $i\in [n]$ by $$p_i = \ell_{n\gamma}(x_i)/d_{\rm eff}(\gamma).$$ Then, RLS sampling goes as follows: one samples independently with replacement $m$ landmarks indexed by $\mathcal{C} = \{i_1, \dots, i_m\}\subseteq [n]$ according to the discrete probability distribution given by $p_i$ for $1\leq i\leq n$. Then, the associated sampling matrix is $S = C \Diag(1/\sqrt{m p})$ with $p = (p_{i_1}, \dots , p_{i_m})^\top$.
\begin{proposition}[\cite{ElAlaouiMahoney} Thm. 2 and App. A, Lem. 1]\label{prop:RLS}
Let $\gamma>1/n$ be the regularization parameter. Let the failure probability be $0< \delta< 1$ and fix a parameter  $0< t< 1$. If we sample $m$ landmarks $\mathcal{C} = \{i_1, \dots, i_m\}\subseteq [n]$ with RLS sampling with
$$
m\geq \frac{2d_{\rm eff}(\gamma) + 1/3}{t^2}\log\left(\frac{n}{\delta}\right)
$$
and define the sampling matrix $S= C \Diag(1/\sqrt{m p})$,
then, with probability at least $1-\delta$,
we have 
$$
K - L_{n\gamma,S}(K)\preceq \frac{n\gamma}{1-t}P_{n\gamma} (K)\preceq \frac{n\gamma}{1-t} \mathbb{I}.
$$
\end{proposition}
It is interesting to notice that the bound on Nystr\"om approximation is improved if $\gamma$ decreases, which also increases the effective dimension of the problem.
To fix the ideas, it is common to take $t=1/2$ in this type of results as it is done in~\citet{ElAlaouiMahoney} and~\citet{MuscoMusco} for symmetry reasons within the proofs.
\begin{remark}[The role of column sampling and the projector kernel matrix]
The basic idea for proving Proposition~\ref{prop:RLS} is to decompose the kernel matrix as $K=B^\top B$. Then, we have $P_{n\gamma} (K) = \Psi^\top \Psi$ with $\Psi = (BB^\top + n\gamma \mathbb{I})^{-1/2} B$ as it can be shown by the push-through identity given in Lemma~\ref{lem:push-through} in Appendix. Next, we rely on an independent sampling of columns of $\Psi$ which yields $\Psi S$, where $S$ is a weighted sampling matrix. Subsequently,  a matrix concentration inequality, involving a sum of columns sampled from $\Psi$, is used to provide an upper bound on the largest eigenvalue of $ \Psi \Psi^\top - \Psi SS^\top\Psi^\top$. Finally, the following key identity 
$$
K - L_{n\gamma,S}(K) = n\gamma \Psi^\top\left(\Psi SS^\top \Psi^\top + n\gamma (B B^\top +n\gamma\mathbb{I})^{-1}\right)^{-1}\Psi,
$$
is used to connect this bound with Nystr\"om approximation. A similar strategy is used within this paper to yield guarantees for RAS hereafter.
\end{remark}
\subsection{DPP sampling}
Let us discuss the use of DPP sampling for Nystr\"om approximation.
We define below a \rev{specific} class of DPPs, called $\mathrm{L}$-ensembles,  by following~\citet{hough2006} and \citet{KuleszaT12}. 
\begin{definition}[$\mathrm{L}$-ensemble]
Let $L$ be \rev{an} $n\times n$ symmetric positive-semidefinite matrix\footnote{\rev{For a generalization to non-symmetric matrices, see e.g.~\citet{Gartrell}.}}.
A finite Determinantal Point Process $\mathbf{Y}$ is \rev{an} $\mathrm{L}$-ensemble  if the probability to sample the set $\mathcal{C}\subseteq [n]$ is given by \[\Pr(\mathbf{Y} = \mathcal{C}) = \frac{\det( L_{\mathcal{C}\mathcal{C}})}{ \det (\mathbb{I} + L)}\rev{,}\]
\rev{where $L_{\mathcal{C}\mathcal{C}}$ denotes the submatrix obtained by selecting the rows and columns indexed by $\mathcal{C}$.} The corresponding process is denoted here by $\mathrm{DPP_L}(L)$ \rev{where the subscript $_\mathrm{L}$ indicates it is an $\mathrm{L}$-ensemble DPP, whereas the corresponding matrix $L$ is specified as an argument.}
\end{definition}
In the above definition, the normalization factor is obtained by requiring that the probabilities sum up to unity.
Importantly, for the specific family of matrices $$L = K/(n\gamma), \text{ such that } K\succeq 0 \text{ is a kernel matrix and } \gamma>0,$$ the marginal inclusion probabilities of this process $\Pr(\mathcal{C} \subseteq \mathbf{Y}) = \det( P_{\mathcal{C}\mathcal{C}})$
are associated with the following marginal (or correlation) kernel\footnote{\rev{The marginal (or correlation) kernel of a DPP, usually denoted by $K$, is denoted here by $P$ to prevent any confusion with the kernel matrix.}} $P = P_{n\gamma}(K)$, which is defined in~\eqref{eq:proj}. The expected sample size of $\mathrm{DPP_L}(K/\alpha)$ \rev{for some $\alpha >0$} is given by
  $$\mathbb{E}_{\mathcal{C}\sim \mathrm{DPP_L}(K/\alpha)}(|\mathcal{C}|) = \Tr \left(P_\alpha(K)\right).$$
For more details, we refer to~\cite{KuleszaT12}.
  
 Let us briefly explain the repulsive property of $\mathrm{L}$-ensemble sampling.
 It is well-known that the determinant of a positive definite matrix can be interpreted as a squared volume. A diverse sample spans a large volume and, therefore, it is expected to come with a large probability. For completeness, we give here a classical argument relating diversity and DPP sampling by looking at marginal probabilities. Namely, the probability that  a random subset contains $\{i,j\}$ is given by
\[
\Pr(\{i,j\}\subseteq \mathbf{Y}) =\det\begin{pmatrix}
P_{ii} & P_{ij}\\
P_{ji} & P_{jj}
\end{pmatrix} = \Pr(\{i\}\subseteq \mathbf{Y})\Pr(\{j\}\subseteq \mathbf{Y})-P_{ij}^2.
\]
So, if $i$ and $j$ are very dissimilar, i.e., if $|P_{ij}|$ or $K_{ij}$ is small, the probability that $i$ and $j$ are sampled together is approximately the product of their marginal probabilities. Conversely, if $i$ and $j$ are very similar, the probability that they belong to the same sample is suppressed.
\subsubsection{Nystr\"om approximation error with $\mathrm{L}$-ensemble sampling}
 We now explicitate the interest of landmark sampling with DPPs in the context of kernel methods.
Sampling with $\mathrm{L}$-ensemble DPPs was shown recently to achieve very good performance for Nystr\"om approximation and comes with theoretical guarantees. In particular,  the expected approximation error is given by the compact formula
 \begin{equation}
      \mathbb{E}_{\mathcal{C}\sim \mathrm{DPP_L}(K/\alpha)}\left( K -L_{0,C}(K)\right) = \alpha P_\alpha(K)\preceq \alpha \mathbb{I},\label{eq:ExpErr}
  \end{equation}
 for all $\alpha>0$ as shown independently in~\citet{Fanuel2020DiversitySI} and \citet{Derezinski2020ImprovedGA}. \rev{  In~\eqref{eq:ExpErr}, $C$ denotes the sampling matrix associated to the set $\mathcal{C}$; see Definition~\ref{def:Sampling}}. The analogy with Proposition~\ref{prop:RLS} is striking if we take $\alpha = n\gamma$. Again, a lower error bound on Nystr\"om approximation error is simply $0\preceq K -L_{0,C}(K)$ \rev{(this is a consequence of Lemma 6 in Supplementary Material F of~\cite{MuscoMusco})}, which holds almost surely. The interpretation of the identity~\eqref{eq:ExpErr} is that Nystr\"om approximation error is controlled by the parameter $\alpha>0$ which also directly influences the expected sample size
  $ \Tr \left(P_\alpha(K)\right)$.
Indeed, a smaller value of $\alpha$ yields a larger expected sample size and gives therefore a more accurate Nystr\"om approximation as explained in~\cite{Fanuel2020DiversitySI}.
We want to emphasize that it is also common to define \rev{$m$}-DPPs which are DPPs conditioned on a given subset size. While it is convenient to use \rev{$m$}-DPPs in practice, their theoretical analysis is more involved; see~\cite{Fastdpp} and~\cite{Schreurs2020diversity}.

\subsubsection{Sequential DPP sampling.}
Let $\mathcal{C}$ and $\tilde{\mathcal{C}}$ be disjoint subsets of $[n]$. The general conditional probability of a DPP both involves a condition of the type $\mathcal{C}\subseteq \mathbf{Y}$ and $\mathbf{Y} \cap  \tilde{\mathcal{C}} =\emptyset$. The formula for the general conditional probability of a DPP is particularly useful for sequential sampling, which is also related to the sampling algorithms proposed in this paper.

It was noted recently in~\citet{Poulson} and \citet{Desolneux} that DPP \rev{$\mathbf{Y}$} sampling can be done in a sequential way. We consider here the case of \rev{an} $\mathrm{L}$-ensemble DPP. The sequential sampling of a DPP involves a sequence of $n$ \rev{Bernoulli} trials. Each \rev{Bernoulli} outcome determines if one item is accepted or rejected. The algorithm loops over the set $[n]$ and constructs sequentially the disjoint sets $\mathcal{C}$ and $\tilde{\mathcal{C}}$ which are initially empty. At any step, the set $\mathcal{C}$ contains accepted items  while $\tilde{\mathcal{C}}$ consists of items that were rejected. At step $i$, the element $i\in [n]$ is sampled if the outcome of a  $\text{Bern}(p_i)$ random variable is successful. The success probability is defined as
$$p_i = \Pr(i \in \mathbf{Y} |\mathcal{C}\subseteq \mathbf{Y}, \mathbf{Y} \cap  \tilde{\mathcal{C}} =\emptyset),$$
where $\mathbf{Y}\sim \mathrm{DPP_{L}}(L)$. Then, if $i$ is sampled successfully, we define $\mathcal{C} \leftarrow \mathcal{C} \cup \{i\}$ and otherwise $\tilde{\mathcal{C}} \leftarrow \tilde{\mathcal{C}} \cup \{i\}$.
The general conditional probability is given by
\begin{align}\Pr(z \in \mathbf{Y} |\mathcal{C}\subseteq \mathbf{Y}, \mathbf{Y} \cap  \tilde{\mathcal{C}}=\emptyset) = [H^{\tilde{\mathcal{C}}}- L_{0,C}(H^{\tilde{\mathcal{C}}}) ]_{zz},\label{eq:genCond}
\end{align}
where $\mathcal{C}\cap \tilde{\mathcal{C}} = \emptyset$ and the `conditional marginal kernel' is \begin{equation}H^{\tilde{\mathcal{C}}}= P + P_{\tilde{C}} \big(\mathbb{I} - P_{\tilde{C}\tilde{C}}  \big)^{-1} P_{\tilde{C}}^\top,\label{eq:H}
\end{equation}
with $P = L(L+\mathbb{I})^{-1}$. The expression~\eqref{eq:genCond} can be found in a slightly different form in~\citet[Corollary~1]{Desolneux}. From a computational perspective, $P$ (or $\mathbb{I} - P$) may still have numerically very small eigenvalues so that the inverse appearing in the calculation of the Nystr\"om approximation~\eqref{eq:diff} or in~\eqref{eq:H} has to be regularized.

Exact sampling of a discrete DPP of a set of $n$ items requires in general $O(n^3)$ operations. It is therefore interesting to analyse different sampling methods which can also yield diverse sampling, since exact DPP sampling is often prohibitive for large $n$. 
The role of DPP sampling for approximating kernel matrices motivates the next section where we extend the analogy between Christoffel functions and general conditional probabilities of $\mathrm{L}$-ensemble DPPs.

\section{Christoffel functions and Determinantal Point Processes \label{sec:Christoffel}}
\subsection{Kernelized Christoffel functions \label{sec:KernChristoffel}}
 In the case of an empirical density, the value of the regularized Christoffel function at $x\in \mathbb{R}^d$ is obtained as follows:
\begin{equation}
C_{\eta}(x) = \inf_{f\in\mathcal{H}}\sum_{i=1}^{n} \frac{\eta_i}{n} f(x_i)^2 + \gamma \|f\|_\mathcal{H}^2 \text{ s.t. } f(x) = 1,\tag{$\text{CF}$}\label{eq:Christoffel}
\end{equation}
where $\eta_i> 0$ for $i\in[n]$ are pre-determined weights which can be used to give more or less emphasis to data points. A specific choice of weights is studied in Section~\ref{sec:CondDPP}.
 Notice that, in the above definition, there is an implicit dependence on the parameter $\gamma>0$, the empirical density and the RKHS itself.   The functions defined by~\eqref{eq:Christoffel} are related to the empirical density in the sense that $C_\eta(x)$ takes larger values where the data set is dense and smaller values in sparse regions. This is the intuitive reason why Christoffel functions can be used for outlier detection~\citep{ChristOutliers}.  As it is explained in~\citet[Eq. (2)]{Pauwels}, the Christoffel function takes the following value on the data set
 \begin{equation}
      C_{\eta}(x_i) = \left( e_i^\top K\Big(K\Diag\left(\frac{\eta}{n}\right)K+ \gamma K \Big)^{-1}K e_i\right)^{-1} \text{ for all } i\in [n]. \label{eq:PauwelsChristoffel}
 \end{equation}
The function defined by~\eqref{eq:Christoffel} naturally provides importance scores of data points. The connection with RLSs (see~\eqref{eq:proj} and~\eqref{eq:RLS}) when $\eta = 1$ can be seen thanks to the equivalent expression for $\eta >0$
  \begin{equation}
      C_{\eta}(x_i) =  \frac{\eta_i}{n}\left(e_i^\top K\left(K+ n\gamma \Diag(\eta)^{-1} \right)^{-1} e_i\right)^{-1} \text{ for all } i\in [n], \label{eq:PauwelsChristoffel2}
 \end{equation}
 which is readily obtained thanks to the matrix inversion lemma.
 For many data sets, the `importance' of points is not uniform over the data set, which motivates the use of RLS sampling  and our proposed methods as it is illustrated in Figure~\ref{fig:Example}. For completeness, we give a derivation of~\eqref{eq:PauwelsChristoffel} in Appendix. Clearly, if $\eta>0$, this Christoffel function can also be understood as \rev{an} RLS associated to a reweighted kernel matrix $\bar{K}(\eta) = \Diag(\eta)^{-1/2} K \Diag(\eta)^{-1/2}$, i.e.,
\begin{equation}
    C_\eta(x_i) = n^{-1}/ [\bar{K}(\eta) \left(\bar{K}(\eta) + n\gamma \mathbb{I}  \right)^{-1}]_{ii}\label{eq:ReWRLS}
\end{equation}
 for all $i\in [n]$.
 
 In Section~\ref{sec:CondChrist} below, we first define conditioned Christoffel functions and show their connection with a conditional probability for \rev{an} $\mathrm{L}$-ensemble DPP. This helps to provide some intuition and is a warm-up for the more general results of Section~\ref{sec:CondDPP}.
\subsection{Conditioned Christoffel functions \label{sec:CondChrist}}
We extend~\eqref{eq:Christoffel} in order to provide importance scores of data points which are in the complement $\mathcal{C}^{c}$ of a set  $\mathcal{C}\subseteq [n]$. This consists in excluding points of $\mathcal{C}$ -- and, implicitly, to some extend also their neighbourhood -- in order to look for other important points in $\mathcal{C}^{c}$.  Namely, we define a conditioned Christoffel function as follows:
\begin{equation}
 C_{\mathcal{C},\eta}(x_z)
= \inf_{f\in\mathcal{H}}\sum_{i=1}^{n}\frac{\eta_i}{n} f(x_i)^2 + \gamma \|f\|_\mathcal{H}^2 \text{ s.t. } \begin{cases}
f(x_z) = 1 \text{ and}\\
f(x_s)=0\ \forall s\in \mathcal{C}
 \end{cases},\tag{$\text{CondCF}$}
\label{eq:addConstraints}
\end{equation}
where we restrict here to $x_z$ such that $z\in \mathcal{C}^{c}$. In order to draw a connection with leverage scores, we firstly take uniform weights $\eta_i = 1$ for all $i\in \mathcal{C}^{c}$. In this paper, we propose to solve several problems of the type~\eqref{eq:addConstraints} for different sets $\mathcal{C}\subseteq [n]$. \rev{Then, we will select landmarks $x_z$ which minimize $C_\mathcal{C,\eta}(x_z)$ in the data set for a certain choice of $\eta$.}
\begin{remark}[Extended weights]
Non-uniform weights are considered below in Proposition~\ref{Prop:Weighted}. We anticipate that the constraint $f(x_s)=0$ for all $s\in \mathcal{C}$ can be implemented by considering $\eta_s = +\infty$ for all $s\in \mathcal{C}$, \rev{ in the spirit of an indicator function of a set as it is commonly used in mathematical optimization.}
\end{remark}

Now, we give a connection between Nystr\"om approximation and conditioned Christoffel functions.
Hereafter, Proposition~\ref{prop:obj} gives a closed expression for the conditioned Christoffel function in terms of the common Nystr\"om approximation of the projector kernel matrix. 
\begin{proposition}\label{prop:obj}
Let $k$ be a strictly positive definite kernel and let $\mathcal{H}$ be the associated RKHS. For $\mathcal{C}\subset [n]$, we have
\begin{equation}
C_{\mathcal{C},1}(x_z) = n^{-1}/[P- P_{\mathcal{C}}P_{\mathcal{C}\mathcal{C}}^{-1}P^\top_{\mathcal{C}}]_{zz}, \label{eq:diff}
\end{equation}
for all $ z\in \mathcal{C}^{c}$.
\end{proposition}
Clearly, in the absence of conditioning, i.e., when $\mathcal{C}=\emptyset$, one recovers the result of~\citet{Pauwels} which has related the Christoffel function to the inverse leverage score.
For $\mathcal{C}\neq \emptyset$, the second term at the denominator of~\eqref{eq:diff} is clearly the common Nystr\"om approximation (see Definition~\ref{def:Nystr}) of the projector kernel based on the subset of columns indexed by $\mathcal{C}$.  Also, notice that $P- P_{\mathcal{C}}P_{\mathcal{C}\mathcal{C}}^{-1}P^\top_{\mathcal{C}}$ is the posterior predictive variance of a Gaussian process regression associated with the projector kernel $P$ in the absence of regularization~\citep{rasmussen2006gaussianprocessesformachinelearning}. The accuracy of the Nystr\"om approximation based on $\mathcal{C}$ is given in terms of the max norm as follows:
\[
 \min_{z\in \mathcal{C}^{c}}  C_{\mathcal{C},1}(x_z)=n^{-1}/\|P - L_{0,\mathcal{C}}(P)\|_{\infty} ,
\]
where $L_{0,C}$ is the common low rank Nystr\"om approximation. 

Now, we derive a connection between  conditioned Christoffel functions and conditional probabilities for a specific DPP: \rev{an} $\mathrm{L}$-ensemble which is built thanks to  a rescaled kernel matrix. To begin, we provide a result relating conditioned Christoffel functions and determinants in the simple setting of the last section.
Below, Proposition~\ref{prop:det}, which can also be found in~\citet{Belabbas369} in a slightly different setting, gives an equivalent expression for a conditioned Christoffel function.
\begin{proposition}[Determinant formula]\label{prop:det}
Let $\mathcal{C}\subseteq [n]$ a non-empty set and $z\in  \mathcal{C}^{c}$. Denote $\mathcal{C}_z = \mathcal{C}\cup \{z\}$. For $\eta =1$, the solution of~\eqref{eq:addConstraints} has the form:
\begin{equation}
C_{\mathcal{C},1}(x_z) = n^{-1} \frac{\det P_{\mathcal{C}\mathcal{C}}}{\det P_{\mathcal{C}_z\mathcal{C}_z}}.\label{eq:det}
\end{equation}
Furthermore, let $B$ be a matrix such that $P = B^\top B$. Then,  we have the alternative expression
$
C_{\mathcal{C},1}(x_z) = n^{-1}/\| b_z - \pi_{V_{\mathcal{C}}} b_z\|_2^2,
$
where $b_z\in\mathbb{R}^{N\times 1}$ is the $z$-th column of $B$ and $\pi_{V_{\mathcal{C}}}$ is the projector on $V_{\mathcal{C}} = {\rm span}\{b_s|s\in \mathcal{C}\}$.
\end{proposition}
Interestingly, the determinants in~\eqref{eq:det} can be interpreted in terms of probabilities associated to \rev{an} $\mathrm{L}$-ensemble DPP with $L = K/(n\gamma)$. 
In other words,  Proposition~\ref{prop:det} states that the conditioned Christoffel function is inversely proportional to the conditional probability
\begin{equation}
\Pr(z \in \mathbf{Y} |\mathcal{C}\subseteq \mathbf{Y})=\frac{\det P_{\mathcal{C}_z\mathcal{C}_z}}{\det P_{\mathcal{C}\mathcal{C}}},\label{eq:condProb1}
\end{equation}
for $\mathbf{Y}\sim \mathrm{DPP_L}(K/(n\gamma))$.
 Hence, minimizing $C_{\mathcal{C}, 1}(x_z)$ over $z\in [n]$ produces an optimal $z^\star$ which is diverse with respect to $\mathcal{C}$. 
 Inspired by the identity~\eqref{eq:condProb1}, we now provide a generalization in Section~\ref{sec:CondDPP}, which associates a common Christoffel function to a general conditional probability of a DPP by using a specific choice of weights $\eta$.

\subsection{Regularized conditional probabilities of a DPP and Christoffel functions \label{sec:CondDPP}} 

\begin{figure}
     \centering
     \includegraphics[scale=0.45]{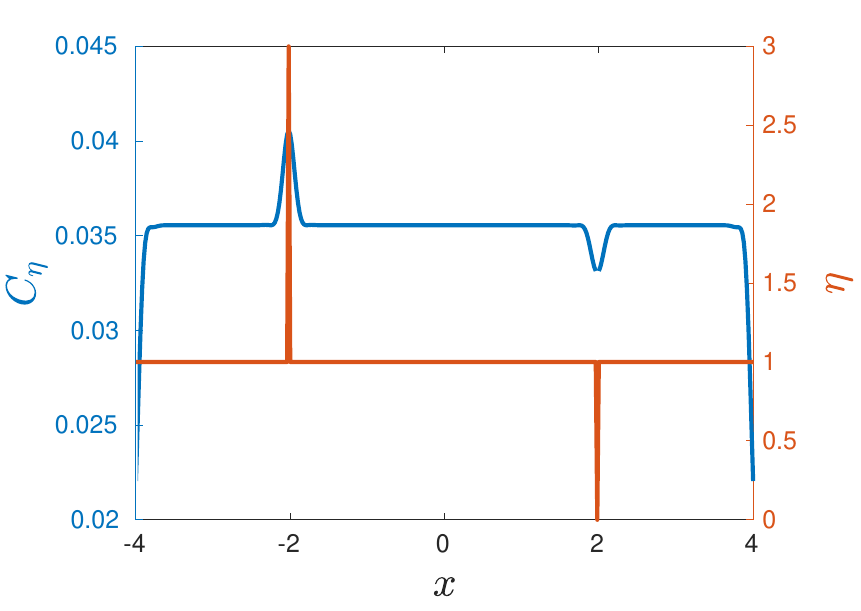}
     \caption{Christoffel function $C_{\eta(\mu,\nu)}(x_i)$ (blue line); see Proposition~\ref{Prop:Weighted} for the precise definition. We take $\gamma = 0.01$ and the Gaussian kernel's bandwidth is $\sigma = 0.2/\sqrt{2}$. The data set is the following grid $x_i = -4+0.02i$ for all the integers $0\leq i\leq 400$. The weights $\eta_{i}(\mu,\nu)$ for $\mu = 0.5$ and $\nu = 0.01$ are displayed in red whereas the corresponding sets are $\mathcal{C} = \{100\}$  and  $\tilde{\mathcal{C}} = \{300\}$.}
     \label{fig:ChristoffelIllustration}
 \end{figure}

 Interestingly, the regularization of the expression of the general conditional probabilities for a DPP in~\eqref{eq:genCond} can be obtained by choosing specific weights $\eta$ in the definition of the Christoffel function.
To do so, the hard constraints in~\eqref{eq:addConstraints} are changed to soft constraints in Proposition~\ref{Prop:Weighted}; see Figure~\ref{fig:ChristoffelIllustration} for an illustration. In other words, the value of $f\in \mathcal{H}$ is severely penalized on $x_i$ when $i\in \mathcal{C}$ in the optimization problem given in~\eqref{eq:addConstraints}. In the same way, $f$ can also be weakly penalized on another disjoint subset $\tilde{\mathcal{C}}$. 
\begin{proposition}[Conditioning by weighting]\label{Prop:Weighted}
Let \rev{$\mu >0$} and $0< \nu < 1$. Let two disjoint subsets $\mathcal{C},\tilde{\mathcal{C}}\subset [n]$ and define $C$ and $\tilde{C}$ the matrices obtained by selecting the columns of the identity indexed by $\mathcal{C}$ and $\tilde{\mathcal{C}}$ respectively. Let $P =  P_{n\gamma}(K)$ and define the following weights: 
\[
\eta_i(\mu,\nu) =\begin{cases}
 1+\mu^{-1} \quad & \text{if } i\in  \mathcal{C}, \\
 \nu \quad & \text{if } i\in  \tilde{\mathcal{C}},\\
  1 \quad &\text{otherwise}.
  \end{cases}
\]
 Then, the Christoffel function defined by~\eqref{eq:Christoffel} associated to $\eta(\mu,\nu)$  is given by
\begin{equation*}
C_{\eta(\mu,\nu)}(x_z) = n^{-1}/ \Big[H^{\tilde{\mathcal{C}}}_{\nu}- L_{\mu ,C}\big(H^{\tilde{\mathcal{C}}}_{\nu}\big) \Big]_{zz}, 
\end{equation*}
where the regularized `conditional marginal kernel' is
\[H^{\tilde{\mathcal{C}}}_{\nu} = P + (1-\nu)P_{\tilde{C}} \big(\mathbb{I} -(1-\nu) P_{\tilde{C}\tilde{C}}  \big)^{-1} P^\top_{\tilde{C}}.
\]
\end{proposition}
Hence, Proposition~\ref{Prop:Weighted} shows that by choosing the weights $\eta$ appropriately the corresponding Christoffel function gives a regularized formulation of conditional probabilities of a DPP if  $\mu>0$ and $0<\nu< 1$ are small enough. Explicitly, we have
$$
\lim_{\mu\to 0}\lim_{\nu\to 0} C_{\eta(\mu,\nu)}(x_z) = n^{-1}/\Pr(z \in \mathbf{Y} |\mathcal{C}\subseteq \mathbf{Y}, \mathbf{Y} \cap  \tilde{\mathcal{C}} =\emptyset),
$$
for $\mathbf{Y}\sim \mathrm{DPP_{L}}(K/(n\gamma))$ and $z\in [n]$, while the conditional probability on the RHS in given in~\eqref{eq:genCond}. 
It is instructive to view the above Christoffel function as a Ridge Leverage Score of a reweighted matrix, bearing in mind the expression of~\eqref{eq:ReWRLS}, 
$$
C_{\eta(\mu,\nu)}(x_i) = n^{-1} \left[\bar{K}(\eta) \left(\bar{K}(\eta) + n\gamma \mathbb{I}  \right)^{-1}\right]_{ii}^{-1}, 
$$
for all $i\in[n]$ and where the reweighted kernel matrix is  $[\bar{K}(\eta)]_{ij} = \frac{K}{\sqrt{\eta_i\eta_j}}$ for all $1\leq i,j\leq n$.
We emphasize that if $\mu$ and $\nu$ take small numerical values, the rows and columns of the reweighted matrix $\bar{K}$ indexed by a subset $\mathcal{C}$ have now very small values while the rows and columns indexed by $\tilde{\mathcal{C}}$ have large numerical values.
\section{Deterministic adaptive landmark selection \label{sec:DAS}}
It is often non trivial to obtain theoretical guarantees for deterministic landmark selection. 
We give here an approach for which some guarantees can be given. Inspired by the connection between DPPs and Christoffel functions, we propose  to select deterministically landmarks by finding the minima of conditioned Christoffel functions defined such that the additional constraints enforce diversity.
As it is described in Algorithm~\ref{Alg1}, we can successively minimize $C_{\mathcal{C}}(x_z)$ over a nested sequence of sets $\mathcal{C}_0\subseteq \mathcal{C}_1\subseteq \dots \subseteq \mathcal{C}_m$ starting with $\mathcal{C}_0 = \emptyset$ by adding one landmark at each iteration. It is not excluded that the discrete maximization problem in Algorithm~\ref{Alg1} can have several maxima. In practice, this is rarely the case. \rev{The ties can be broken uniformly at random or by using any deterministic rule. Hence, in the former case, the algorithm is not deterministic anymore.}
\begin{algorithm}[b]
\begin{algorithmic}
\Statex {\bf input}: Matrix $K\succ 0$, sample size \rev{$m$} and  $\gamma > 0$.
\Statex {\bf initialization}: $\mathcal{C}_0 = \emptyset$ and $l=1$.
\Statex {\bf while}: $l\leq m$ \, {\bf do}
\Statex \quad  $s_{l} \in \arg\max \Diag\left(P-P_{\mathcal{C}_l}P_{\mathcal{C}_l\mathcal{C}_l}^{-1}P^\top_{\mathcal{C}_l}\right)$. 
\Statex \quad  $\mathcal{C}_{l} \leftarrow \mathcal{C}_{l-1}\cup\{s_l\}$ and $l\leftarrow l+1$.
\Statex {\bf end while}
\Statex {\bf return} $\mathcal{C}_m$.
\end{algorithmic} 
\caption{DAS.\label{Alg1}}
\end{algorithm}
In fact, Algorithm~\ref{Alg1} is a greedy reduced basis method as it is defined in~\citet{DeVore2013}, which has been used recently for landmarking on manifolds in~\citet{GaoKovalskyDaubechies} with a different kernel; see also~\citet{SantinHaasdonk}. Its convergence is studied in Proposition \ref{Corol:conv}.
\begin{proposition}[Convergence of DAS]\label{Corol:conv}
Let $\Lambda_1\geq \dots \geq\Lambda_n>0$ be the eigenvalues of the projector kernel matrix $P$ defined in~\eqref{eq:proj}. If $\mathcal{C}_m$ is sampled according to Algorithm~\ref{Alg1}, we have:
\[
\|P-P_{\mathcal{C}_m}P_{\mathcal{C}_m\mathcal{C}_m}^{-1}P^\top_{\mathcal{C}_m}\|_\infty \leq 2 \|P\|_\infty \Lambda^{1/2}_{\left \lfloor m/2\right \rfloor+1}, \text{ for all } 2\leq m<n.
\]

\end{proposition}
Notice that $\|P\|_\infty$ is indeed the largest leverage score related to the maximal marginal degrees of freedom defined in~\citet{Bach2013}.
Furthermore, we remark that the eigenvalues of $P$ are related to the eigenvalues $\lambda_1\geq \dots \geq \lambda_n>0$ of the kernel matrix $K$ by the Tikhonov filtering $\Lambda_m = \frac{\lambda_m}{\lambda_m + n \gamma}$.
In the simulations, we observe that DAS performs well when the spectral decay is fast. This seems to be often the case in practice as explained in \citet{Papailiopoulos} and \citet{McCurdy}.
Hence, since DPP sampling is efficient for obtaining an accurate Nystr\"om approximation, this deterministic adaptation can be expected to produce satisfactory results. However, it is more challenging to obtain theoretical guarantees for the accuracy of the Nystr\"om approximation obtained from the selected subset.
For a set of landmarks $\mathcal{C}$, Lemma~\ref{lem:Kfrom|K} relates the error on the approximation of $K$ to the error on the approximation of $P$.
\begin{lemma}\label{lem:Kfrom|K}
Let $\mu> 0$ and $\tilde{\mu} = \mu/(\lambda_{max}(K)+n\gamma)$. Then, it holds that: \[ 0\preceq \frac{K-L_{\mu,S}(K)}{\lambda_{max}(K)+n\gamma} \preceq  P_{n\gamma}(K)-L_{\tilde{\mu},S}(P_{n\gamma}(K)).\]
\end{lemma} 
This structural result connects the approximation of $K$ to the approximation of $P$ as a function of the regularization $n\gamma$.
However, the upper bound on the approximation of $K$ can be  loose since  $\lambda_{max}(K)$ can be large in general.
It is also true that we might expect the Nystr\"om approximation error to go to zero when the number of landmarks goes to $n$. Nevertheless, the rate of convergence to zero is usually unknown for deterministic landmark selection methods. For more instructive theoretical guarantees, we study a randomized version of Algorithm~\ref{Alg1} in the next section.
\begin{remark}[Related approaches]
Several successful approaches for landmarking are based on Gaussian processes~\citep{PaisleyLandmarking} or use a Bayesian approach~\citep{PaisleyActiveLearning} in the context of active learning. These methods are similar to DAS with the major difference that DAS uses the projector kernel $P = K(K+n\gamma\mathbb{I})^{-1}$, while the above mentioned methods rely on a kernel function $k(x,y)$ or its corresponding kernel matrix.
\end{remark}

\section{Randomized adaptive landmark sampling \label{sec:RAS}}
\rev{RLS} sampling and $\mathrm{L}$-ensemble DPP sampling both allow for obtaining guarantees for kernel approximation as it is explained in Section~\ref{sec:KernelApproximation}. Randomized Adaptive Sampling (RAS) can be seen as an intermediate approach which gives a high probability guarantee for Nystr\"om approximation error with a reduced computational cost compared with DPP sampling. \rev{We think RAS} is best understood by considering the analogy with sequential DPP sampling and Christoffel functions.

As it is mentioned in Section~\ref{sec:CondDPP}, the sequential sampling of a DPP randomly picks items conditionally to the knowledge of previously accepted and rejected items, i.e., the disjoint sets $\mathcal{C}$ and $\tilde{\mathcal{C}}$ defined in Section~\ref{sec:CondDPP}. The computation of the  matrix $H$ in \eqref{eq:H} involves solving a linear system to account for the past rejected samples. RAS uses a similar approach with the difference that the knowledge of the past rejected items is disregarded. Another difference with respect to DPP sequential sampling is that the sampling matrix used for constructing the regularized Nystr\"om approximation is weighted contrary to the analogous formula in~\eqref{eq:genCond}. As explained in Algorithm~\ref{Alg2}, RAS constructs a sampling matrix by adaptively choosing to add (or not) a column to the sampling matrix obtained at the previous step. 
\begin{algorithm}[H]
\begin{algorithmic}
\Statex {\bf input}: Matrix $K\succ 0$, oversampling $c> 0$ and regularizers $\gamma>0$, $\epsilon\in(0,1)$ and $t=1/2$.
\Statex {\bf initialization}: $S_{0}=\emptyset$.
\Statex {\bf for} i=1,\dots, n {\bf do}
\Statex \quad calculate $s_i = \frac{1}{\epsilon} \left[P_{n\gamma}(K) -L_{\epsilon,S_{i-1}}(P_{n\gamma}(K))\right]_{ii}.$
\Statex \quad calculate $p_i = \min\{1, c\tilde{l}_i\}$ with $\tilde{l}_i = \min\{1,(1+t)s_i\}$.
\Statex \quad draw $X_i\sim {\rm \rev{Bernoulli}}(p_i)$.
\Statex \quad {\bf if} $X_i = 1$,
\State \quad \quad add a column to the sampling matrix as follows $S_{i} \leftarrow (S_{i-1} | \frac{e_i}{\sqrt{p_i}} )$.
\Statex \quad {\bf else}
\State\quad \quad $S_{i} \leftarrow S_{i-1}$ (no update).
\Statex \quad {\bf end if}
\Statex {\bf end for}
\Statex {\bf return} $S_n$.
\end{algorithmic}
\caption{RAS. \label{Alg2}}
\end{algorithm}
We now explain the sampling algorithm which goes sequentially over all $i\in[n]$.
Initialize  $\mathcal{S}_{0} = \emptyset$. Let $i\in[n]$ and  denote by $S_{i-1}$ a sampling matrix associated to the set $\mathcal{S}_{i-1}$ at iteration $i-1$. 
A \rev{Bernoulli} random variable determines if $i$ is added to the landmarks or not. If ${\rm \rev{Bernoulli}}(p_i)$ is successful, we update the sampling matrix as follows $S_{i} = (S_{i-1} | \frac{e_i}{\sqrt{p_i}} )$, otherwise $S_{i} = S_{i-1}$. The success probability $p_i$ is defined to depend on the score 
\[
s_i = \frac{1}{\epsilon} \left[P_{n\gamma}(K) -L_{\epsilon,S_{i-1}}(P_{n\gamma}(K))\right]_{ii},\]
by analogy with the conditional probability~\eqref{eq:condProb1} and the Christoffel function value~\eqref{eq:diff}.
The sampling probability is defined as
\begin{equation}
p_i = \min\{1, c\tilde{l}_i\} \ \text{ with } \ \tilde{l}_i = \min\{1,(1+t) s_i\}\label{eq:tildeli},
\end{equation}
with $t>0$ and $c>0$.
\begin{remark}[Oversampling parameter]
In this paper, the positive parameter $c$ is considered as an oversampling factor and is chosen such that $c>1$. This lower bound is less restrictive than the condition on $c$ in Theorem~\ref{thm:Main}. In our simulations, the oversampling parameter is always strictly larger than $1$ so that the expression of the sampling probability simplifies to $p_i = \min\{1, c (1+t) s_i\}$.
\end{remark}
\paragraph{Interpretation of the sampling probability.}
Here, $t>0$ is a parameter that will take a fixed value and plays the same role as the parameter $t$ in Proposition~\ref{prop:RLS}. Therefore $t$ should not take a large value. Then, $\tilde{l}_i$ can be viewed as a leverage score.
To allow for varying the number of sampled points, this leverage score is multiplied by an oversampling factor $c>0$ as it is also done for recursive RLS sampling in~\citet{MuscoMusco}. The resulting score can be larger than 1 and therefore it is capped at $1$.
The idea is that the score $s_i$ is an approximation of an RLS.
Indeed, if we define the following factorization 
\begin{align}
B^\top B = P_{n\gamma}(K),\label{eq:definitionOfB}
\end{align}
an alternative expression of the score is given by \[s_i =[B^\top(B S_{i-1}S_{i-1}^\top B^\top + \epsilon\mathbb{I})^{-1}B]_{ii}\] with $i\in [n]$,  as a consequence of Lemma~\ref{lem:WNys} in Appendix. 
Then, the score $s_i$ is an approximation of the ridge leverage score
$$
 [ B^\top(B B^\top + \epsilon\mathbb{I})^{-1}B]_{ii}=[P_\epsilon(P_{n\gamma}(K))]_{ii},
$$
for $i\in[n]$, with regularization parameter $\epsilon>0$, as it can be shown thanks to the push-through identity\footnote{See Lemma~\ref{lem:push-through} in Appendix.}. In view of Section~\ref{sec:RLS}, the above ridge leverage score can be used for sampling a subset to compute a Nystr\"om approximation of $P_{n\gamma}(K)$ thanks to Proposition~\ref{prop:RLS}. However, we are interested in the Nystr\"om approximation of $K$. The following identity allows us to relate $s_i$ to the approximation of $K$.
\begin{lemma}\label{lem:CompositionRule}
Let $K\succ 0$. Then, we have $P_\epsilon(P_{n\gamma}(K)) = \frac{1}{1+\epsilon}P_{\frac{\epsilon n\gamma}{1+\epsilon}}(K)$.
\end{lemma}
The second regularization parameter $\epsilon$ is typically chosen as $\epsilon=10^{-10}$ so that, in this circumstance, the adaptivity in Algorithm~\ref{Alg2} promotes diversity among the landmarks in view of the discussion in Section~\ref{sec:CondChrist}.
\begin{remark}[Subset size]
Clearly, RAS outputs a subset of random size. This is also the case for $\mathrm{L}$-ensemble DPP sampling, with the notable difference that the mean and variance of the subset size is known exactly in the latter case. For RAS, we empirically observe that the size variance is not large. The parameter $\gamma>0$ influences the effective dimension of the problem and therefore helps to vary the amount of points sampled by RAS. A small $\gamma$ yields a large effective dimension and conversely. Increasing $c$ also increases the sample size which might also reduce diversity.
\end{remark}
\paragraph{Theoretical guarantees.} An error bound for Nystr\"om approximation with RAS can be obtained by analogy with RLS sampling. Our main result in Theorem~\ref{thm:Main} deals with the adaptivity of RAS to provide an analogue of Proposition~\ref{prop:RLS} which was designed for independent sampling.

We can now explain how RAS can be used to obtain a Nystr\"om approximation of $K$ by introducing an appropriate column sampling strategy.
We define the factorization $$P_\epsilon(P_{n\gamma}(K)) = \Psi^\top \Psi,$$ with $\Psi = (B B^\top+\epsilon \mathbb{I})^{-1/2}B$ and with $B$ defined in~
\eqref{eq:definitionOfB}.
Therefore, thanks to Lemma~\ref{Lem:KernelApprox} below, we know that if we can sample appropriately the columns of $\Psi$, denoted by $(\psi_i)_{i=1}^n$, then the error on the corresponding Nystr\"om approximation will be bounded by a small constant.
\begin{lemma}[Nystr\"om approximation from column subset selection]\label{Lem:KernelApprox}
Let $\epsilon>0$ and $0<t<(1+\epsilon)^{-1}$. Denote by $\Psi$ a matrix such that $\Psi^\top \Psi = P_\epsilon(P_{n\gamma}(K))$. If there is a sampling matrix $S$ such that $$\lambda_{max}( \Psi \Psi^\top - \Psi S S^\top  \Psi^\top) \leq t,$$ then,  it holds that
$$
0\preceq K-L_{\frac{\epsilon n\gamma}{1+\epsilon}, S}(K)\preceq \frac{\epsilon n\gamma}{1-t(1+\epsilon)}P_\epsilon(P_{n\gamma}(K)).
$$
\end{lemma}
Some remarks are in order. Firstly, Lemma~\ref{Lem:KernelApprox} has been adapted from~\citet{ElAlaouiMahoney} in order to match our setting and its proof is given in Appendix for completeness. Secondly, the parameter $t$ does not play a dominant role in the above error bound as long as it is not close to $(1+\epsilon)^{-1}$. Indeed, the error bound can be simply improved by choosing a smaller value for the product $\epsilon n\gamma$. Hence, if we require $\epsilon < 1$, then we can choose $t=1/2$ to clarify the result statement. Then, we have the simplified error bound
$$
0\preceq K-L_{\frac{\epsilon n\gamma}{1+\epsilon}, S}(K)\preceq \frac{2\epsilon n\gamma}{1-\epsilon} \mathbb{I},
$$
where we used that $P_\epsilon(P_{n\gamma}(K))\preceq \mathbb{I}$ in Lemma~\ref{Lem:KernelApprox}.

In order to use this result, we want to find a good subset of columns of $\Psi$, i.e., a set of landmarks $\mathcal{S}$ and probabilities $(p_j)_{j\in \mathcal{S}}$ such that 
$$
\lambda_{max}\left( \sum_{i=1}^n\psi_i\psi_i^\top - \sum_{j\in \mathcal{S}}\psi_j\psi_j^\top/p_j\right)\leq t,
$$
for $t>0$, with high probability. This condition can be achieved for well-chosen probabilities $\{p_j : j\in \mathcal{S}\}$ by using a martingale concentration inequality which helps to deal with the adaptivity of RAS. This matrix martingale inequality plays the role of the matrix Bernstein inequality used in the proof of Proposition~\ref{prop:RLS} in~\cite{ElAlaouiMahoney}.

Below, Theorem~\ref{thm:Main} indicates that, if the oversampling $c$ is large enough,  RAS produces a sampling matrix allowing for a `good' Nystr\"om approximation of $K$ with high probability. For convenience, we recall the definition of the negative branch of the Lambert function $W_{-1}:[-e^{-1}, 0)\to \mathbb{R}$ which satisfies $\lim_{y \to 0_{+}} W_{-1}(-y)/\log(y) = 1.$ Also, define  the function $g(a) = -W_{-1}(-1/a)$ for all $a\geq e$, which has the asymptotic behaviour $g(a)\sim \log(a)$.
\begin{theorem}[Main result]\label{thm:Main}
Let $\epsilon\in(0,1)$ and $\gamma>0$. \rev{Let $\delta\in (0,1)$.} If the oversampling parameter satisfies:
\[
c\geq  \frac{28}{3}g\left(\frac{700 d_{\rm eff}(\frac{\epsilon n\gamma}{1+\epsilon})}{3(1+\epsilon)\delta}\right) \vee \frac{1+\sqrt{37}}{3}, 
\]
then, with a probability at least $1-\delta$, Algorithm~\ref{Alg2} yields a weighted sampling matrix $S$ such that
$$
\|K-L_{\frac{\epsilon n\gamma}{1+\epsilon}, S}(K)\|_{2}\leq  \frac{2\epsilon n\gamma}{1-\epsilon}.
$$
\end{theorem}
\begin{remark}[Common Nystr\"om approximation]
In view of the remarks of Section~\ref{sec:DefNystr\"om}, the result of Theorem~\ref{thm:Main} directly implies a bound on the common Nystr\"om approximation,  that is $\|K-L_{0, S}(K)\|_{2}\leq  \frac{2\epsilon n\gamma}{1-\epsilon}$.
\end{remark}
\paragraph{Discussion.}
Let us list a few comments about Theorem~\ref{thm:Main}.
The proof of Theorem~\ref{thm:Main}, given in Appendix, uses the martingale-based techniques developed in~\citet{OnlineRowSampling} in the context of online sampling, together with a refinement by~\citet{MINSKER} of the Freedman-type inequality for matrix martingales due to~\citet{tropp2011}. An advantage of the Freedman inequality obtained in~\citet{MINSKER} is that it does not directly depend on the matrix dimensions but rather on a certain form of intrinsic dimension. 
The martingale-based techniques allow to deal with the adaptivity of our algorithm which precludes the use of statistical results using independent sampling. From the theoretical perspective, the adaptivity does not improve the type of guarantees obtained for independent sampling. The adaptivity of RAS rather makes the theoretical analysis more involved. However, our simulations indicates an empirical improvement.  We emphasize that the dependence on the effective dimension in Theorem~\ref{thm:Main} does not appear in~\citet{OnlineRowSampling} although the proof technique is essentially the same. This extra dependence on $d_{\rm eff}(\frac{\epsilon n\gamma}{1+\epsilon})$ helps to account for the kernel structure although the effective dimension can be however rather large in practice if $\epsilon\approx 10^{-10}$. A small enough $\epsilon$ indeed promotes diversity of sampled landmarks.
The bound of Theorem~\ref{thm:Main} is pessimistic since it depends on the effective dimension which does not account for the subsample diversity. In its current state, the theory cannot yield improved guarantees over independent sampling. 
In practice, the chosen value for the oversampling factor is taken much smaller.  
A major difference with~\citet{ElAlaouiMahoney} and~\citet{MuscoMusco} is that the landmarks $\mathcal{S}$ are \emph{not} sampled independently in Algorithm~\ref{Alg2}.
Algorithmically, RAS also corresponds to running Kernel Online Row Sampling~\citep{calandriello2017second} with the difference that it is applied here on the projector kernel $P_{n\gamma}(K)$.
As it is mentioned hereabove, a natural link can be made between RAS and sequential sampling of a DPP.
RAS involves a regularized score
 $[P-L_{\epsilon,{S}_{i-1}}(P)]_{ii}$ at step $i$, where $S_{i-1}$ is a (weighted) sampling matrix associated to a set $\mathcal{C}_{i-1}\subset [n]$. Indeed, this score depends on the points $\mathcal{C}_{i-1}$ which were chosen before step $i$, but not on the points $\tilde{\mathcal{C}}_{i-1}$ that were \emph{not} chosen (corresponding to a failed \rev{Bernoulli}). Sequential sampling of a DPP at step $i$ also involves drawing a \rev{Bernoulli} with probability  
$\Pr(i \in \mathbf{Y} |\mathcal{C}_{i-1}\subseteq \mathbf{Y}, \mathbf{Y} \cap  \tilde{\mathcal{C}}_{i-1} =\emptyset)$
as given by a conditional kernel in~\eqref{eq:H}. The first major difference with sequential sampling of a DPP is that we do \emph{not} calculate the conditional kernel~\eqref{eq:H}. The second difference is that the linear systems that are solved by RAS are \emph{regularized} and Theorem~\ref{thm:Main} accounts for this regularization. Thirdly, RAS involves a \emph{weighting} of the sampling matrix which does not appear in DPP sampling.
\section{Numerical results \label{sec:Numerics}}

\subsection{Exact algorithms \label{sec:ExactAlgo}}
The performance of deterministic adaptive landmark sampling (DAS) and its randomized variant (RAS) is evaluated on the \texttt{Boston housing}, \texttt{Stock},  \texttt{Abalone} and  \texttt{Bank 8FM} \rev{datasets} which are described in Table~\ref{Table:data}.
\begin{table}[h]
\caption{Small scale data sets and parameters\label{Table:data}}

\begin{center}
         \begin{tabular}{rcccc}
                        \toprule
                        data set & $n$ & $d$ & $c/\epsilon$ & $\sigma$ \\ \midrule
                        \texttt{Housing} & 506 & 13 & 100 & 5 \\
                        \texttt{Stock} & 950 & 10 & 100 & 5 \\
                        \texttt{Abalone} & 4177 & 8 & 150 & 5 \\
                        \texttt{Bank 8FM} & 8192 & 8 & 150 & 10 \\ \bottomrule
                \end{tabular}
  \end{center}

  \end{table}
These public data sets\footnote{\url{https://www.cs.toronto.edu/~delve/data/data sets.html}, \url{https://www.openml.org/d/223}} have been used for benchmarking  m-DPPs in~\citet{Fastdpp}. 
The quality of the landmarks $\mathcal{C}$ is evaluated by calculating $\|K-\hat{K}\|_2/\|K\|_2$ with $\hat{K} = L_{\varepsilon,C}(K)$ with $\varepsilon = 10^{-12}$ for numerical stability.
\begin{figure}[t]
	\centering
	\begin{center}
\begin{minipage}{1.0\textwidth}

	\begin{subfigure}[t]{0.49\textwidth}
	\includegraphics[width=\textwidth]{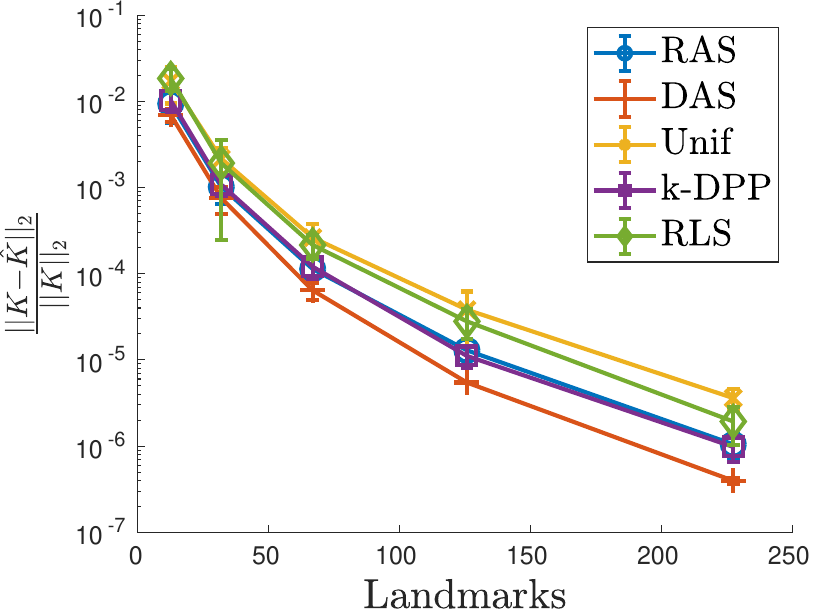}
		\caption{Stock}
	\end{subfigure}
	\quad
	\begin{subfigure}[t]{0.49\textwidth}
	\includegraphics[width=\textwidth]{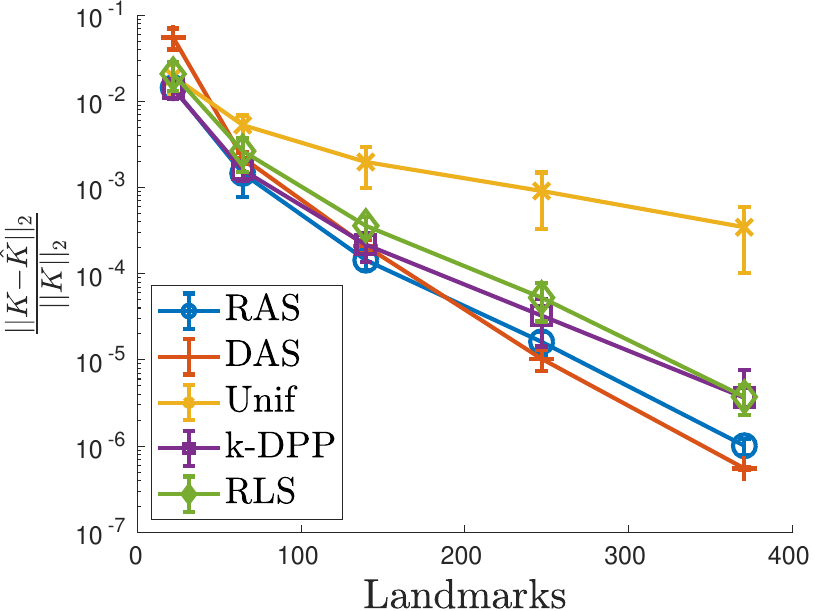}
		\caption{Housing}
	\end{subfigure}
	\end{minipage}
	
		\begin{minipage}{1.0\textwidth}
	\begin{subfigure}[b]{0.49\textwidth}
	\includegraphics[width=\textwidth]{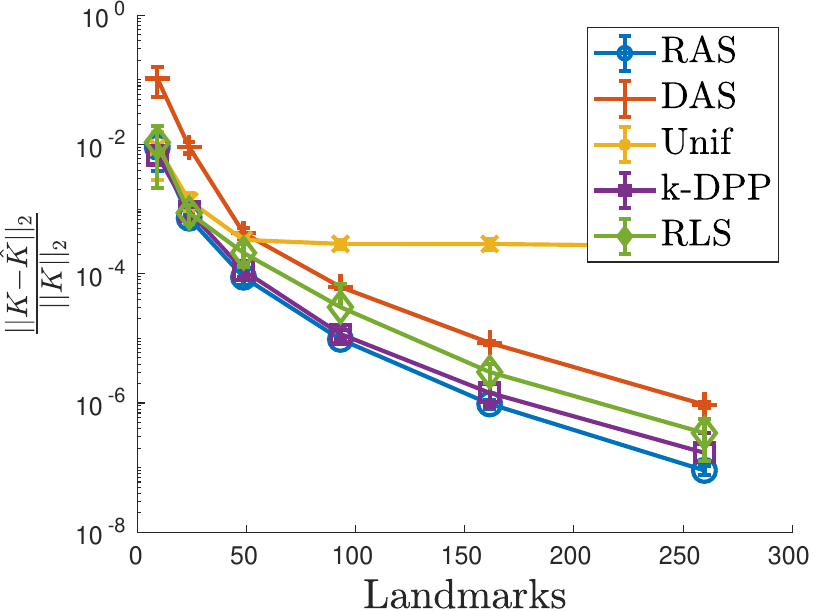}
		\caption{Abalone}
	\end{subfigure}
	\quad
	\begin{subfigure}[b]{0.49\textwidth}
	\includegraphics[width=\textwidth]{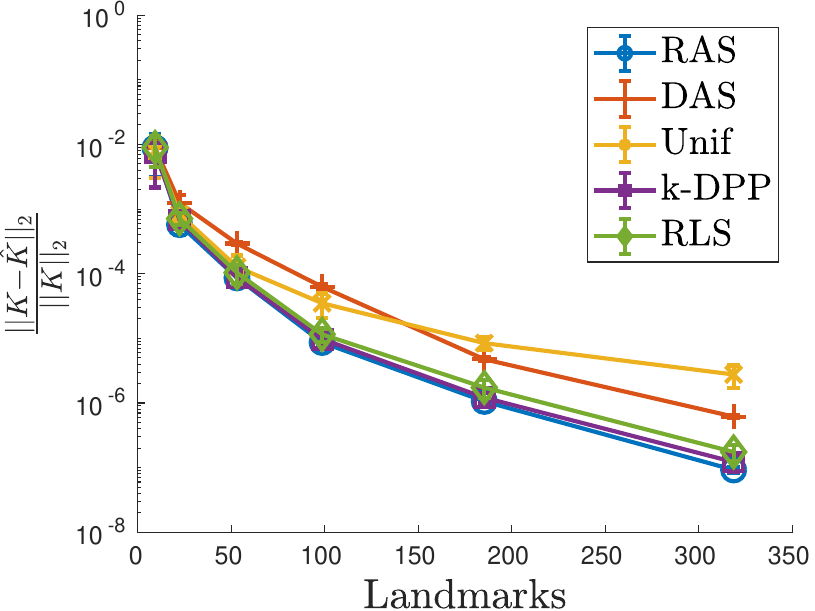}
		\caption{Bank8FM}
	\end{subfigure}	
	\end{minipage}

	\end{center}
	\caption{Relative operator norm of the Nystr\"om approximation error as a function of the number of landmarks. The error is plotted on a logarithmic scale, averaged over 10 trials. The error bars are standard deviations.\label{fig:OP}}
\end{figure}
A Gaussian kernel is used with a fixed bandwidth $\sigma>0$  after standardizing the data. We compare the following exact algorithms: Uniform sampling (Unif.), m-DPP\footnote{Matlab code: \url{https://www.alexkulesza.com/}.}
, exact ridge leverage score sampling (RLS)
and DAS.  The experiments have the following structure:
\begin{enumerate}
    \item First, RAS is executed for an increasing $\gamma \in \{10^{0}, \dots , 10^{-4}\}$ for the \texttt{Boston housing} and \texttt{Stock} data sets, $\gamma \in \{10^{0}, \dots , 10^{-5}\}$ for the \texttt{Abalone} and  \texttt{Bank 8FM} data sets and $\epsilon= 10^{-10}$. With each regularization parameter $\gamma$, RAS returns a different subset of $m$ landmarks. The same subset size $m$ is used for the comparisons with the other methods. 
    \item For each subset size $m$, the competing methods are executed for multiple hyperparameters $\gamma \in \{10^0 , 10^{-1} , \dots , 10^{-6} \}$. The parameter $\gamma$ here corresponds to the regularization parameter for the RLSs (see \eqref{eq:RLS}), 
    and the regularization parameter of DAS. The best performing $\gamma$ is selected for each subset size $m$ and sampling algorithm. In this manner, we each time compare the best possible sampling every algorithm can provide given the subset size $m$.
\end{enumerate}
This procedure is repeated 10 times and the averaged results are visualized in Figure~\ref{fig:OP}. Notice that the error bars are small. 
DAS performs well on the \texttt{Boston housing} and \texttt{Stock} data set, which show a fast decay in the spectrum of $K$ (see Figure~\ref{fig:Eig} in Appendix). This confirms the expectations from Proposition~\ref{Corol:conv}.  If the decay of the eigenvalues is not fast enough, the randomized variant RAS has a superior performance which is similar to the performance of a m-DPP. For our implementation, RAS is faster compared with m-DPP, as it is illustrated in Figure~\ref{fig:Time} in Appendix. The main cost of RAS is the calculation of the projector kernel matrix which requires $O(n^3)$ operations. The exact sampling of a m-DPP has a similar cost. As a measure of diversity the $\mathrm{log}\mathrm{det}(K_{\mathcal{C}\mathcal{C}})$ is given in Figure~\ref{fig:DetEr} in Appendix. The DAS method shows the largest determinant, RAS has a similar determinant as the m-DPP.

\subsection{Approximate RAS}
For medium-scale problems, we propose an approximate RAS method. First, the Gaussian kernel matrix $K$ is approximated by using random Fourier features~\citep{RandomFeatures}, that is,  a generic method which is not data-adaptive.
In practice, we calculate $F\in \mathbb{R}^{n\times n_F}$  with  $n_F = 4000$ random Fourier features such that $FF^\top$ approximates $K$. Then,  we obtain $\hat{P}_{n\gamma} = F \left(F^\top F + n\gamma \mathbb{I}\right)^{-1} F^\top$, where the linear system is solved  by using a Cholesky decomposition. This approximate projector matrix is used in the place of $P_{n\gamma}$ in Algorithm~\ref{Alg2}. Next, we use the matrix inversion lemma in order to update the matrix inverse needed to calculate the sampling probability in Algorithm~\ref{Alg2}. 
The quality of the obtained landmarks is evaluated by calculating the averaged error $\|K_{\mathcal{A}\mathcal{A}}-\hat{K}_{\mathcal{A}\mathcal{A}}\|_F$ over $50$ subsets $\mathcal{A}\subset [n]$ of $2000$ points sampled uniformly at random. The Frobenius norm is preferred to the $2$-norm to reduce the runtime. A comparison is done with the recursive leverage score sampling method\footnote{Matlab code:  \url{https://www.chrismusco.com/}.} (RRLS) developed in~\citet{MuscoMusco} and BLESS which was proposed in~\citet{Rudi2018}.
The parameters of BLESS are chosen in order to promote accuracy rather than  speed (see Table~\ref{tab:BLESSparameters} in Appendix). This algorithm was run several times until the number of sampled landmarks $n_{BLESS}$ satifies $|\mathcal{C}|\leq n_{BLESS}\leq |\mathcal{C}| + 20$. 
\begin{figure}[h]
	\centering
	\begin{subfigure}[b]{0.45\textwidth}
	\includegraphics[width=0.9\textwidth]{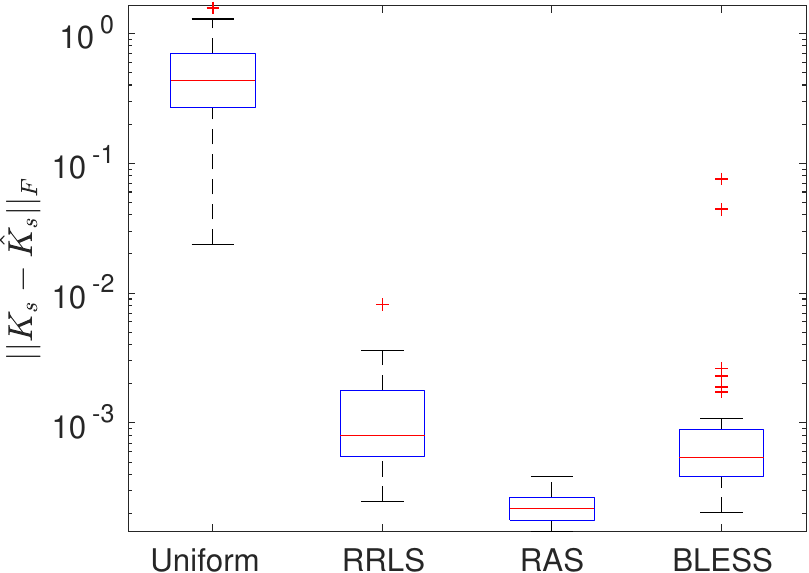}
		\caption{$|\mathcal{C}| =2378 $, $\sigma = 5$.}
	\end{subfigure}
	\hfill
	\begin{subfigure}[b]{0.45\textwidth}
	\includegraphics[width=0.9\textwidth]{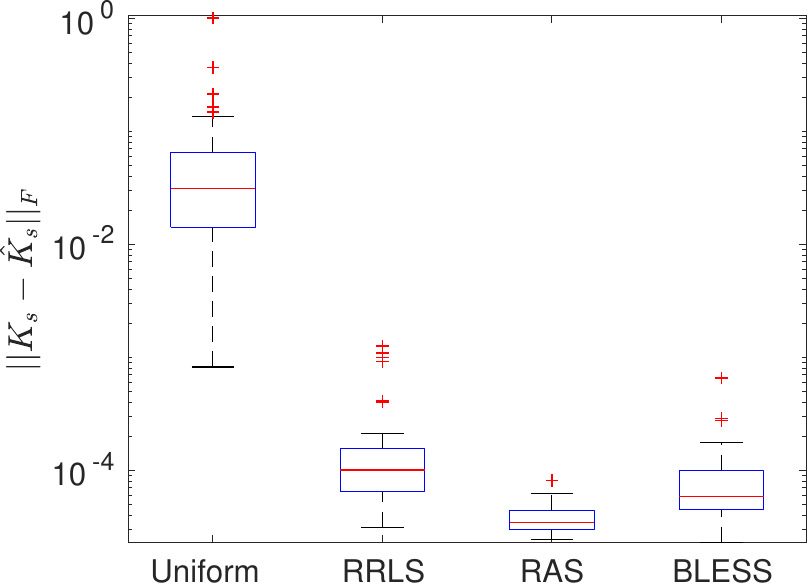}
		\caption{$|\mathcal{C}| = 924$, $\sigma = 10$.}
	\end{subfigure}
	\caption{Boxplot of the Nystr\"{o}m approximation error in Frobenius norm for \texttt{MiniBooNE}. The runtimes are: (a) $t_{RRLS} = 8$s, $t_{RAS} = 710$s, $t_{BLESS} = 18$s and (b) $t_{RRLS} = 3$s, $t_{RAS} = 167$s, $t_{BLESS} = 5$s.\label{fig:MiniBoon}}
\end{figure}
An overview of the data sets and parameters used in the experiments\footnote{\url{https://archive.ics.uci.edu/ml/data sets.php}} is given in Table \ref{Table:dataLarge}. Note that here we take a fixed $\sigma$ and $\gamma$ and not repeat the experiment for multiple subsets sizes $|\mathcal{C}|$. 

\begin{table}[h]
\caption{Medium scale data sets and parameters\label{Table:dataLarge}}
    
\begin{center}
 
                \begin{tabular}{rccccc}
                        \toprule
                        data set & $n$ & $d$ & $c/\epsilon$ & $\sigma$ & $\gamma$ \\ \midrule
                        \texttt{Super} & 21263 & 81 & 1 & 3 & $10^{-4}$ \\
                        \texttt{CASP} & 45730 & 9 & 1 & 1 & $10^{-4}$ \\
                        \texttt{Adult} & 48842 & 14 & 200 & 10 & $10^{-7}$ \\
                        \texttt{MiniBooNE} & 130065 & 50 & 200 & 5,10 & $10^{-7}$ \\
                        \texttt{cod-RNA} & 331152 & 8 & 200 & 4 & $10^{-6}$ \\
                        \texttt{Covertype} & 581012 & 54 & 200 & 10 & $10^{-6}$ \\
\bottomrule 
                \end{tabular}
\end{center}
\end{table}

\begin{figure}[h]
	\centering
	\begin{subfigure}[b]{0.32\textwidth}	\includegraphics[width=\textwidth]{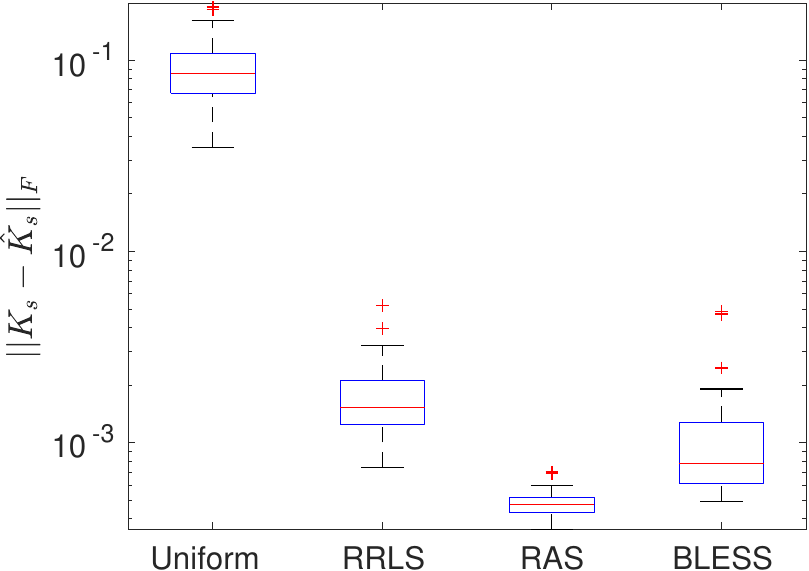}
		\caption{\texttt{Adult}, $|\mathcal{C}| =1816$.}
	\end{subfigure}
	\hfill
	\begin{subfigure}[b]{0.32\textwidth}
	\includegraphics[width=0.93\textwidth]{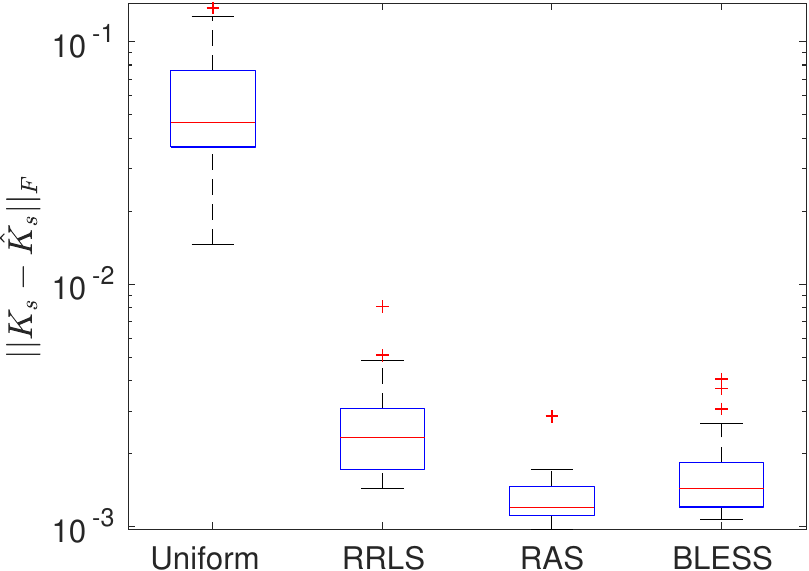}
		\caption{\texttt{cod-RNA}, $|\mathcal{C}| =1280$.}
	\end{subfigure}
	\begin{subfigure}[b]{0.32\textwidth}
	\includegraphics[width=\textwidth]{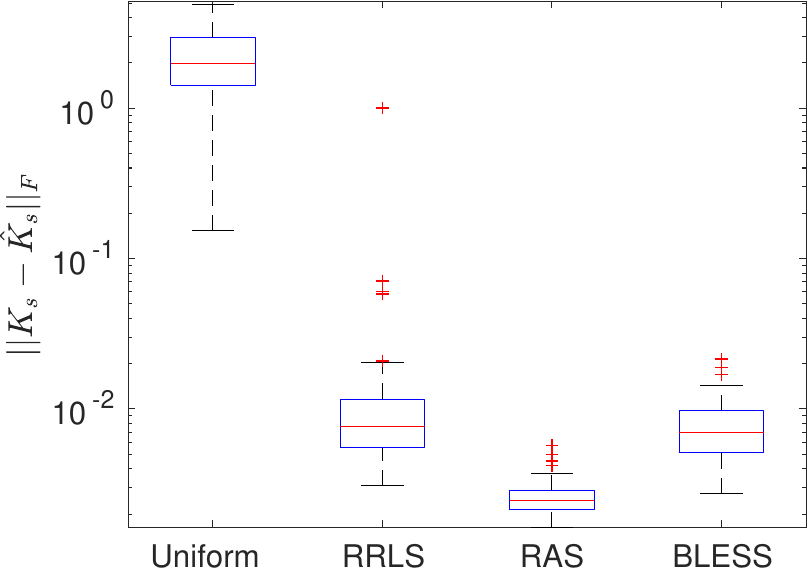}
		\caption{\texttt{Covertype}, $|\mathcal{C}| =3016$.}
	\end{subfigure}
	\caption{Boxplot of the Nystr\"{o}m approximation error in Frobenius norm.  The runtimes are: (a) $t_{RRLS} = 3$s, $t_{RAS} = 197$s, $t_{BLESS} = 4$s, (b) $t_{RRLS} = 11$s, $t_{RAS} = 555$s, $t_{BLESS} = 11$s and (c) $t_{RRLS} = 70$s, $t_{RAS} = 3131$s, $t_{BLESS} = 71$s.\label{fig:BigData}}
\end{figure}

The experiments are repeated 3 times, except for \texttt{Covertype}, for which the computation is done only once. The sampling with lowest average error over the $50$ subsets is shown for each method. Hence, the error bars intuitively indicate the variance of the stochastic estimate of the Frobenius error.
In Figure~\ref{fig:MiniBoon}, the sampling of landmarks with RAS for two different kernel parameters is compared to RRLS and BLESS on \texttt{MiniBooNE}. The boxplots show a reduction of the error, as well as a lower variance to the advantage of RAS. The increase in performance becomes more apparent when fewer landmarks are sampled. Additional experiments are given in Figure~\ref{fig:BigData} for \texttt{Adult}, \texttt{cod-RNA} and  \texttt{Covertype}. The indicative runtimes for RRLS, RAS and BLESS are only given for completeness.

\subsection{Performance for Kernel Ridge Regression}
\begin{figure}[t]
	\centering
	\begin{subfigure}[b]{0.32\textwidth}	\includegraphics[width=\textwidth]{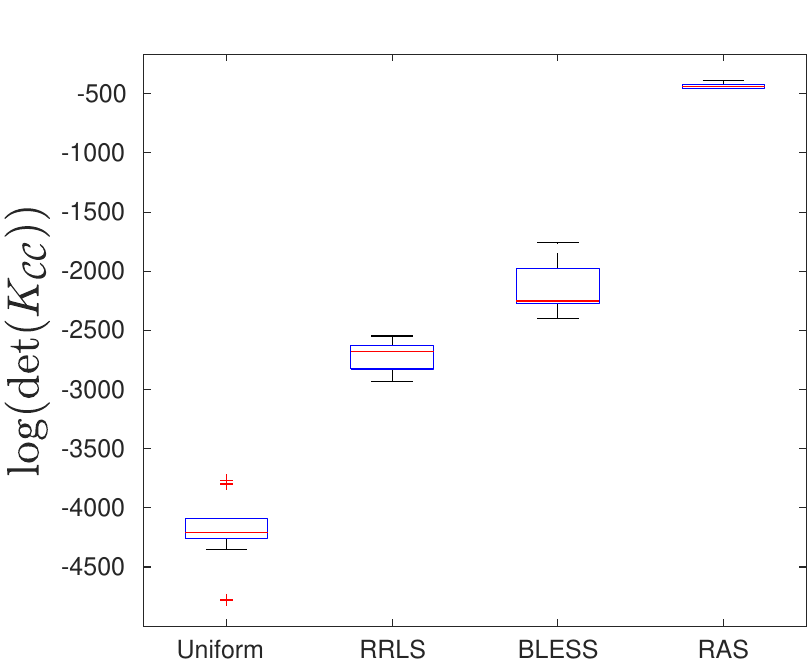}
		\caption{\texttt{Super}: $\mathrm{log}\mathrm{det}(K_{\mathcal{C}\mathcal{C}})$}
	\end{subfigure}
	\hfill
	\begin{subfigure}[b]{0.32\textwidth}
		\includegraphics[width=0.93\textwidth]{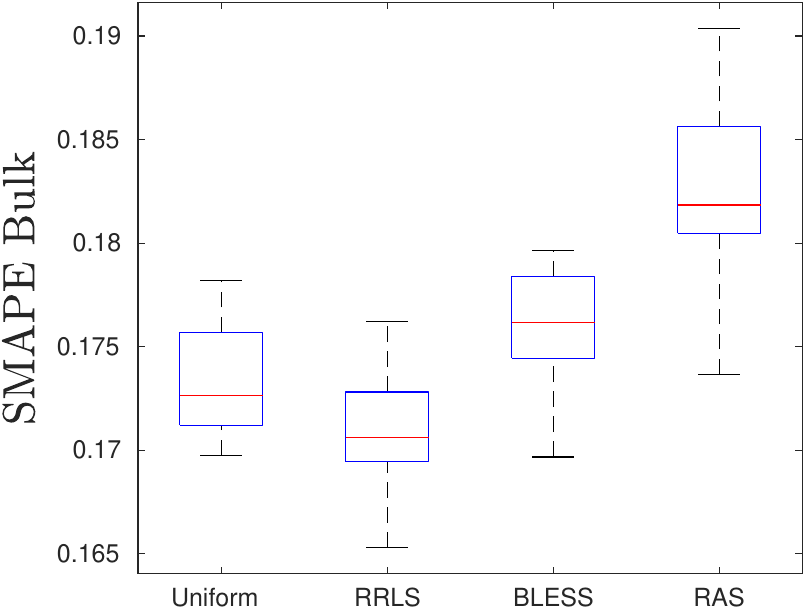}
		\caption{\texttt{Super}: SMAPE Bulk.}
	\end{subfigure}
	\begin{subfigure}[b]{0.32\textwidth}
		\includegraphics[width=\textwidth]{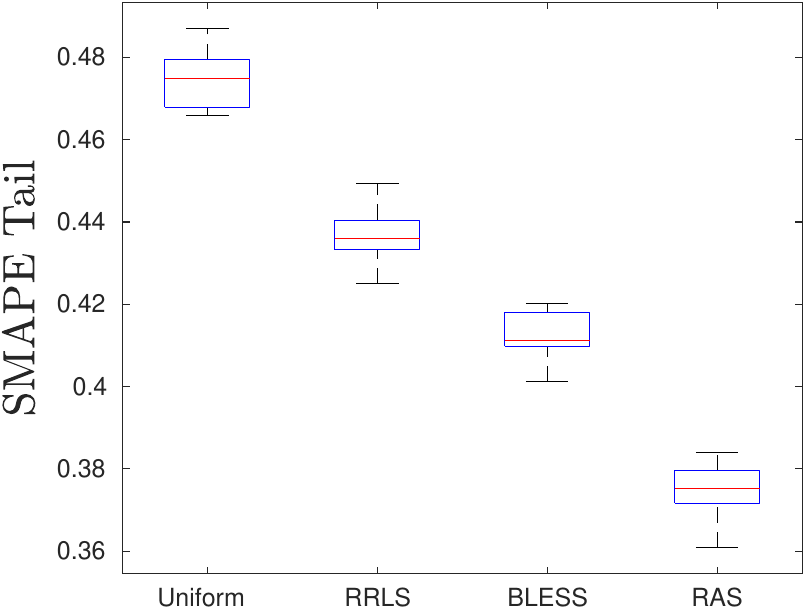}
		\caption{\texttt{Super}: SMAPE Tail.}
	\end{subfigure}
	\begin{subfigure}[b]{0.32\textwidth}	\includegraphics[width=\textwidth]{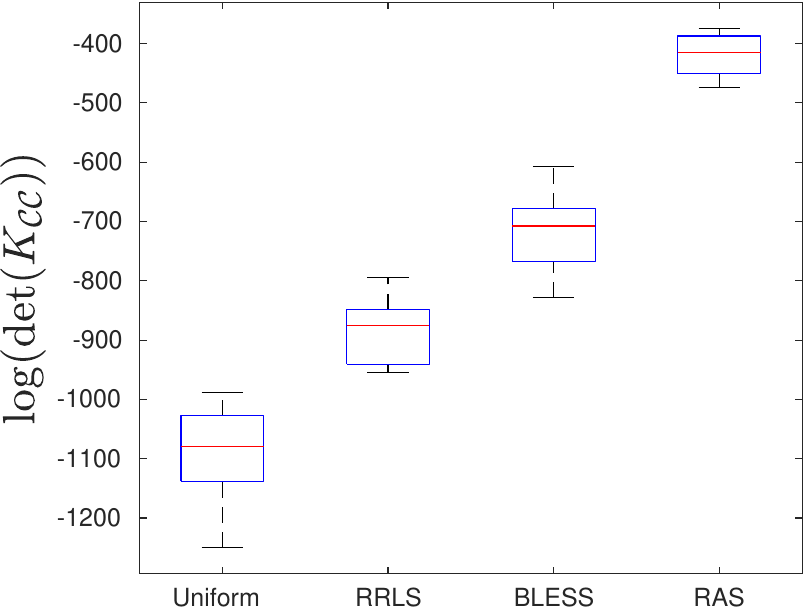}
		\caption{\texttt{CASP}: $\mathrm{log}\mathrm{det}(K_{\mathcal{C}\mathcal{C}})$}
	\end{subfigure}
	\hfill
	\begin{subfigure}[b]{0.32\textwidth}
		\includegraphics[width=0.93\textwidth]{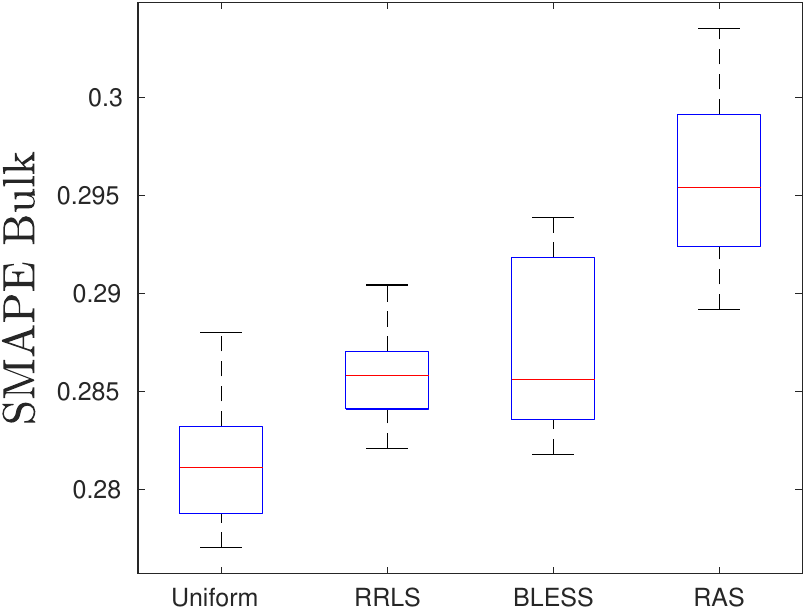}
		\caption{\texttt{CASP}: SMAPE Bulk.}
	\end{subfigure}
	\begin{subfigure}[b]{0.32\textwidth}
		\includegraphics[width=\textwidth]{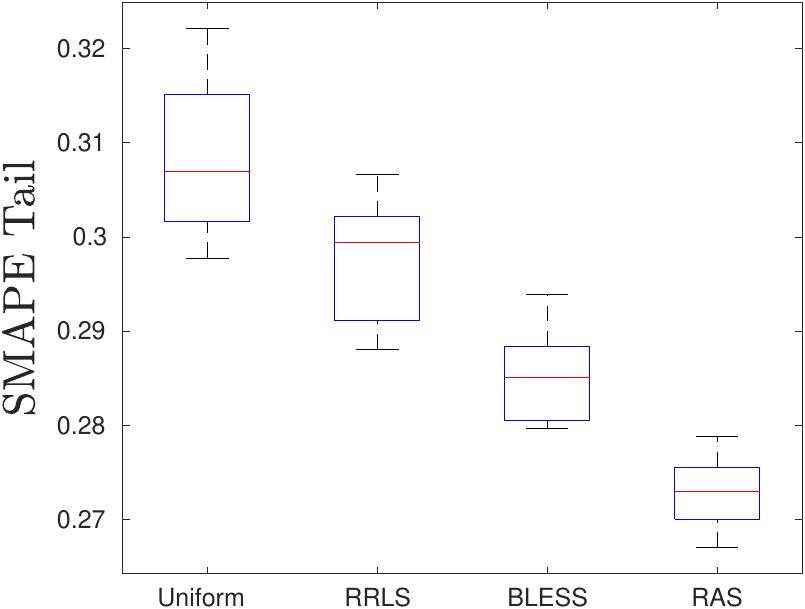}
		\caption{\texttt{CASP}: SMAPE Tail.}
	\end{subfigure}
	\caption{Boxplot of the $\mathrm{log}\mathrm{det}(K_{\mathcal{C}\mathcal{C}})$, SMAPE in the bulk and tail.  The runtimes are: (\texttt{CASP}) $t_{RRLS} = 0.4$s, $t_{RAS} = 55$s, $t_{BLESS} = 19$s, (\texttt{Super}) $t_{RRLS} = 0.6$s, $t_{RAS} = 68$s, $t_{BLESS} = 14$s.}
	\label{fig:Regression}
\end{figure}
To conclude, we verify the performance of the proposed method on a supervised learning task. Consider the input-output pairs $\{(x_i,y_i)\in \mathbb{R}^d \times \mathbb{R}\}_{i\in [n]}$. The Kernel Ridge Regression (KRR) using the subsample $\mathcal{C}$ is determined by solving
\begin{equation*}
\label{eq:ridgeRegression}
f^\star=  \arg\min_{f\in \mathcal{H}_{\mathcal{C}}}\frac{1}{n}\sum_{i=1}^{n}(y_i-f(x_i))^2 + \lambda \|f\|_{\mathcal{H}}^2, \text{ with } \lambda>0,
\end{equation*}
where $\mathcal{H}_{\mathcal{C}} = {\rm span} \{ k(x_i, \cdot)|i\in \mathcal{C}\}$. The regressor is $f^\star (\cdot) = \sum_{i\in \mathcal{C}}\alpha^\star_i k(x_i,\cdot)$  with  
\begin{equation}
\alpha^\star = (K_{\mathcal{C}}^\top K_{\mathcal{C}} + n \lambda K_{\mathcal{C}\mathcal{C}})^{-1}K_{\mathcal{C}}^\top  y.
\end{equation}
The experiment is done on the following data sets: \texttt{Superconductivity} (\texttt{Super}) and \texttt{Physicochemical Properties} \texttt{of} \texttt{Protein Tertiary} \texttt{Structure} (\texttt{CASP})\footnote{\url{https://archive.ics.uci.edu/ml/data sets.php}}, where we use a Gaussian Kernel after standardizing data. The symmetric mean absolute percentage error (SMAPE) between the outputs  and the regressor 
$$
\text{SMAPE} = \frac{1}{n} \sum_{i=1}^n \frac{|y_i - f^\star(x_i)|}{(|y_i| + |f^\star(x_i)|)/2},
$$
is calculated for the methods: Uniform sampling, RRLS, BLESS and approximate RAS. First, approximate RAS is run with $n_F = 4000$ random Fourier features. Afterwards Uniform, RRLS and BLESS sample the same number of landmarks as approximate RAS. To evaluate the performance of different methods on the full space of data sets, the test set is stratified, i.e., the test set is divided into `bulk' and `tail' as follows. The bulk corresponds to test points where the RLSs are smaller than or equal to the 70\% quantile, while the tail of data corresponds to test points where the ridge leverage score is larger than the 70\% quantile. The RLSs are approximated using RRLS with 8000 data points. This way, we can evaluate if the performance of the different methods is also good for `outlying' points. A similar approach was conducted in~\cite{Fanuel2020DiversitySI}. The data set is split in $50\%$ training data and $50\%$ test data, so to make sure the train and test set have similar RLS distributions. Information on the data sets and hyperparameters used for the experiments is given in Table~\ref{Table:dataLarge}. The regularization parameter $\lambda \in \{10^{-4},10^{-6},10^{-8},10^{-12} \}$ of Kernel Ridge Regression is determined by using 10-fold-cross-validation. The results in Figure~\ref{fig:Regression} show that approximate RAS samples a more diverse subset. As a result, the method performs much better in the tail of the data, while yielding a slightly worse performance in the bulk. The performance gain in the tail of the data is more apparent for the \texttt{Super}, where the RLS distribution has a longer tail. The data set has more `outliers', increasing the importance of using a diverse sampling algorithm. The RLS distributions of the data sets are visualized in Appendix.

\section{Conclusion}
Motivated by the connection between leverage scores, DPPs and kernelized Christoffel functions, we propose two sampling algorithms: DAS and RAS, with a theoretical analysis. RAS allows to sample diverse and important landmarks with a good overall performance. An additional approximation is proposed so that RAS can be applied to large data sets in the case of a Gaussian kernel. Experiments for kernel ridge regression show that the proposed method performs much better in the tail of the data, while maintaining a comparable performance for the bulk data.

\subsection*{Acknowledgements}
We thank Alessandro Rudi and Luigi Carratino for providing us with BLESS code. Most of this work was done when MF was at KU Leuven. EU: The research leading to these results has received funding from the European Research Council under the European Union's Horizon 2020 research and innovation program / ERC Advanced Grant E-DUALITY (787960). This paper reflects only the authors' views and the Union is not liable for any use that may be made of the contained information. Research Council KUL: Optimization frameworks for deep kernel machines C14/18/068 Flemish Government: FWO: projects: GOA4917N (Deep Restricted Kernel Machines: Methods and Foundations), PhD/Postdoc grant This research received funding from the Flemish Government (AI Research Program). Johan Suykens and Joachim Schreurs are affiliated to Leuven.AI - KU Leuven institute for AI, B-3000, Leuven, Belgium. Ford KU Leuven Research Alliance Project KUL0076 (Stability analysis and performance improvement of deep reinforcement learning algorithms)

\appendix

\section{Christoffel function and orthogonal polynomials \label{app:Christoffel}}
For completeness, we simply recall here the definition of the Christoffel function~\citep{Pauwels}. Let $\ell \in \mathbb{N}$ and $p(x)$ be an integrable real function on $\mathbb{R}^d$. Then, the Christoffel function reads:
\[
z\mapsto \min_{P\in\mathbb{R}_\ell (X)}\int P^2(x) p(x)\rmd x \text{ s.t. } P(z)=1,
\]
where $\mathbb{R}_\ell (X)$ is the set of $d$ variate polynomials of degree at most $\ell$.

In comparison, the optimization problem~\eqref{eq:Christoffel} involves a minimization over \rev{an} RKHS and includes an additional regularization term. Let $f\in \mathcal{H}$ and $\alpha\in\mathbb{R}^{n}$. Then, denote the sampling operator $\mathfrak{S}:\mathcal{H}\to \mathbb{R}^n$ , such that $\mathfrak{S}f = \frac{1}{\sqrt{n}}(f(x_1) \dots f(x_n))^\top$ and its adjoint $\mathfrak{S}^*:\mathbb{R}^n\to\mathcal{H}$ given by $\mathfrak{S}^* \alpha = \frac{1}{\sqrt{n}}\sum_{i=1}^n \alpha_i k_{x_i}$.
In order to rephrase the optimization problem in the RKHS, we introduce the covariance operator $\mathfrak{S}^*\mathfrak{S} =\frac{1}{n} \sum_{i=1}^n k_{x_i}\otimes k_{x_i}$ and the kernel matrix $\mathfrak{S}\mathfrak{S}^* = \frac{1}{n}K $ which share the same non-zero eigenvalues.
Then, the problem~\eqref{eq:addConstraints} reads:
\begin{equation}
C_\mathcal{C}(x_z) = \inf_{f\in\mathcal{H}}\langle f, (\mathfrak{S}^*\mathfrak{S}+ \gamma \mathbb{I}) f\rangle_\mathcal{H} \text{ s.t. } \langle k_{x_z}, f\rangle_\mathcal{H} = 1 \text{ and } f\in ({\rm span} \{ k_{x_s} | s\in \mathcal{C}\})^\perp.\label{eq:addConstraints2}
\end{equation}
 In other words, the formulation~\eqref{eq:addConstraints2} corresponds to the definition~\eqref{eq:Christoffel}  where the RKHS $\mathcal{H}$ is replaced by a specific subspace of $\mathcal{H}$.

\section{Useful results and Deferred Proofs}
In Section~\ref{sec:TechnicalResults}, some technical lemmata are gathered, while the next subsections provide the proofs of the results of this paper.
\subsection{Useful results \label{sec:TechnicalResults}}
We firstly give in Lemma~\ref{Lem:project} a result which is used to compute the expression of conditioned Christoffel functions in Proposition~\ref{prop:obj}.
\begin{lemma}\label{Lem:project}
Let $H$ be a Hilbert space. Let $\{v_s\in H| s\in \mathcal{C}\subseteq [n]\}$ a collection of vectors and denote $V = {\rm span}\{v_s\in H| s\in \mathcal{C}\subseteq [n]\}$. Define $\pi_{V^\perp}$ the linear projector onto the orthogonal of $V$. Let $u\in H$ such that $\pi_{V^\perp} u\neq 0$.
Then, the optimal value of
\begin{equation}
m^\star = \min_{x\in H} \langle x, x\rangle \text{ s.t. } \begin{cases}
\langle x, u\rangle = 1\\
\langle x, v_s\rangle = 0, \text{ for all } s\in \mathcal{C}
\end{cases}
\end{equation}
is  given by $m^\star = 1/\|\pi_{V^\perp} u\|^2$.
\end{lemma}
\begin{proof}
Let $x = x_\perp + x_\parallel$ where $x_\perp= \pi_{V^\perp} x$ and $x_\parallel = \pi_{V} x$. Then, the objective satisfies
$$
\langle x, x\rangle = \langle x_\perp, x_\perp\rangle + \langle x_\parallel, x_\parallel\rangle\geq \langle x_\perp, x_\perp\rangle.
$$
This indicates that we can simply optimize over $x\in V^\perp$. The problem is then
\begin{equation*}
m^\star = \min_{x\in V^\perp} \langle x, x\rangle \text{ s.t. } 
\langle x,  \pi_{V^\perp}u\rangle = 1.
\end{equation*}
By using a Lagrange multiplier $\lambda\in \mathbb{R}$, we find $2x + \lambda \pi_{V^\perp}u = 0$. Next, we use the constraint $\langle x,  \pi_{V^\perp}u\rangle = 1$. Thus, we find that the multiplier has to satisfy $\lambda = -2/ \|\pi_{V^\perp} u\|^2$. After a substitution, this yields finally $x^\star = \pi_{V^\perp} u/\|\pi_{V^\perp} u\|^2$.
\end{proof}
Next, we recall in Lemma~\ref{lem:push-through} an instrumental identity which is often called `push-through identity'.
\begin{lemma}[Push-through identity]\label{lem:push-through}
Let $X$ and $Y$ be $n\times m$ matrices such that $X^\top Y+\mathbb{I}_m$ and $YX^\top + \mathbb{I}_n$ are invertible. Then, we have $$(X^\top Y+\mathbb{I}_m)^{-1}X^\top = X^\top(YX^\top + \mathbb{I}_n)^{-1}.$$
\end{lemma}
The following identity is a slightly generalized version of a result given  in~\citet{MuscoMusco}, which originally considers only the diagonal elements of the matrix $K- L_{\mu, S}(K)$ below.
\begin{lemma}[Lemma~6 in  Appendix F of \citet{MuscoMusco}]\label{lem:WNys}
Let $\mu>0$ and $S$ be a sampling matrix. Let $B$ such that $K= B^\top B$. We have
$$
K- L_{\mu, S}(K) = \mu B^\top (BSS^\top B^\top + \mu \mathbb{I})^{-1} B.
$$
\end{lemma}
\begin{proof}
This is a consequence of the push-through identity in Lemma~\ref{lem:push-through}.
\end{proof}
In what follows, we list two results in Lemma~\ref{Lemma:B} and Lemma~\ref{Lemma:M} which are used in the proof of Proposition~\ref{Prop:Weighted}.
\begin{lemma}\label{Lemma:B}
Let $B$ be \rev{an} $n\times n$ matrix. Let $S$ be \rev{an} $n\times k$ matrix and $\alpha\in \mathbb{R}$, such that $\mathbb{I} + \alpha B^\top SS^\top B$ and $\mathbb{I} + \alpha S^\top B B^\top S$ are non-singular. Then, we have
\begin{align*}
B(\mathbb{I} + \alpha B^\top SS^\top B)^{-1}B^\top = BB^\top -\alpha BB^\top S (\mathbb{I} + \alpha S^\top B B^\top S)^{-1} S^\top BB^\top.
\end{align*}
\end{lemma}
\begin{proof}
We consider the expression on the RHS. It holds that
\begin{align*}
&BB^\top -\alpha BB^\top S (\mathbb{I} + \alpha S^\top B B^\top S)^{-1} S^\top BB^\top \\&= B(\mathbb{I} -\alpha B^\top S (\mathbb{I} + \alpha S^\top B B^\top S)^{-1} S^\top B)B^\top\\
&= B(\mathbb{I} -\alpha B^\top S S^\top B (\mathbb{I} + \alpha  B^\top SS^\top B)^{-1} )B^\top = B(\mathbb{I} + \alpha B^\top SS^\top B)^{-1}B^\top,
\end{align*}
where the next to last equality  is due to the push-through identity in Lemma~\ref{lem:push-through} with $X = \alpha S^\top B$ and $Y = S^\top B$. This yields the desired result.
\end{proof}
\begin{lemma}\label{Lemma:M}
Let $K\succeq 0$ and let $M$ be $n\times n$ symmetric matrix such that $K+M \succ 0$. Define $P_M = K^{1/2}\left(K + M\right)^{-1}K^{1/2}$. Let $S$ be \rev{an} $n\times k$ matrix and $\alpha\in \mathbb{R}$ be such that $K + M +\alpha K^{1/2}SS^\top K^{1/2}$  and $\mathbb{I}+\alpha S^\top P_M S$ are non-singular. Then, we have
\[
K^{1/2}\left(K + M +\alpha K^{1/2}SS^\top K^{1/2}\right)^{-1}K^{1/2}
= P_M- \alpha P_MS\left(\mathbb{I}+\alpha S^\top P_M S\right)^{-1}S^\top P_M.
\]
\end{lemma}
\begin{proof}
The results follows from Lemma~\ref{Lemma:B} by choosing $B = K^{1/2}(K+M)^{-1/2}$ such that $BB^\top = P_M$.
\end{proof}

\subsection{Proof of Proposition~\ref{prop:obj}}
\begin{proof}
We start by setting up some notations. Let $f\in \mathcal{H}$ and $\alpha\in\mathbb{R}^{n}$. Then, define the sampling operator $\mathfrak{S}:\mathcal{H}\to \mathbb{R}^n$ , such that $\mathfrak{S}f = \frac{1}{\sqrt{n}}(f(x_1) \dots f(x_n))^\top$. Its adjoint writes $\mathfrak{S}^*:\mathbb{R}^n\to\mathcal{H}$ and is given by $\mathfrak{S}^* \alpha = \frac{1}{\sqrt{n}}\sum_{i=1}^n \alpha_i k_{x_i}$.

Firstly, since $z\in X$, the objective~\eqref{eq:addConstraints} and the constraints depend only on the discrete set $X = \{x_1, \dots, x_n\}$. Then, we can decompose $f = f_X + f_X^\perp$ where $\langle f_X, f_X^\perp\rangle_\mathcal{H} = 0$ and $f_X \in {\rm span}\{k_{x_{i}} | i\in [n]\}$. Since, by construction, $f_X^\perp(x_i) = 0$ for all $i\in [n]$, the objective satisfies:
\[
\frac{1}{n}\sum_{i=1}^{n} f_X(x_i)^2 + \gamma \|f\|_\mathcal{H}^2\geq \frac{1}{n}\sum_{i=1}^{n} f_X(x_i)^2 + \gamma \|f_X\|_\mathcal{H}^2.
\]
Hence, we can rather minimize on $f\in {\rm span}\{k_{x_{i}} | i\in [n]\}$. This means that we can find $\alpha\in \mathbb{R}^n$ such that $f^\star = \sum_{i=1}^n\alpha_i k_{x_i} = \sqrt{n}\mathfrak{S}^*\alpha$. The objective~\eqref{eq:addConstraints} reads
\begin{align*}
\langle f^\star, (\mathfrak{S}^*\mathfrak{S}+ \gamma \mathbb{I}) f^\star\rangle_\mathcal{H} &= \sqrt{n} \alpha^\top \mathfrak{S}(\mathfrak{S}^*\mathfrak{S}+ \gamma \mathbb{I})\mathfrak{S}^* \alpha \sqrt{n}\\
& = n\alpha^\top(\mathfrak{S}\mathfrak{S}^*+ \gamma \mathbb{I}) \mathfrak{S}\mathfrak{S}^* \alpha\\
& = n\alpha^\top(K/n+ \gamma \mathbb{I}) (K/n) \alpha\\
& = n^{-1}\alpha^\top(K+ n\gamma \mathbb{I}) K \alpha.
\end{align*}

For simplicity, denote by $K_{x_i} = K e_i$ a column of the matrix $K$, where $e_i$ is the $i$-th element of the canonical basis of $\mathbb{R}^n$.
 The problem becomes then:
\[
\min_{\alpha\in \mathbb{R}^n} \alpha^\top M \alpha \text{ such that } \alpha^\top K_{x_z} = 1 \text{ and } \alpha^\top K_{x_s} = 0 \text{ for all } s\in \mathcal{C},
\]
where $M =n^{-1} K(K+n\gamma \mathbb{I})$. Recall that, since we assumed $k\succ 0$, then $K$ is invertible. By doing the change of variables $\alpha_M = M^{1/2} \alpha$, we obtain: 
\[
\min_{\alpha_M\in \mathbb{R}^n} \alpha_M^\top  \alpha_M \text{ such that } \alpha_M^\top K^{(M)}_{x_z} = 1 \text{ and } \alpha_M^\top K^{(M)}_{x_s} = 0 \text{ for all } s\in \mathcal{C},
\]
with $K^{(M)}_{x_z}  = M^{-1/2}K_{x_z}$ and $K^{(M)}_{x_s} =  M^{-1/2}K_{x_s}$. Let $V = {\rm span} \{K^{(M)}_{x_s}| s\in \mathcal{C}\}$ and $K^{(M)}\in \mathbb{R}^{n\times |\mathcal{C}|}$ be the matrix whose columns are the vectors $K^{(M)}_{x_s}$.  Notice that the elements of $\{K^{(M)}_{x_s}| s\in \mathcal{C}\}$ are linearly independent since $K^{(M)}_{x_s} =  M^{-1/2}K e_s$. Then, we use Lemma~\ref{Lem:project}, and find $C^{-1}_{\mathcal{C}}(x_z) = \|\pi_{V^\perp}K^{(M)}_{x_z} \|^2$. By using $\pi_{V} = K^{(M)} G^{-1} K^{(M)\top}$ with the Gram matrix $G = K^{(M)\top}K^{(M)}$, we find:
\begin{equation}
\|\pi_{V^\perp}K^{(M)}_{x_z} \|^2 = K^{(M)\top}_{x_z} \left(\mathbb{I} - K^{(M)} G^{-1} K^{(M)\top}\right) K^{(M)}_{x_z}.\label{eq:diffkernel}
\end{equation}
By substituting $M = n^{-1} K(K+n\gamma \mathbb{I})$, we find that $G_{ss'} =n e^\top_{s}K(K+n\gamma \mathbb{I})^{-1} e_{s'}$ for all $s,s'\in \mathcal{C}$. Hence, we find:
\[
\|\pi_{V^\perp}K^{(M)}_{x_z} \|^2 =n\left\{ e_z^\top K(K+n\gamma \mathbb{I})^{-1} e_z - e_z^\top K(K+n\gamma \mathbb{I})^{-1} G^{-1} K(K+n\gamma \mathbb{I})^{-1}e_z\right\},
\]
which yields the desired result. 
\end{proof}
\subsection{Details of derivations of Section~\ref{sec:KernChristoffel}}
By using a similar approach and the same notations as in the previous subsection, we find
$$
C_\eta(x_z) = \min_{\alpha\in \mathbb{R}^n}\alpha^\top \left(\frac{1}{n} K\Diag(\eta)K +\gamma K\right) \alpha \text{ such that } \alpha^\top K_{x_z} = 1,
$$
for some $z\in [n]$. By using Lemma~\ref{Lem:project} we find the expression given  in~\eqref{eq:PauwelsChristoffel}
$$
C_\eta(x_z) =\left(K_{x_z}^\top \left( \frac{1}{n} K\Diag(\eta)K + \gamma K\right)^{-1}K_{x_z} \right)^{-1} = \frac{n^{-1}}{[K \left( K\Diag(\eta)K + n\gamma K\right)^{-1}K]_{zz}}.
$$
The second expression for Christoffel functions given in~\eqref{eq:PauwelsChristoffel2} is merely obtained by using the matrix inversion lemma as follows
\begin{align*}
K \left( K\Diag(\eta)K + n\gamma K\right)^{-1}K &= K\left(\frac{K^{-1}}{n\gamma} - \left(\Diag(\eta)^{-1} + \frac{K}{n\gamma}\right)^{-1} \right) K\\
&=\frac{1}{n\gamma}\left( K - K\left( K+n\gamma \Diag(\eta)\right)^{-1}\right)\\
&= K\left(K+n\gamma\Diag(\eta)^{-1}\right)^{-1} \Diag(\eta)^{-1}.
\end{align*}
By taking the $(z,z)$-th entry of the matrix above  gives directly~\eqref{eq:PauwelsChristoffel2}.
\subsection{Proof of Proposition~\ref{prop:det}}

\begin{proof}
The objective $C_\mathcal{C}(x_z)^{-1}$ can be simply obtained by calculating the following determinant
\begin{align*}
n^{-1}C_\mathcal{C}(x_z)^{-1} &= \det\left(
\begin{pmatrix}
P_{zz} & P_{z,\mathcal{C}}\\
P_{z,\mathcal{C}}^\top & P_{\mathcal{C},\mathcal{C}}
\end{pmatrix} \cdot 
\begin{pmatrix}
1 & 0\\
0^\top & P_{\mathcal{C},\mathcal{C}}
\end{pmatrix}^{-1}\right) \\
&= \det
\begin{pmatrix}
P_{zz} & P_{z,\mathcal{C}}P_{\mathcal{C},\mathcal{C}}^{-1}\\
P_{z,\mathcal{C}}^\top & \mathbb{I}
\end{pmatrix} = P_{zz}-P_{z,\mathcal{C}}P_{\mathcal{C},\mathcal{C}}^{-1}P_{z,\mathcal{C}}^\top,
\end{align*}
where we use the well-known formula for the determinant of block matrices at the last equality. This yields the desired result.
\end{proof}

\subsection{Proof of Proposition~\ref{Prop:Weighted}}
\begin{proof}
By using Lemma~\ref{Lem:project} or the results of~\citet{Pauwels}, we find the expression
\begin{equation}
n^{-1}(e_i^\top \underbrace{K(K\Diag(\eta)K + n\gamma K)^{-1}K}_{T}e_i)^{-1} = \inf_{f(x_z) =1}\frac{1}{n}\sum_{i=1}^{n} \eta_i f(x_i)^2 + \gamma \|f\|_\mathcal{H}^2,\label{eq:eq1Prop4}
\end{equation}
where the weights are defined as $\eta_i  = 1+\mu^{-1}$ for all $i\in  \mathcal{C}$ and $\eta_i  = \nu$ if $i\in \tilde{\mathcal{C}}$ and $\eta_i  = 1$ otherwise. Recall that $\mu>0$ and $0< \nu < 1$ so that $\Diag(\eta)$ is the positive diagonal matrix
\[
\Diag(\eta) = \mathbb{I} +\mu^{-1}CC^\top + (\nu - 1)\tilde{C}\tilde{C}^\top,
\]
where $C$ (resp. $\tilde{C}$) is a sampling matrix obtained by sampling the columns of the identity matrix belonging to $\mathcal{C}$ (resp. $\tilde{\mathcal{C}}$). Notice that $CC^\top$ (resp. $\tilde{C}\tilde{C}^\top$) is the diagonal matrices with entries equal to $1$ for elements indexed by $\mathcal{C}$ (resp. $\tilde{\mathcal{C}}$) and equal to zero otherwise. First, we establish an equivalent expression for $T$ at the LHS of~\eqref{eq:eq1Prop4}, as follows
\begin{align*}
T &= K(K\Diag(\eta)K + n\gamma K)^{-1}K\\
& = K^{1/2}\left(K^{1/2}\left(\mathbb{I}+\mu^{-1}CC^\top + (\nu - 1)\tilde{C}\tilde{C}^\top\right) K^{1/2} +n\gamma \mathbb{I}\right)^{-1}K^{1/2}.
\end{align*}
Then, the idea is to define the symmetric matrix $M = (\nu-1)K^{1/2}\tilde{C}\tilde{C}^\top K^{1/2}+n\gamma \mathbb{I}$, which is such that
\begin{equation}
K + M =  K^{1/2}(\mathbb{I}-\tilde{C}\tilde{C}^\top) K^{1/2} + \nu K^{1/2}\tilde{C}\tilde{C}^\top K^{1/2}+n\gamma \mathbb{I}\succ 0,\label{eq:K+M}
\end{equation}
since $\mathbb{I}-\tilde{C}\tilde{C}^\top\succeq 0$.
Then, define the following notation (also used in~Lemma~\ref{Lemma:M}):
\[
P_M = K^{1/2}(K+M)^{-1} K^{1/2} = K^{1/2}\left(K+n\gamma \mathbb{I} -  (1-\nu)K^{1/2}\tilde{C}\tilde{C}^\top K^{1/2} \right)^{-1}K^{1/2},\]
which satisfies $P_M \succ 0$ since $K+M\succ 0$ and $K\succ 0$.
Next, by using Lemma~\ref{Lemma:M} with $\alpha = \mu^{-1}$ and the matrix $M$ defined above~\eqref{eq:K+M}, we find the following equivalent expression for $T$ defined in~\eqref{eq:eq1Prop4}, i.e.,
\begin{align*}
 T &=  K^{1/2}\left(K+  \underbrace{(\nu - 1)K^{1/2}\tilde{C}\tilde{C}^\top K^{1/2}+n\gamma \mathbb{I}}_M+ \mu^{-1}K^{1/2}CC^\top K^{1/2}\right)^{-1}K^{1/2}\\
 &= P_M- \mu^{-1} P_MC\left(\mathbb{I}+\mu^{-1} C^\top P_M C\right)^{-1}C^\top P_M.
\end{align*}
Now, in order to match the notation of Proposition~\ref{Prop:Weighted}, we define $H = P_M$. This gives
\begin{align}
T = H-  \mu^{-1}H C\left(\mathbb{I} + \mu^{-1} C^\top H C\right)^{-1}C^\top H 
 = H - L_{\mu, C}(H).\label{eq:T_prop_weighted}
\end{align}
 Next, to find an equivalent expression for \[H = K^{1/2}\left(K+n\gamma \mathbb{I} +  (\nu -1)K^{1/2}\tilde{C}\tilde{C}^\top K^{1/2} \right)^{-1}K^{1/2},\] we use once more Lemma~\ref{Lemma:M}, but this time we take $\alpha = \nu-1$ and $M = n\gamma \mathbb{I}$. Notice that $\alpha$ does not need to be positive to be able to apply Lemma~\ref{Lemma:M}. An equivalent expression for $H$ is
\begin{equation}
H = P + (1-\nu) P\tilde{C}\left(\mathbb{I}-(1-\nu)\tilde{C}^\top P \tilde{C}\right)^{-1}\tilde{C}^\top P,\label{eq:final_result}
\end{equation}
with $P = K(K+n\gamma \mathbb{I})^{-1}$. Remark that the assumptions of Lemma~\ref{Lemma:M} are satisfied: on the one hand, the matrix inverse in~\eqref{eq:final_result} exits since $0\preceq \tilde{C}^\top P \tilde{C}\preceq \mathbb{I}$ and $0<\nu < 1$; on the other hand, the matrix $K+n\gamma \mathbb{I} +  (\nu -1)K^{1/2}\tilde{C}\tilde{C}^\top K^{1/2}$ is non-singular in the light of~\eqref{eq:K+M}. By combining \eqref{eq:T_prop_weighted} and~\eqref{eq:final_result}, we have the desired result.
\end{proof}

\section{Proof of the results of Section \ref{sec:DAS}}
\subsection{Proof of Proposition~\ref{Corol:conv}}
\begin{proof}
Let us explain how Algorithm~\ref{Alg1} can be rephrased  by  considering first Proposition~\ref{prop:det}. First, we recall the spectral factorization of the projector kernel matrix $P = B^\top B$. For simplicity, denote the Hilbert space\footnote{Here, we do not endow $H$ with \rev{an} RKHS structure.} $H = \mathbb{R}^n$. We also introduce the notation $\mathcal{F} = \{b_1,\dots , b_n\}$ for the set of columns of $B\in \mathbb{R}^{n\times n}$ which is considered as a compact subset of $H$. The vector subspace generated by the columns indexed by the subset $\mathcal{C}_m$ is denoted by
$V_m  = {\rm span}\{b_s | s\in \mathcal{C}_{m}\}\subset H.$
With these notations and thanks to Proposition~\ref{prop:det},  the \emph{square root} of the objective maximized by Algorithm~\ref{Alg1} can be written $\sigma_m = n^{-1/2}C^{-1/2}_{\mathcal{C}_m}(x_{s_{m+1}}) = \|b_{s_{m+1}}-\pi_{V_m}b_{s_{m+1}}\|_2$.
 Hence, by using the definition of the Christoffel function, we easily see that 
$\sigma_{m+1}\leq  \sigma_m.$
Then, the algorithm can be viewed as a greedy reduced subspace method, namely:
\[ \sigma_m =  \max_{b\in \mathcal{F}}{\rm dist}(b, V_m), \]
where the distance between a column $b\in \mathcal{F}$ and the subspace $V_m$ is given by:
\begin{equation}
{\rm dist}(b, V_m) = \min_{X\in V_m} \|b-X\|_2.
\end{equation}
As explained in~\citet{DeVore2013,SantinHaasdonk,GaoKovalskyDaubechies,Binev2011}, the convergence of this type of algorithm can be studied as a function of the Kolmogorov width:
\begin{equation}
    d_m = \min_{Y_m} \max_{b\in \mathcal{F}} {\rm dist}(b,Y_m),\label{eq:m-width}
\end{equation}
where the minimization is over all $m$-dimensional vector subspace $Y_m$ of $H$. Intuitively, $d_m$ is the best approximation error calculated over all possible subspace of dimension $m$ with $m<n$.
Then, Theorem 3.2 together with Corollary 3.3 in~\citet{DeVore2013} gives the upper bound:
\begin{equation}
\sigma_{2m} \leq \sqrt{2 \|P\|_\infty d_m},\quad  \text{ for all } m\geq 1,\label{eq:sigmad}
\end{equation}
which is a close analogue of eq. (4.9) in~\cite{GaoKovalskyDaubechies}.
Hence, by upper bounding the Kolmogorov width, we  provide a way to calculate a convergence rate of the greedy algorithm. To do so, we first recall that $\mathcal{F} = \{b_1,\dots b_n\}$ and where the $i$-th column of B is denoted by $b_i = B e_i$. Therefore, we have:
\begin{equation}
\max_{b\in \mathcal{F}} \min_{y\in Y_m} \|b-y\|_2 = \max_{i\in [N]} \min_{y\in Y_m} \|B e_i-y\|_2\leq  \max_{\|x\|_2\leq 1} \min_{y\in Y_m} \|Bx-y\|_2,\label{eq:bound_width}
\end{equation}
where the inequality is obtained by noticing that $e_i$ is such that $\|e_i\|_2 = 1$ and by observing that maximizing an objective over a larger set yields a larger objective value.
Then, in the light of this remark and according to~\citet{Pinkus1979}, we define a modified width as follows:
\[
\tilde{d}_m(B) = \min_{Y_m} \max_{\|x\|_2\leq 1} \min_{y\in Y_m} \|Bx-y\|_2,
\]
which naturally satisfies $d_m\leq \tilde{d}_m(B)$ as it can be seen by taking the minimum of both sides of~\eqref{eq:bound_width} over the $m$-dimensional vector spaces $Y_m$ and by recalling the Kolmogorov width definition in~\eqref{eq:m-width}. Then, $\tilde{d}_m(B)$ can be usefully upper bounded thanks to Theorem~\ref{thm:Pinkus} that we adapt from~\citet{Pinkus1979}.
\begin{theorem}[Thm 1.1 in \citet{Pinkus1979}]\label{thm:Pinkus}
Let $B$ be a real $p\times q$ matrix. Let $\lambda_1\geq \dots \geq \lambda_p\geq 0$ denote the $p$ eigenvalues of $BB^\top$ and let $v_1, \dots, v_p$ denote a corresponding set of orthonormal eigenvectors. Then, $\tilde{d}_m(B) =\lambda_{m+1}^{1/2}$ if  $ m<p$ and  $\tilde{d}_m(B) =0$ if $m\geq p$.
Furthermore if $m<p$, then an optimal subspace for $\tilde{d}_m$ is ${\rm span}\{v_1,\dots, v_m\}$.
\end{theorem}
Our convergence result given in Corollary~\ref{Corol:conv} simply follows from~\eqref{eq:sigmad}, the inequality  $d_m\leq \tilde{d}_m$ and Theorem~\ref{thm:Pinkus} by noting that the non-zero eigenvalues of $BB^\top$ are the non-zero eigenvalues of $B^\top B$. Notice that $B$ is obtained by the spectral factorization of $P = B^\top B$ and therefore $B$ is full rank.
\end{proof}
\subsection{Proof of Lemma \ref{lem:Kfrom|K}}
\begin{proof}
To prove this result, we rely on Lemma~\ref{lem:WNys} and on the definition of $P = K(K+n\gamma \mathbb{I})^{-1}$. Firstly, by following Lemma~\ref{lem:WNys}, we have
\begin{align}
K-L_{\mu,S}(K) = \mu K^{1/2} (K^{1/2} S S^\top K^{1/2}+ \mu \mathbb{I})^{-1} K^{1/2},\label{eq:eqLemma1}
\end{align}
with $\mu>0$.
Then, we can obtain a similar expression for the projector kernel
\begin{align*}
    P - L_{\mu,S}(P)  &=\mu P^{1/2} (P^{1/2} S S^\top P^{1/2}+ \mu \mathbb{I})^{-1} P^{1/2}\\
    &= \mu K^{1/2} \left(K^{1/2}SS^\top K^{1/2} + \mu (K+n\gamma \mathbb{I})\right)^{-1}K^{1/2},\label{eq:eqLemma1}
\end{align*}
where we substituted the definition $P^{1/2} = K^{1/2}(K+n\gamma\mathbb{I})^{-1/2}$ in the last equality. Next, we use $K\preceq\lambda_{max}(K)\mathbb{I}$, and the property that $A^{-1}\succeq B^{-1}$ if $B\succeq A$ with $A$ and $B\succ 0$. We obtain:
\[
 \big(K^{1/2} S S^\top K^{1/2}+ \mu  (K+n\gamma \mathbb{I}) \big)^{-1}   \succeq \big(K^{1/2} S S^\top K^{1/2}+ \mu (\lambda_{max}(K)+n\gamma) \mathbb{I}\big)^{-1},
\]
and the result follows from the following inequality
\begin{align*}
  P-L_{\mu,S}(P) &\succeq \mu K^{1/2}\big(K^{1/2} S S^\top K^{1/2}+ \mu (\lambda_{max}(K)+n\gamma) \mathbb{I}\big)^{-1} K^{1/2}\\
   & = (\lambda_{max}(K)+n\gamma)^{-1} \tilde{\mu} K^{1/2}\big(K^{1/2} S S^\top K^{1/2}+ \tilde{\mu} \mathbb{I}\big)^{-1} K^{1/2}\\
   & = (\lambda_{max}(K)+n\gamma)^{-1}    \left(K-L_{\tilde{\mu},S}(K) \right),
 \end{align*}
 where we used again~\eqref{eq:eqLemma1} in the last equality and defined $\tilde{\mu} = \mu(\lambda_{max}(K)+n\gamma)$.
\end{proof}
\section{Proof of the results of Section~\ref{sec:RAS}}
\subsection{Proof of Lemma~\ref{Lem:KernelApprox}}
The statement of Lemma~\ref{Lem:KernelApprox} has been adapted from~\cite{ElAlaouiMahoney} to the setting of this paper. Its proof, given for completeness, follows the same lines as in Lemma~1 in the appendix of~\cite{ElAlaouiMahoney}.
\begin{proof}
For simplicity let $\theta = \epsilon n\gamma /(1+\epsilon)$.
By assumption, we have
$$ \Psi \Psi^\top - \Psi S S^\top  \Psi^\top \preceq t \mathbb{I},$$
with $P_\epsilon(P_{n\gamma}(K)) = \Psi^\top \Psi$. By using the composition rule given in Lemma~\ref{lem:CompositionRule}, we find the equivalent expression
$$\Psi^\top \Psi= P_\epsilon(P_{n\gamma}(K)) = \frac{1}{1+\epsilon}P_{\theta}(K) \text{ with } \theta = \frac{\epsilon n\gamma}{1+\epsilon}.$$
Therefore, we can choose the factorization given by $\Psi = (1+\epsilon)^{-1/2}(K+\theta\mathbb{I})^{-1/2} K^{1/2} $. By using our starting hypothesis and by denoting for simplicity $P_\theta = P_\theta(K)$, we have 
$$
\frac{1}{1+\epsilon}\left( P_\theta - P_\theta^{1/2} S S^\top P_\theta^{1/2} \right)\preceq t\mathbb{I}.
$$
After a substitution of $P_\theta(K)= K(K+\theta\mathbb{I})^{-1}$, this gives the following inequality 
$$
(K+\theta\mathbb{I})^{-1/2}\left(K- K^{1/2}SS^\top K^{1/2}\right)(K+\theta\mathbb{I})^{-1/2}\preceq t(1+\epsilon) \mathbb{I}.
$$
Equivalently, we have the following inequality
$$
K- K^{1/2}SS^\top K^{1/2} \preceq t(1+\epsilon) (K+\theta\mathbb{I}),
$$
which gives
\begin{equation}
    K^{1/2}SS^\top K^{1/2}\succeq K-t(1+\epsilon)(K+\theta\mathbb{I}). \label{eq:ineqLemma3}
\end{equation}
Then, if $1-t(1+\epsilon)>0$,  it holds that
\begin{align*}
K-L_{\theta, S}(K) &= \theta K^{1/2}\big(K^{1/2}SS^\top K^{1/2}+\theta \mathbb{I}\big)^{-1}K^{1/2} &\text{(by Lemma~\ref{lem:WNys})}\\
& \preceq \theta K^{1/2}\big(K+\theta \mathbb{I}-t(1+\epsilon) (K+\theta \mathbb{I})\big)^{-1}K^{1/2} & \text{(by using~\eqref{eq:ineqLemma3})}\\
&= \frac{\theta}{1-t(1+\epsilon)} P_{\theta}(K), &
\end{align*}
where we used $P_\theta(K) = K(K+\theta \mathbb{I})^{-1}$ at the last equality.
Now, by using Lemma~\ref{lem:CompositionRule}, we find
\begin{align*}
K-L_{\frac{\epsilon n\gamma }{1+\epsilon}, S}(K)  \preceq  \frac{\epsilon n\gamma /(1+\epsilon)}{1-t(1+\epsilon)} (1+\epsilon) P_{\epsilon}(P_{n\gamma}(K)),
\end{align*}
which is the stated result.
\end{proof}

\subsection{Proof of Theorem~\ref{thm:Main}}
The proof structure follows closely~\citet{OnlineRowSampling} with suitable adaptations.
In particular and in contrast with~\citet{OnlineRowSampling}, a main ingredient of the proof of Theorem~\ref{thm:Main}
is the following improved Freedman inequality.
\begin{theorem}[Thm 3.2 in \citet{MINSKER}]\label{thm:Minsker}
Let $X_1, \dots, X_n$ be martingale differences valued in the $n\times n$ symmetric matrices \rev{and let $U>0$ be deterministic real number such that $\|X_i\|_{2}\leq U$ almost surely}. Denote the variance $W_n = \sum_{i=1}^{n} \mathbb{E}_{i-1}(X_i^2)$. Then, for any $t\geq \frac{1}{6}(U + \sqrt{U^2 + 36 \sigma^2})$, we have:
\begin{equation}
\Pr\left(\|\sum_{i=1}^{n} X_i\|_{2} >t , \lambda_{max}(W_n)\leq \sigma^2\right)\leq 50 \Tr\left(\min\{\frac{t}{U}\frac{\mathbb{E}(W_n)}{\sigma^2},1\}\right) \exp\left(\frac{-t^2/2}{\sigma^2+tU/3}\right).\label{eq:MinskerBound}
\end{equation}
\end{theorem}
Notice that the function $\min\{M,1\}$, defined for all symmetric $M\succeq 0$, acts on the eigenvalues of $M$. In Theorem~\ref{thm:Minsker}, $\sigma^2$ denotes an upper bound on the predictable quadratic variation and not the kernel bandwidth parameter.

We now give the proof of Theorem~\ref{thm:Main}.
\begin{proof}
Let $t>0$ and $0<\epsilon<1$. We consider here the sampling procedure given by RAS; see Algorithm~\ref{Alg2}. 
We recall that we defined the matrix $B$ such that $B^\top B = P_{n\gamma}(K)$. Then, this proof considers the problem of sampling columns of the matrix $\Psi$ which is defined  such that $$\Psi = (B B^\top+\epsilon \mathbb{I})^{-1/2}B.$$

\noindent\textbf{Proof strategy.} 
The key idea of this proof is to define a matrix martingale $Y_0, \dots, Y_n$ involving a sum of rank 1 matrices built from the columns of $\Psi$ so that, at step $n$, we have
$$
Y_n =  \Psi S_n S_n^\top \Psi^\top - \Psi \Psi^\top,
$$ 
where $S_n$ is the sampling matrix at step $n$. If RAS samples enough landmarks, i.e., if $S_n$ samples enough columns of $\Psi$, then $Y_n$ will be small with high probability.
In other words, we want to show that for a large enough $c>0$, we have $$ \| Y_n \|_2\leq t,$$ and therefore $\lambda_{\rm max}( Y_n ) \leq t$ with a large probability.
Then, by using Lemma~\ref{Lem:KernelApprox} and by choosing $t=1/2$, we can obtain an error bound on a regularized Nystr\"om approximation $$
0\preceq K-L_{\frac{\epsilon n\gamma}{1+\epsilon}, S}(K)\preceq \frac{2\epsilon n\gamma}{1-\epsilon} \mathbb{I},
$$
which is the result that we want to prove. 

The proof goes as follows. The matrix martingale is firstly defined. In order to use the martingale concentration inequality, we upper bound the norm of the difference sequence which boils down to find a lower bound on the sampling probability. Finally, we use the concentration inequality find a lower bound on $c$ so that $ \| Y_n \|_2\leq t$ with a large probability.\\

\noindent\textbf{Defining a matrix martingale.} Let $S_i$ be the sampling matrix at step $i$ in Algorithm~\ref{Alg2} and $p_i$ be the sampling probability. 
Define now a matrix martingale $Y_0, Y_1, \dots, Y_n \in \mathbb{R}^{n\times n}$ such that $Y_0 = 0$  and with difference sequence $X_1, \dots, X_n$ given by:
\begin{align}
    X_i = \begin{cases}
(\frac{1}{p_i}-1) \psi_i \psi_i^\top \text{ if column } b_i \text{ is sampled,}\\
-\psi_i \psi_i^\top  \text{ otherwise,}
\end{cases}\label{eq:X_i}
\end{align}
where $p_i$ is given in~\eqref{eq:tildeli} and where $\psi_i$ is the $i$-th column of $\Psi$. \rev{So, we have $X_i = Y_i -Y_{i-1}$ for $1\leq i\leq n$. Notice also that if $p_i = 1$, then $X_i=0$.} It remains to show that we indeed defined a martingale. \rev{To do so, we define $\mathbb{E}_{i-1}$ as the expectation conditioned on the previous steps from $0$ to $i-1$ included, more precisely $$\mathbb{E}_{i-1}[Z] = \mathbb{E}[Z|Y_0, X_1, \dots, X_{i-1}],$$ for some random $Z$.} We simply have $\mathbb{E}_{i-1}(X_i) =(p_i(\frac{1}{p_i}-1) - (1-p_i))\psi_i \psi_i^\top =0$ by a direct calculation. 

Next, denote by $B_{[i-1]}$ the rectangular matrix obtained by selecting the $i-1$ first columns of $B$, so that $B_{[n]} = B$.  So, the martingale value at step $i-1$ reads
\[
Y_{i-1} =  (B B^\top+\epsilon \mathbb{I})^{-1/2}\left(B S_{i-1}S_{i-1}^\top B^\top - B_{[i-1]} B_{[i-1]}^\top\right)(B B^\top+\epsilon \mathbb{I})^{-1/2}.
\]
\rev{We now describe a modification of the martingale allowing for using Theorem~\ref{thm:Minsker} by identifying the constants $U$ and $\sigma^2$ in the tail bound.}

We first observe that, if $\|Y_{i-1}\|_{2}<t$ at step $i-1$, then by construction $\|Y_{j}\|_{2}<t$ for all previous steps $j<i-1$. If, at some point $\|Y_{i-1}\|_{2}\geq t$, then we set $X_j=0$ for all subsequent steps $j\geq i$. By abusing notations, we denote this  stopped process by the same symbol $(Y_i)_i$\rev{; for more details about stopped martingales, we refer to~\citet[Section 7]{TroppLectureNotes}. The basic idea is that, if the norm of the stopped martingale at step $i-1$ is strictly smaller than $t$, the same is true for the original martingale defined in~\eqref{eq:X_i}. The stopped martingale facilitates the analysis, since as soon as it has stopped the difference sequence satisfies $X_i = 0$.}
Again, \rev{if the martingale has not yet stopped $\|Y_{i-1}\|_{2}<t$ and $X_i$ is given by~\eqref{eq:X_i}}, and we have 
$$
(B B^\top+\epsilon \mathbb{I})^{-1/2}\left(B S_{i-1}S_{i-1}^\top B^\top - B_{[i-1]} B_{[i-1]}^\top\right)(B B^\top+\epsilon \mathbb{I})^{-1/2}\prec t \mathbb{I},
$$
which gives the following inequality
\begin{equation}
    B S_{i-1}S_{i-1}^\top B^\top \prec B_{[i-1]} B_{[i-1]}^\top + t (B B^\top+\epsilon \mathbb{I}).\label{eq:ineqProofThm1}
\end{equation}
Notice that, by definition, $B$ contains the columns of $B_{[i-1]}$, and therefore,  the following inequality holds
\begin{equation}
B_{[i-1]} B_{[i-1]}^\top \preceq B B^\top.\label{eq:ineqB}
\end{equation}
\noindent\textbf{Lower bounding the sampling probability.} At this point, we want to show that the norm of the difference sequence $X_i$ is bounded in order to use the matrix martingale concentration result which is given below in Theorem~\ref{thm:Minsker}. We have the elementary upper bound on the difference sequence
\[\|X_i\|_{2}\leq \max\{1, 1/p_i -1\} \|\psi_i\psi_i^\top\|_{2} \leq\frac{1}{p_i} \psi_i^\top \psi_i.\]
However, the RHS of the above bound can be arbitrary large if $p_i$ goes to zero. By using \eqref{eq:ineqProofThm1} and \eqref{eq:ineqB}, we show that the sampling probability $p_i$ is lower bounded as follows
\begin{equation}
    p_i\geq \min\{c \psi_i^\top \psi_i, 1\}.\label{eq:bound_on_pi}
\end{equation}
The argument to prove~\eqref{eq:bound_on_pi} goes as follows.
First, we recall the definition of the sampling probability at step $i$, as it is given in Section~\ref{sec:RAS} by
$$
p_i \triangleq \min\{1, c\tilde{l}_i\},
$$
with $\tilde{l}_i \triangleq \min\{1,(1+t) s_i\}$ and $$s_i \triangleq b_i^\top(B S_{i-1}S_{i-1}^\top B^\top + \epsilon\mathbb{I})^{-1}b_i
= \frac{1}{\epsilon} \Big[P_{n\gamma}(K) -L_{\epsilon,S_{i-1}}(P_{n\gamma}(K))\Big]_{ii},$$ while $b_i$ is the $i$-th column of $B$. Next, we obtain the following set of inequalities
\begin{align*}
\tilde{l}_i &= \min\{1,(1+t) b_i^\top(B S_{i-1}S_{i-1}^\top B^\top + \epsilon\mathbb{I})^{-1}b_i\} & \\
 &\geq \min\{1,(1+t) b_i^\top\Big( B_{[i-1]}B_{[i-1]}^\top + \epsilon\mathbb{I} + t (B B^\top+\epsilon \mathbb{I})\Big)^{-1}b_i\} & \text{(by using \eqref{eq:ineqProofThm1})}\\
 &\geq \min\{1,(1+t) b_i^\top\Big( BB^\top + \epsilon\mathbb{I} + t (B B^\top+\epsilon \mathbb{I})\Big)^{-1}b_i\} & \text{(by using \eqref{eq:ineqB})}\\
  &\geq \min\{1,b_i^\top (B B^\top+\epsilon \mathbb{I})^{-1}b_i\} = b_i^\top (B B^\top+\epsilon \mathbb{I})^{-1}b_i =  \psi_i^\top \psi_i.
\end{align*}
So, we have $p_i\geq \min\{c \psi_i^\top \psi_i, 1\}$. If $p_i = 1$, then $X_i = 0$ as a direct consequence of the definition in~\eqref{eq:X_i}, so that the step $i$ does not change the value of the martingale. Otherwise, if $p_i<1$, we have $p_i\geq c \psi_i^\top \psi_i$.

\noindent\textbf{Concentration bound.} We can now estimate the probability of failure by using a concentration bound given in Theorem~\ref{thm:Minsker}.
Let us check that the hypotheses of Theorem~\ref{thm:Minsker} are satisfied.
Clearly, each increment is bounded as follows: \[\|X_i\|_{2}\leq \max\{1, 1/p_i -1\} \|\psi_i\psi_i^\top\|_{2} \leq\frac{1}{p_i} \psi_i^\top \psi_i\leq 1/c \triangleq U,\]
\rev{almost surely.}
Similarly, by using once more $\psi_i^\top \psi_i\leq p_i/c$, we find:
\[
\mathbb{E}_{i-1}\left(X_i^2\right) \preceq p_i(1/p_i-1)^2 \left(\psi_i \psi_i^\top\right)^2 +(1-p_i) \left(\psi_i \psi_i^\top\right)^2 \preceq \frac{1}{p_i}  \left(\psi_i \psi_i^\top\right)^2\preceq \psi_i \psi_i^\top/c.
\]
Then, the predictable quadratic variation satisfies the following upper bound
\[
W_n \preceq \frac{1}{c}\sum_{i=1}^n \psi_i\psi_i^\top = \frac{1}{c} P_\epsilon(P_{n\gamma}(K)) = \frac{1}{c}\frac{1}{1+\epsilon} P_{\frac{\epsilon n\gamma}{1+\epsilon}}(K) \text{ almost surely,}
\]
where the last equality is due to Lemma~\ref{lem:CompositionRule}. Furthermore,  we have $\lambda_{max}(W_n)\leq 1/c =\sigma^2$ almost surely\footnote{Notice that $\sigma^2$ denotes here an upper bound on the predictable quadratic variation and not the bandwidth of a kernel.}.
\rev{Now, for a symmetric matrix $M$, we recall that $\min\{M,1\}$ is defined on the eigenvalues of $M$. Next, we use that $\min\{\lambda,1\}\leq \lambda $ for all real $\lambda$}.  Hence,
\[
\Tr\left(\min\{\frac{t}{U}\frac{\mathbb{E}(W_n)}{\sigma^2},1\}\right)\leq \frac{tc}{1+\epsilon}\Tr\left(P_{\frac{\epsilon n\gamma}{1+\epsilon}}(K)\right).
\]
Theorem~\ref{thm:Minsker} can now be used to find how large the oversampling parameter $c>0$ has to be so that $\lambda_{\rm max}( Y_n ) \leq t$.

\noindent\textbf{Finding a lower bound on $c$.} In view of Theorem~\ref{thm:Minsker}, we require that $t\geq \frac{1}{c}\frac{1+\sqrt{37}}{6}$. Let the accuracy parameter be $t=1/2$ in order to simplify the result; cfr.~\citet{MuscoMusco} where a similar choice is done.
This implies then that $c\geq \frac{1+\sqrt{37}}{3}$. Another lower bound on $c$ is obtained by bounding the  failure probability. Namely, the failure probability given by \eqref{eq:MinskerBound} in Theorem~\ref{thm:Minsker}  is upper bounded as follows:
\[
\Pr(\|Y_n\|_{2} >t)\leq 50 \Tr\left(\min\{\frac{t}{U}\frac{\mathbb{E}(W_n)}{\sigma^2},1\}\right) e^{\frac{-t^2/2}{\sigma^2+tU/3}} \leq \frac{50 t c}{1+\epsilon} d_{\rm eff} e^{\frac{-t^2/2}{1/c+t/(3c)}},
\]
where $d_{\rm eff} = d_{\rm eff}(\epsilon n\gamma/(1+\epsilon))$.
We now find how $c>0$ can be chosen such that this upper bound on the failure probability is smaller than $\delta$. For $t=1/2$ we have the condition:

\[
\frac{25c}{1+\epsilon} \frac{d_{\rm eff}}{\delta}\leq e^{\frac{c}{8(1+1/6)}}.
\]
Hence, by a change of variables, we can phrase this inequality in the form $\exp(x)\geq a x$, which can be written equivalently as follows:
\[
x\geq - W_{-1}(-1/a),\qquad \text{ for } x\geq 1 \text{ and } a\geq e,
\]
where $W_{-1}:[-1/e, 0(\to \mathbb{R}$ is the negative branch of the Lambert function (see Section 3. of~\citet{Corless}), which is a monotone decreasing function.
 Notice that the asymptotic expansion of this branch of the Lambert function is
$\lim_{y \to 0_{-}} W_{-1}(y)/\log(-y) = 1.$ Therefore, we can write $- W_{-1}(-1/a) \sim \log(a)$ (with $a>0$), so that we finally have the conditions:
\[
c\geq -\frac{28}{3}W_{-1}\left(-\frac{3(1+\epsilon)\delta}{700d_{\rm eff}}\right), \text{ and } c\geq \frac{1+\sqrt{37}}{3},
\]
where $d_{\rm eff} = d_{\rm eff}(\epsilon n\gamma/(1+\epsilon))$. This proves the desired result.

\end{proof}

\section{Ablation study for different bandwidths of the Gaussian kernel \label{app:abBandwidth}}

\begin{figure}[h]
	\centering
	\begin{subfigure}[b]{0.44\textwidth}
		\includegraphics[width=\textwidth]{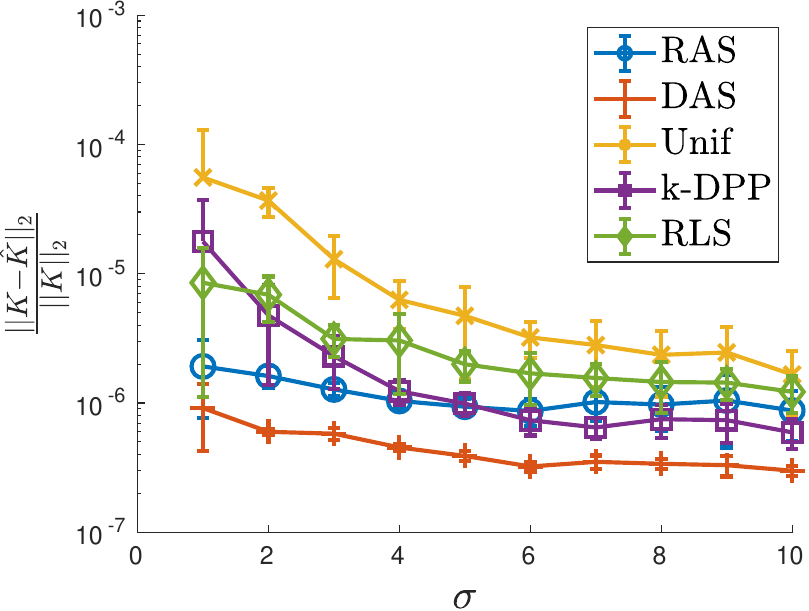}
		\caption{\texttt{Stock}}
	\end{subfigure}
	\hfill
	\begin{subfigure}[b]{0.44\textwidth}
		\includegraphics[width=\textwidth]{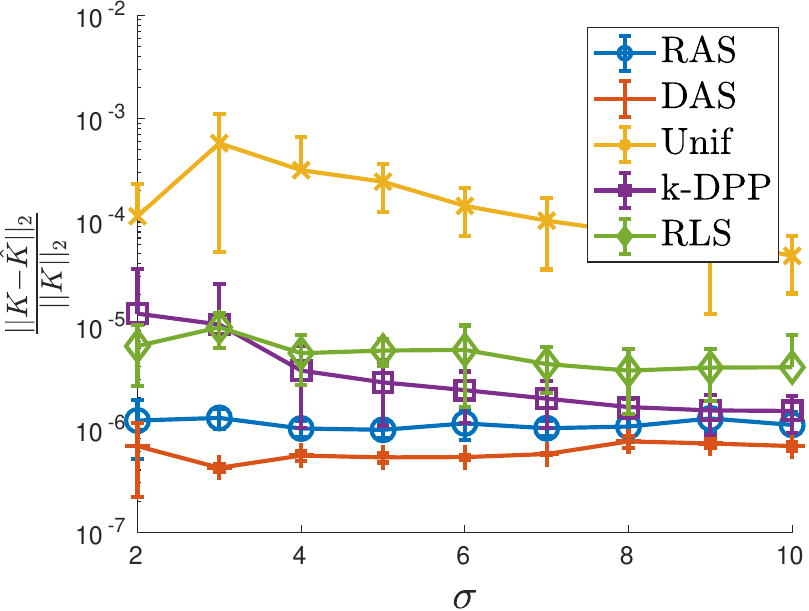}
		\caption{\texttt{Housing}}
	\end{subfigure}
	\begin{subfigure}[b]{0.44\textwidth}
		\includegraphics[width=\textwidth]{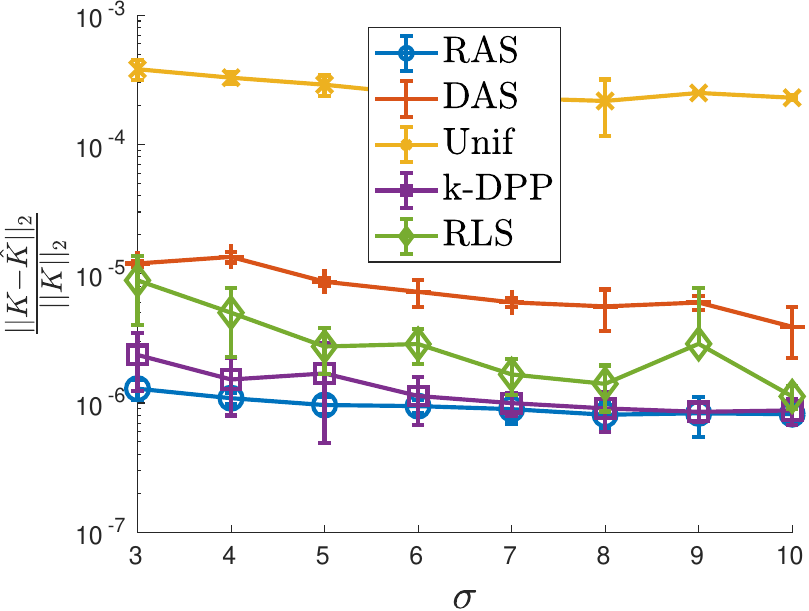}
		\caption{\texttt{Abalone}}
	\end{subfigure}
	\hfill
	\begin{subfigure}[b]{0.44\textwidth}
		\includegraphics[width=\textwidth]{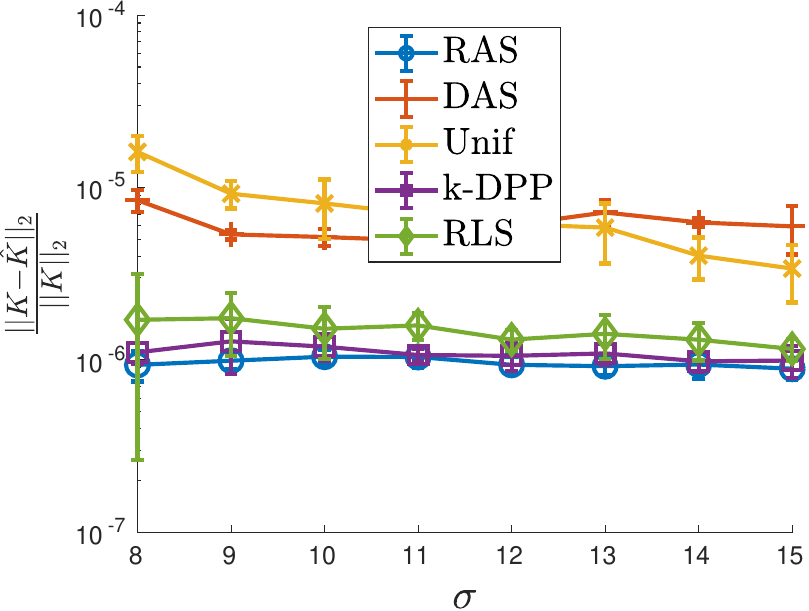}
		\caption{\texttt{Bank8FM}}
	\end{subfigure}
	\caption{The Relative operator norm of the Nystr\"om approximation error as a function of the bandwidth $\sigma$. The error bars are the standard deviations.\label{fig:bwExp_OP}}
\end{figure}

\begin{figure}[h]
	\centering
	\begin{subfigure}[b]{0.44\textwidth}
		\includegraphics[width=\textwidth]{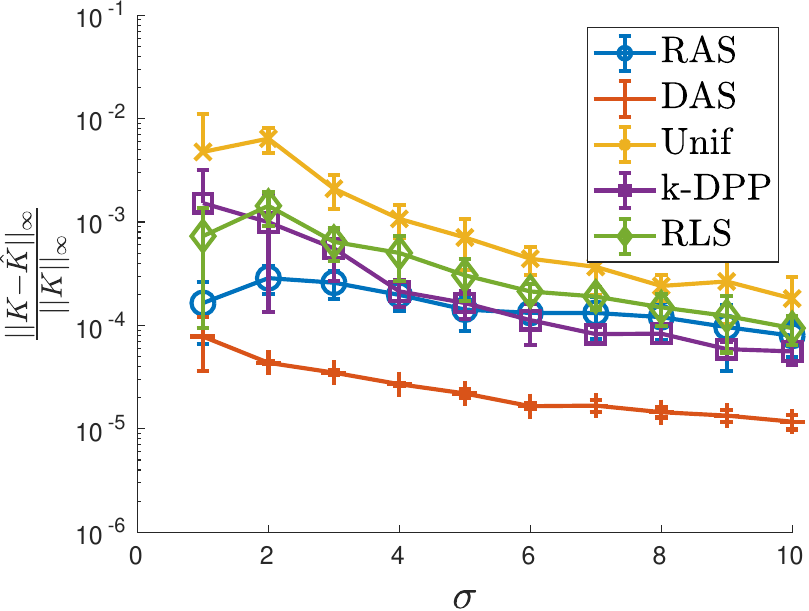}
		\caption{\texttt{Stock}}
	\end{subfigure}
	\hfill
	\begin{subfigure}[b]{0.44\textwidth}
		\includegraphics[width=\textwidth]{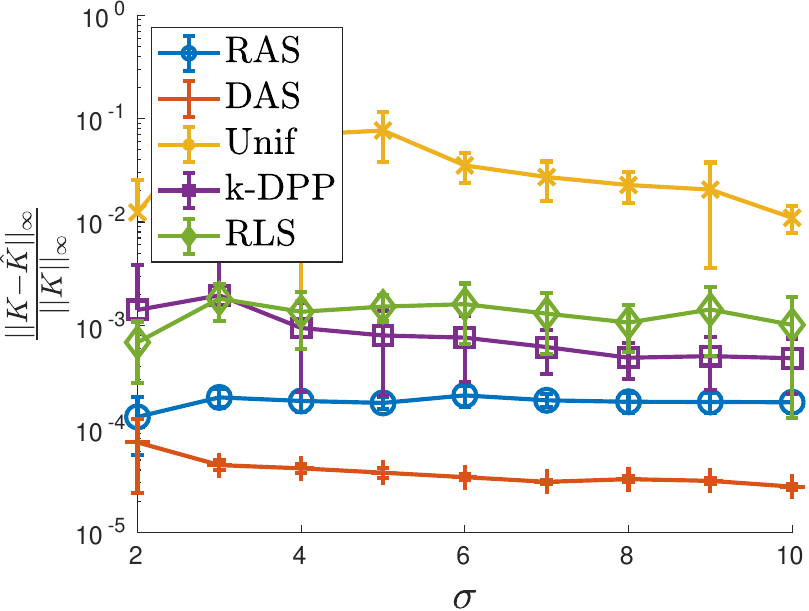}
		\caption{\texttt{Housing}}
	\end{subfigure}
	\begin{subfigure}[b]{0.44\textwidth}
		\includegraphics[width=\textwidth]{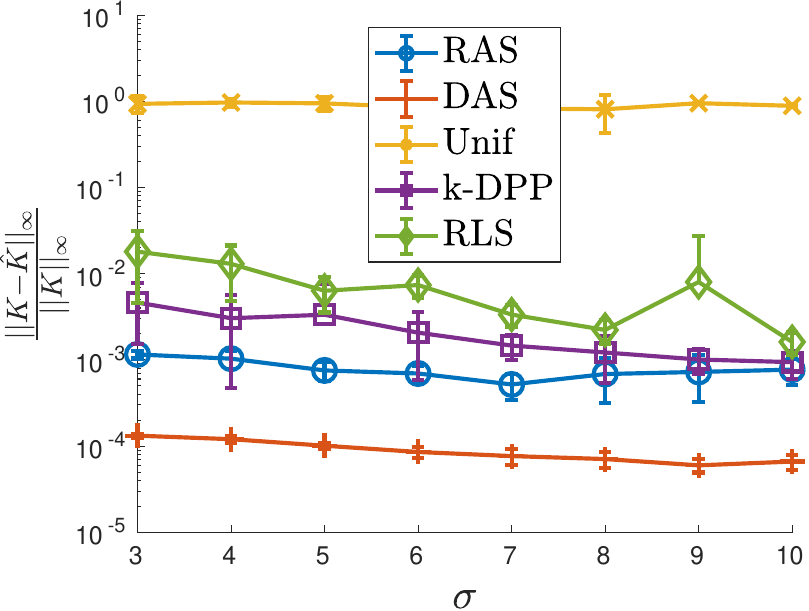}
		\caption{\texttt{Abalone}}
	\end{subfigure}
	\hfill
	\begin{subfigure}[b]{0.44\textwidth}
		\includegraphics[width=\textwidth]{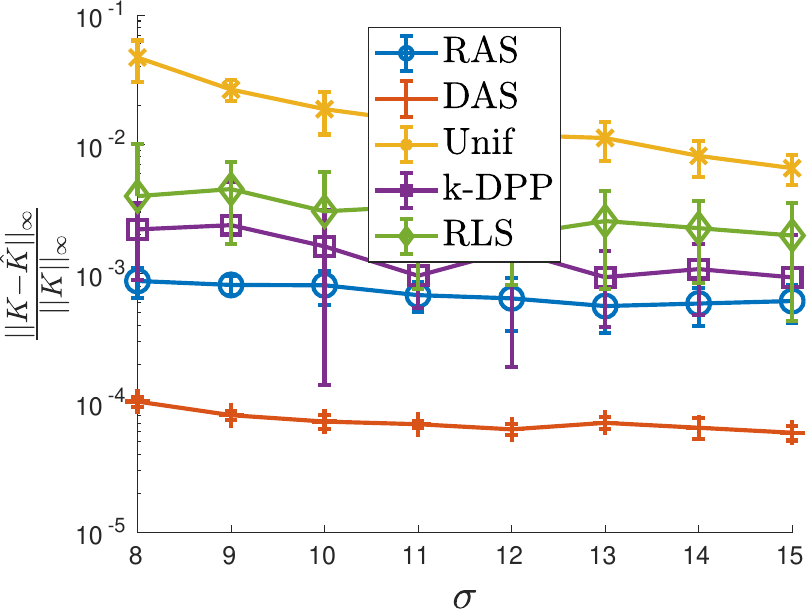}
		\caption{\texttt{Bank8FM}}
	\end{subfigure}
	\caption{The relative max norm of the Nystr\"om approximation error as a function of the bandwidth $\sigma$. The error bars are the standard deviations.\label{fig:bwExp_Max}}
\end{figure}

Here we repeat the Nystr\"om experiment of Section~\ref{sec:ExactAlgo} for different bandwidths of the Gaussian kernel. Namely, we take a fixed regularization $\gamma = 10^{-4}$ in RAS, but vary the bandwidth $\sigma$ of the Gaussian kernel. The parameter $c$ is not changed and remains the same as in Table~\ref{Table:data} for all the datasets. Note that a different $\sigma$ results in a different spectrum of the kernel matrix, consequently the RAS algorithm outputs a different number of landmarks for each setting. The best performing subset is chosen as representative. The experiment is repeated $10$ times and the averaged results are given on Figures~\ref{fig:bwExp_OP} and \ref{fig:bwExp_Max}. The conclusions from the experiment in Section~\ref{sec:ExactAlgo} are shown to be consistent over a range of $\sigma$'s.
\section{Simulation details, additional tables and figures\label{app:AddTable}}
\begin{figure}[h]
	\centering
	\begin{subfigure}[b]{0.24\textwidth}
		\includegraphics[width=\textwidth]{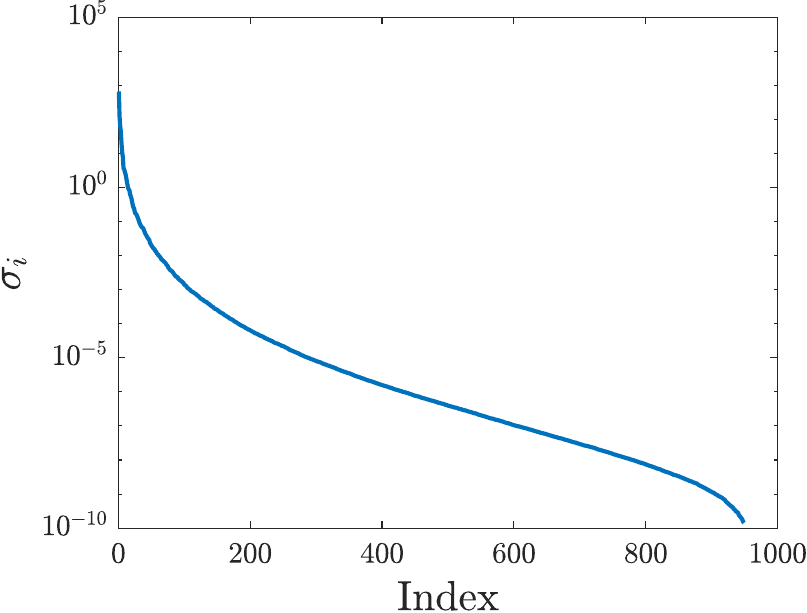}
		\caption{\texttt{Stock}}
	\end{subfigure}
	\hfill
	\begin{subfigure}[b]{0.24\textwidth}
		\includegraphics[width=\textwidth]{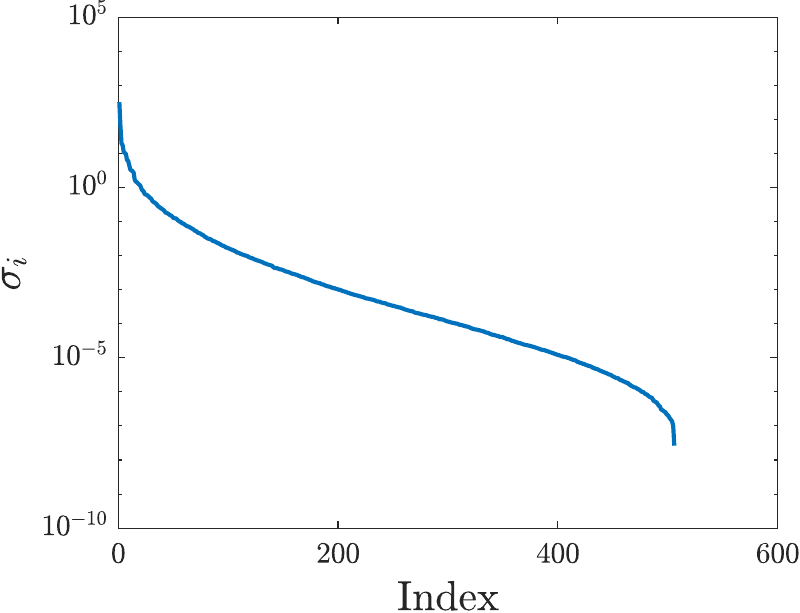}
		\caption{\texttt{Housing}}
	\end{subfigure}
	\begin{subfigure}[b]{0.24\textwidth}
		\includegraphics[width=0.96\textwidth]{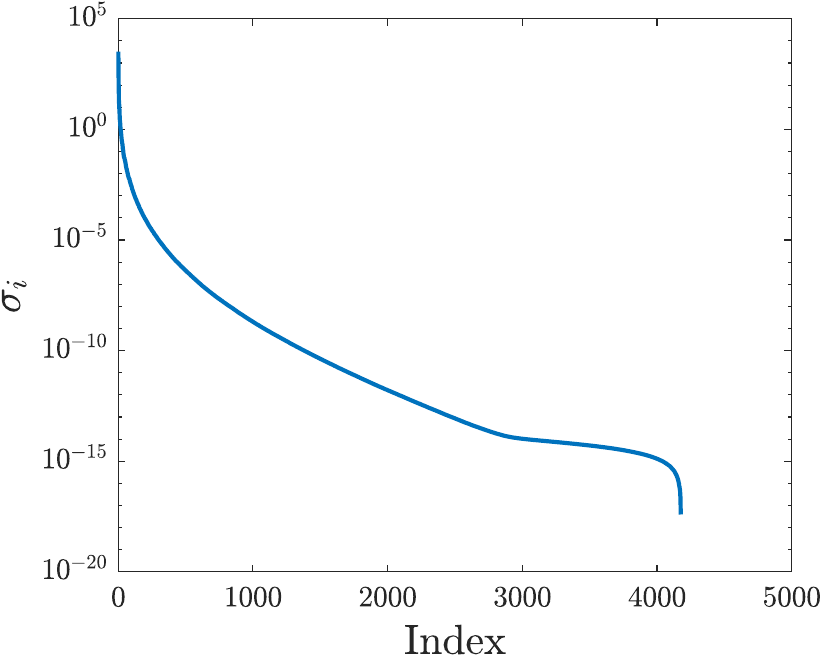}
		\caption{\texttt{Abalone}}
	\end{subfigure}
	\hfill
	\begin{subfigure}[b]{0.24\textwidth}
		\includegraphics[width=0.96\textwidth]{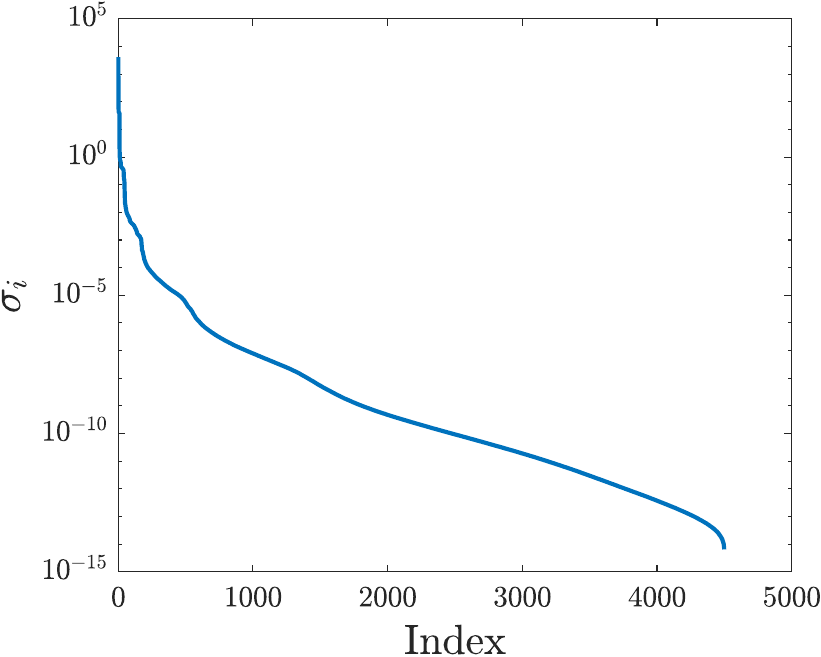}
		\caption{\texttt{Bank8FM}}
	\end{subfigure}
	\caption{Singular value spectrum of the data sets of Table~\ref{Table:data} on a logarithmic scale. For a given index (e.g., 100), the value of the eigenvalues for \texttt{Stock} and \texttt{Housing} are considerably smaller than \texttt{Abalone} and \texttt{Bank8FM}.\label{fig:Eig}}
\end{figure}

\begin{figure}[h]
	\centering
	\begin{minipage}{0.88\textwidth}
		\begin{subfigure}[t]{0.45\textwidth}
			\includegraphics[width=0.9\textwidth]{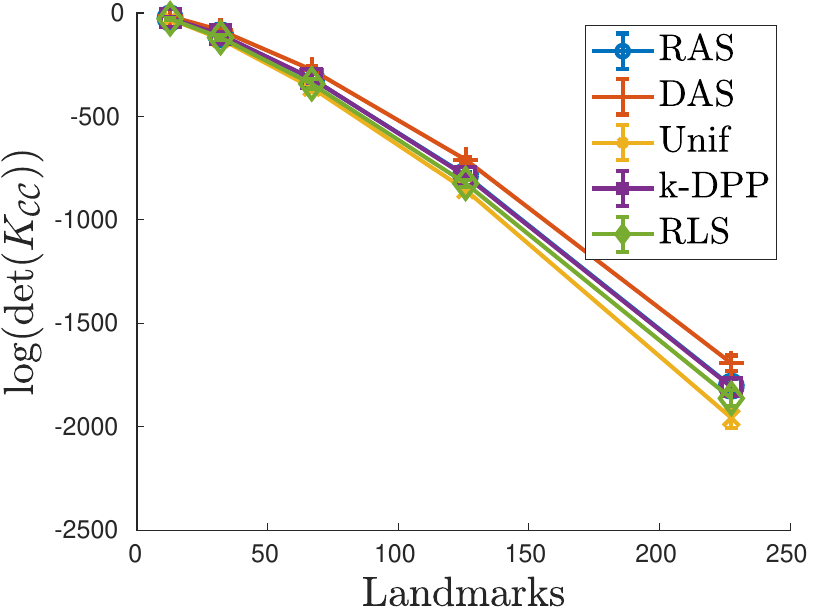}
			\caption{\texttt{Stock}}
		\end{subfigure}
		\quad
		\begin{subfigure}[t]{0.45\textwidth}
			\includegraphics[width=0.9\textwidth]{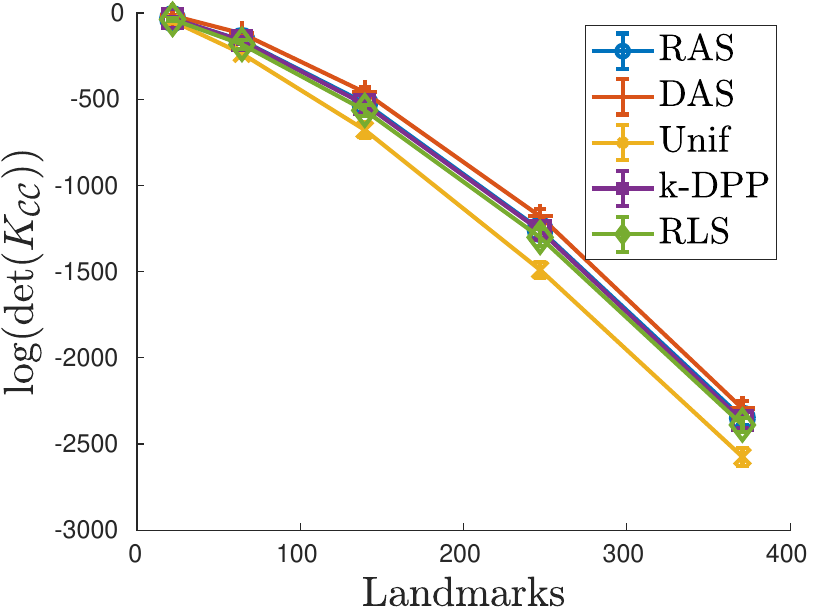}
			\caption{\texttt{Housing}}
		\end{subfigure}
	\end{minipage}
	\begin{minipage}{0.88\textwidth}
		\begin{subfigure}[b]{0.45\textwidth}
			\includegraphics[width=0.9\textwidth]{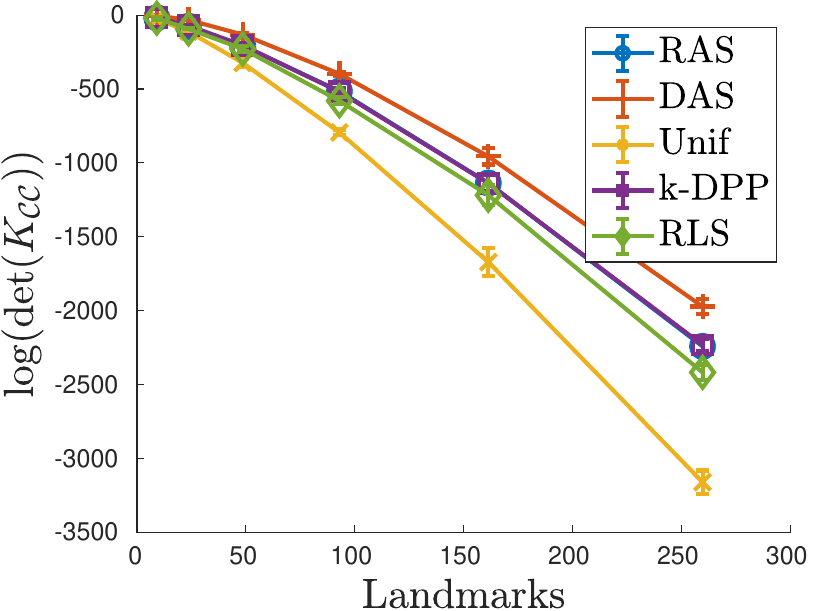}
			\caption{\texttt{Abalone}}
		\end{subfigure}
		\quad
		\begin{subfigure}[b]{0.45\textwidth}
			\includegraphics[width=0.9\textwidth]{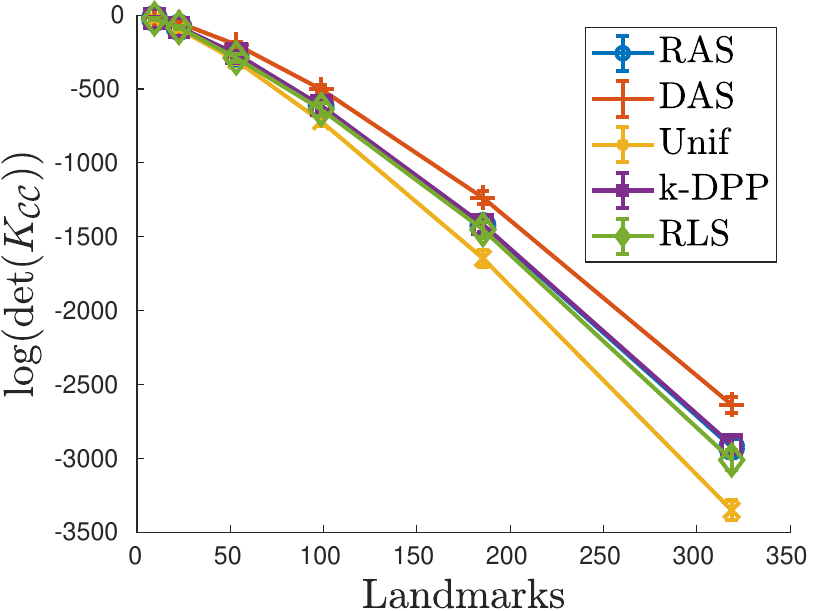}
			\caption{\texttt{Bank8FM}}
		\end{subfigure}
	\end{minipage}
	\caption{Diversity as a function of the number of landmarks. The $\mathrm{log}\mathrm{det}(K_{\mathcal{C}\mathcal{C}})$ is plotted on a logarithmic scale, averaged over 10 trials. The error bars are the standard deviations.\label{fig:DetEr}}
\end{figure}

\begin{figure}[h]
	\centering
	\begin{minipage}{0.88\textwidth}
		\begin{subfigure}[t]{0.45\textwidth}
			\includegraphics[width=0.9\textwidth]{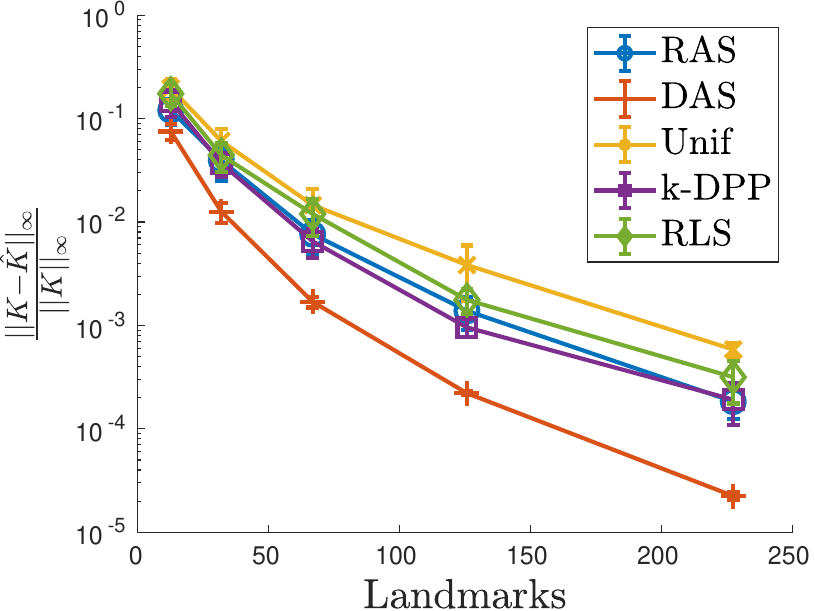}
			\caption{\texttt{Stock}}
		\end{subfigure}
		\quad
		\begin{subfigure}[t]{0.45\textwidth}
			\includegraphics[width=0.9\textwidth]{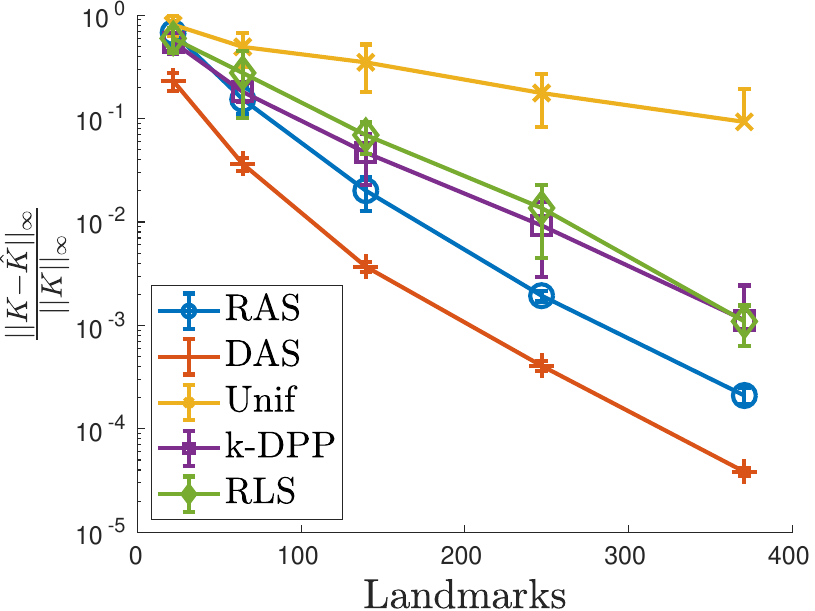}
			\caption{\texttt{Housing}}
		\end{subfigure}
	\end{minipage}
	\begin{minipage}{0.88\textwidth}
		\begin{subfigure}[b]{0.45\textwidth}
			\includegraphics[width=0.9\textwidth]{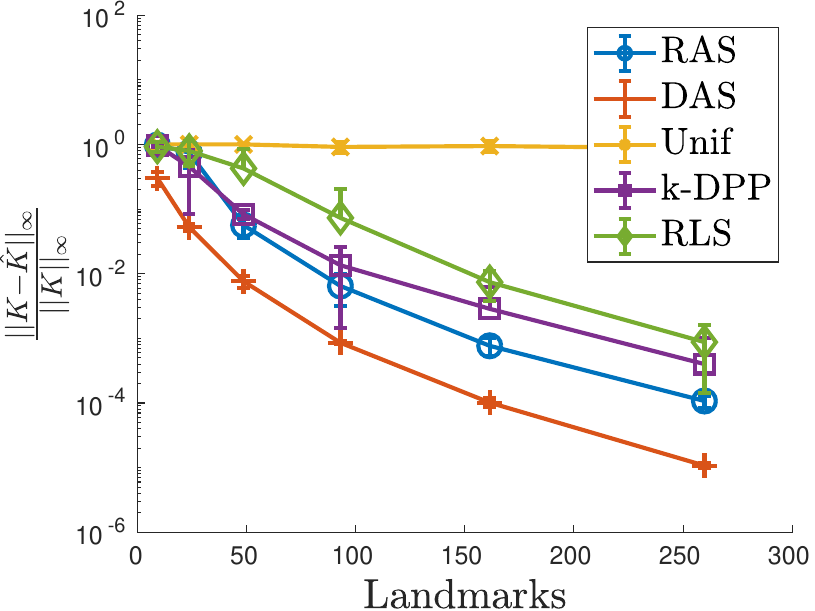}
			\caption{\texttt{Abalone}}
		\end{subfigure}
		\quad
		\begin{subfigure}[b]{0.45\textwidth}
			\includegraphics[width=0.9\textwidth]{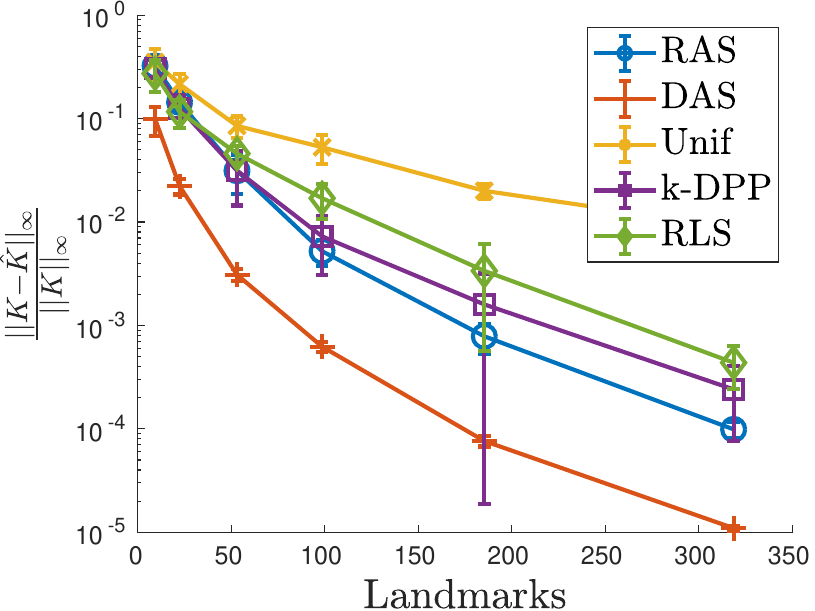}
			\caption{\texttt{Bank8FM}}
		\end{subfigure}
	\end{minipage}
	\caption{Relative max norm with error bars of the approximation as a function of the number of landmarks. The error is plotted on a logarithmic scale, averaged over 10 trials.\label{fig:MaxEr}}
\end{figure}

The computer used for the small-scale simulations has a processor with $8$ cores $3.40GHz$ and $15.5$ GB of RAM. The implementation of the algorithms is done with matlabR2018b.
For the medium scale experiments, the implementation of RAS was done in Julia1.0 while the simulations were performed on a computer with a $12\times 3.7$ GHz processors and $62$ GB RAM. 
\begin{table}[h]
	\begin{center}
			\caption{Parameters for the samplings with BLESS. \label{tab:BLESSparameters}}
		\begin{tabular}{rccccc}
			\toprule
			data set & $\sigma$ & $q_0$ & $c_0$& $c_1$& $c_2$ \\ \midrule
			\texttt{Super} & 3 & 2 & 2 & 3 & 3 \\
			\texttt{CASP} & 1 & 2 & 2 & 3 & 3 \\
			\texttt{Adult} & 10 & 2 & 2 & 3 & 2.2\\
			\texttt{MiniBooNE} & 5,10 & 2 & 2 & 3 & 3.6,4\\
			\texttt{cod-RNA} & 4 & 2 & 2 & 3 & 2.9\\
			\texttt{Covertype} & 10 & 2 & 2 & 3 & 3\\
			\bottomrule
		\end{tabular}
	\end{center}
\end{table}
Also, we provide here additional explanatory figures concerning the data sets of Table~\ref{Table:data}. In Figure~\ref{fig:Eig}, the eigenvalues of the kernel matrices are displayed. The decay of the spectrum is faster for \texttt{Stock} and \texttt{Housing}. Furthermore, the accuracy of the kernel approximation for the max norm is also given in Figure~\ref{fig:MaxEr}. Namely, the DAS method of Algorithm~\ref{Alg1} shows a better approximation accuracy for this specific norm. As a measure of diversity the $\mathrm{log}\mathrm{det}(K_{\mathcal{C}\mathcal{C}}) = \sum_{i =1}^{|\mathcal{C}|} \mathrm{log}(\lambda_i)$ with $\lambda_1,...,\lambda_{|\mathcal{C}|}$ the eigenvalues of $K_{\mathcal{C}\mathcal{C}}$, is given in Figure~\ref{fig:DetEr}. The DAS method shows the largest determinant, RAS has a similar determinant as the k-DPP.

\begin{figure}[h]
	\centering
	\begin{subfigure}[b]{0.49\textwidth}
		\includegraphics[width=\textwidth]{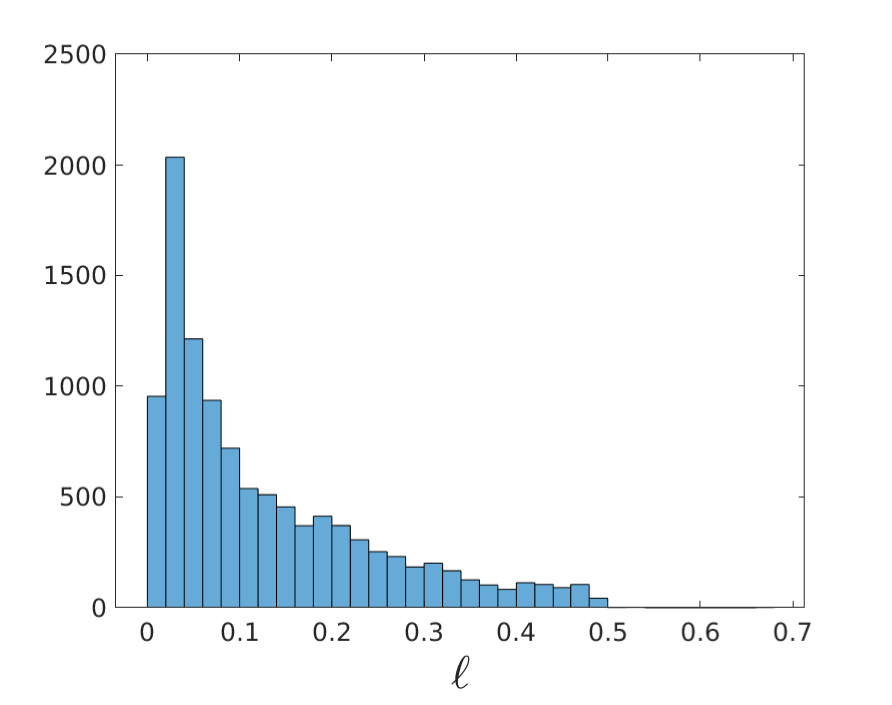}
		\caption{\texttt{Super}}
	\end{subfigure}
	\begin{subfigure}[b]{0.49\textwidth}	\includegraphics[width=\textwidth]{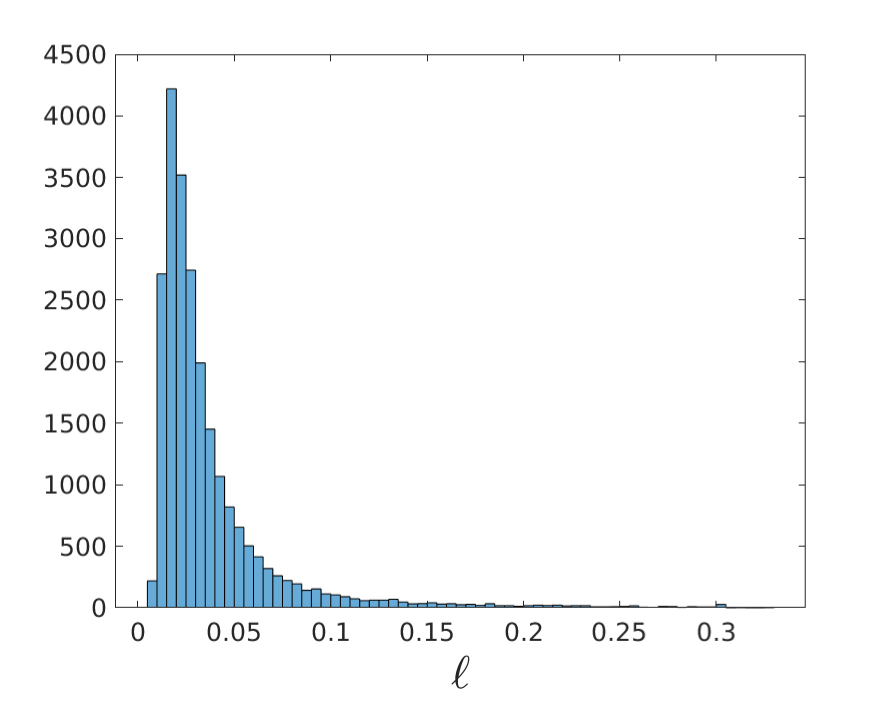}
		\caption{\texttt{CASP}}
	\end{subfigure}
	\caption{The approximate ridge leverage scores accompanying Figure \ref{fig:Regression}.}
	\label{fig:Reg_RLS}
\end{figure}

\begin{figure}[h]
	\centering
\begin{minipage}{0.88\textwidth}
	\begin{subfigure}[t]{0.45\textwidth}
		\includegraphics[width=0.9\textwidth]{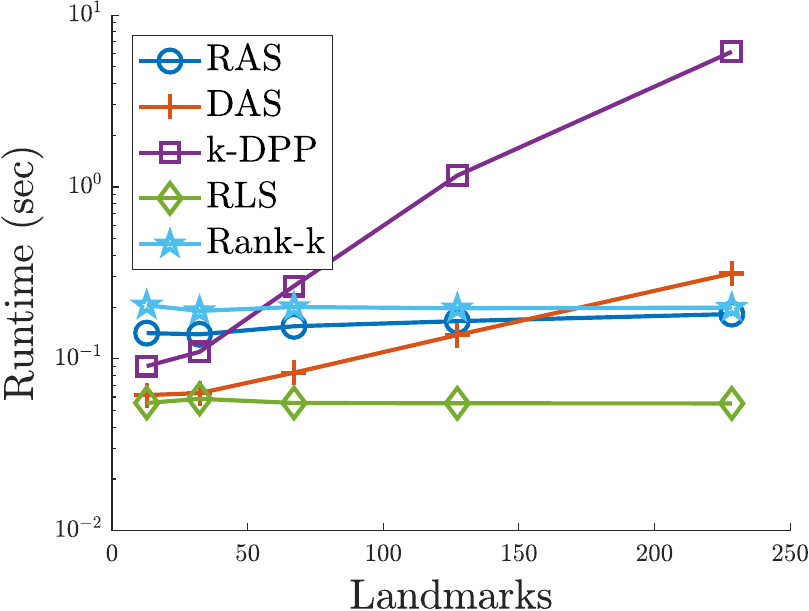}
		\caption{Stock}
	\end{subfigure}
	\quad
	\begin{subfigure}[t]{0.45\textwidth}
		\includegraphics[width=0.9\textwidth]{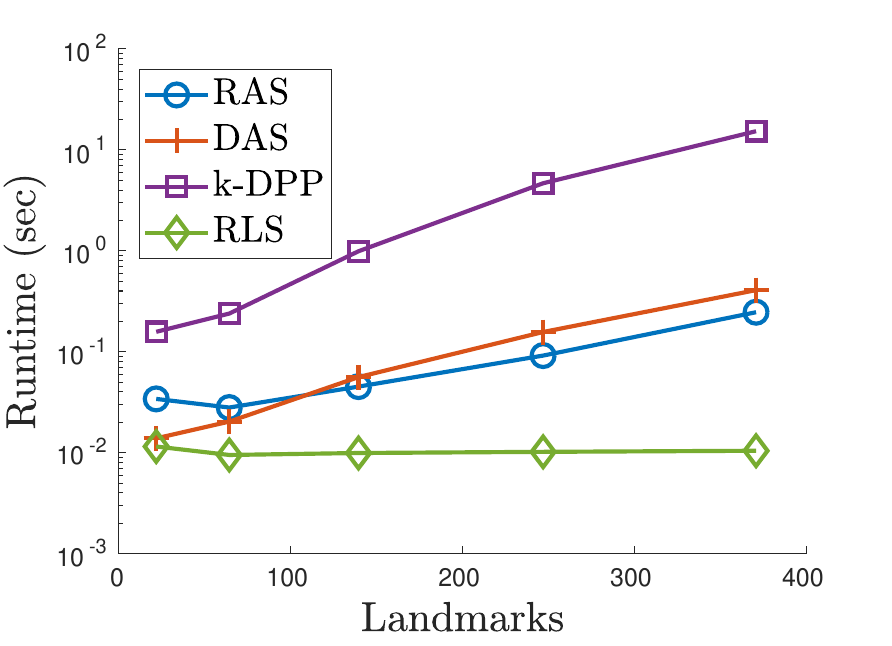}
		\caption{Housing}
	\end{subfigure}
\end{minipage}
\begin{minipage}{0.88\textwidth}
	\begin{subfigure}[b]{0.45\textwidth}
		\includegraphics[width=0.9\textwidth]{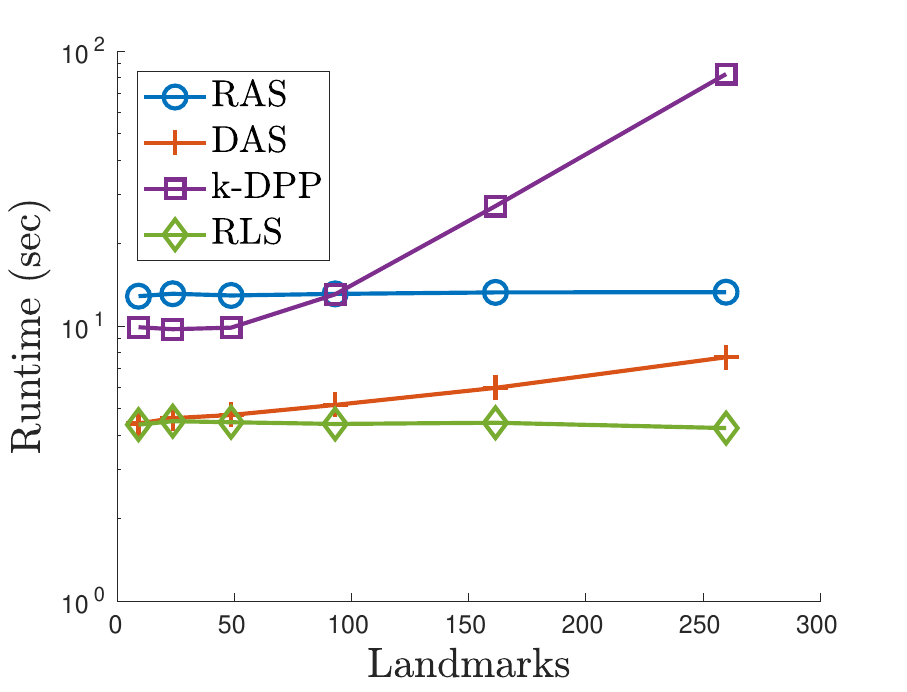}
		\caption{Abalone}
	\end{subfigure}
	\quad
	\begin{subfigure}[b]{0.45\textwidth}
		\includegraphics[width=0.9\textwidth]{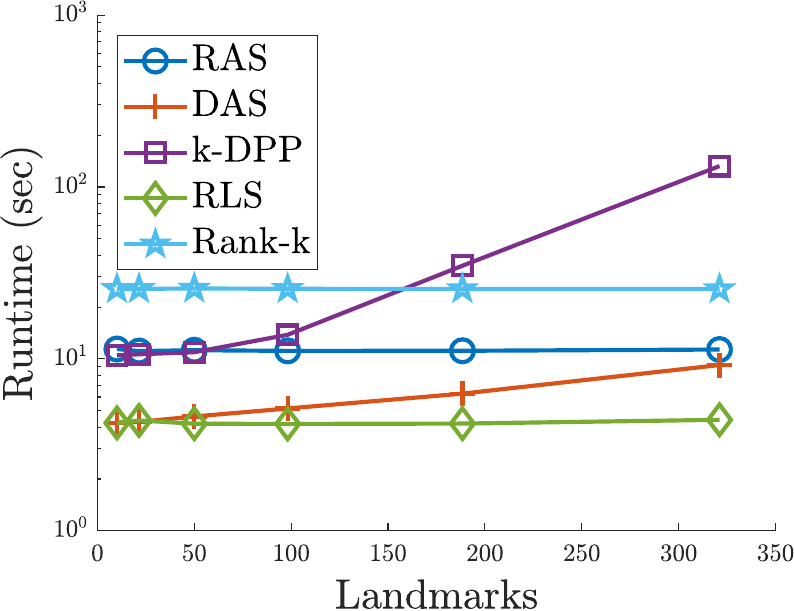}
		\caption{Bank8FM}
	\end{subfigure}
\end{minipage}
	\caption{Timings for the computations of Figure~\ref{fig:OP} as a function of the number of landmarks. The timings are plotted on a logarithmic scale, averaged over 10 trials.\label{fig:Time}}
\end{figure}

\FloatBarrier
\bibliographystyle{spbasic}
\bibliography{References}
\end{document}